%% file: paper.tex
\documentclass{article} 
\usepackage{iclr2017_conference,times}
\usepackage{hyperref}
\usepackage{url}
\usepackage{graphicx}
\usepackage{amsmath}
\usepackage{multirow}

\input{macros}

\title{Adversarial examples for generative models}

\author{Jernej Kos \\
National University of Singapore\\
\And
Ian Fischer \\
Google Research \\
\And
Dawn Song \\
University of California, Berkeley\\
}

\DeclareMathOperator{\argmin}{argmin}

\begin{document}

\maketitle

\begin{abstract}
We explore methods of producing adversarial examples on deep generative models such as the variational autoencoder (VAE) and the VAE-GAN.
Deep learning architectures are known to be vulnerable to adversarial examples, but previous work has focused on the application of adversarial examples to classification tasks.
Deep generative models have recently become popular due to their ability to model input data distributions and generate realistic examples from those distributions.
We present three classes of attacks on the VAE and VAE-GAN architectures and demonstrate them against networks trained on MNIST, SVHN and CelebA.
Our first attack leverages classification-based adversaries by attaching a classifier to the trained encoder of the target generative model, which can then be used to indirectly manipulate the latent representation.
Our second attack directly uses the VAE loss function to generate a target reconstruction image from the adversarial example.
Our third attack moves beyond relying on classification or the standard loss for the gradient and directly optimizes against differences in source and target latent representations.
We also motivate why an attacker might be interested in deploying such techniques against a target generative network.
\end{abstract}

\input{introduction}
\input{background}
\input{problem-definition}

\input{attack-methodology}
\input{measuring-effectiveness}
\input{evaluation}

\input{conclusion}

\subsubsection*{Acknowledgments}

This material is in part based upon work supported by the National Science Foundation under Grant No. TWC-1409915.
Any opinions, findings, and conclusions or recommendations expressed in this material are those of the author(s) and do not necessarily reflect the views of the National Science Foundation.

\bibliography{paper}
\bibliographystyle{iclr2017_conference}

\input{appendix}

\end{document}

%% file: macros.tex
\usepackage{amsmath}
\usepackage{amssymb}
\usepackage{amsthm}

\newcommand{\myvec}[1]{\mathbf{#1}}
\newcommand{\myvecsym}[1]{\boldsymbol{#1}}

\DeclareMathOperator{\KL}{KL}

\newcommand{\vvarepsilon}{\myvecsym{\varepsilon}}

\newcommand{\vmu}{\myvecsym{\mu}}

\newcommand{\vsigma}{\myvecsym{\sigma}}

\newcommand{\ttilde}{\tilde{t}}

\newcommand{\vx}{\myvec{x}}
\newcommand{\vxh}{\hat{\vx}}

\newcommand{\vxstar}{\vx^\myvec{\ast}}
\newcommand{\vxhstar}{\vxh^\myvec{\ast}}

\newcommand{\vy}{\myvec{y}}
\newcommand{\yh}{\hat{y}}

\newcommand{\vz}{\myvec{z}}

\newcommand{\mymathcal}[1]{\mathcal{#1}}

\newcommand{\calL}{\mymathcal{L}}

\newcommand{\calX}{\mymathcal{X}}
\newcommand{\calY}{\mymathcal{Y}}

\newcommand{\fdec}{f_{\mathrm{dec}}}
\newcommand{\fenc}{f_{\mathrm{enc}}}
\newcommand{\fdisc}{f_{\mathrm{disc}}}
\newcommand{\fclass}{f_{\mathrm{class}}}
\newcommand{\vxhadv}{\vxh_{\mathrm{adv}}}
\newcommand{\Gtarg}{G_{\mathrm{targ}}}
\newcommand{\VAE}{\mathrm{VAE}}
\newcommand{\classifier}{\mathrm{classifier}}
\newcommand{\latent}{\mathrm{latent}}

\newcommand{\hmm}[1]{\ifnum\ifhmode\spacefactor\else2000\fi>1000 \uppercase{#1}\else#1\fi}

\newcommand{\ASuntargeted}{AS_{ignore-target}}
\newcommand{\AStargeted}{AS_{target}}
\newcommand{\Oracle}{\textsc{Oracle}}



%% file: introduction.tex
\section{Introduction}

Adversarial examples have been shown to exist for a variety of deep learning architectures.\footnote{
 Adversarial examples are even easier to produce against most other machine learning architectures, as shown in \citet{papernot2016practical}, but we are focused on deep networks.
}
They are small perturbations of the original inputs, often barely visible to a human observer, but carefully crafted to misguide the network into producing incorrect outputs.
Seminal work by~\citet{szegedy2013intriguing} and ~\citet{goodfellow2014explaining}, as well as much recent work, has shown that adversarial examples are abundant and finding them is easy.

Most previous work focuses on the application of adversarial examples to the task of classification, where the deep network assigns classes to input images.
The attack adds small adversarial perturbations to the original input image.
These perturbations cause the network to change its classification of the input, from the correct class to some other incorrect class (possibly chosen by the attacker).
Critically, the perturbed input must still be recognizable to a human observer as belonging to the original input class.\footnote{
 Random noise images and ``fooling'' images \citep{nguyen2014fool} do not belong to this strict definition of an adversarial input, although they do highlight other limitations of current classifiers.
}

Deep generative models, such as \citet{kingma2013auto}, learn to generate a variety of outputs, ranging from handwritten digits to faces \citep{kulkarni2015deep}, realistic scenes \citep{oord2016conditional}, videos \citep{kalchbrenner2016video}, 3D objects \citep{dosovitskiy2016learning}, and audio \citep{vandenoord2016wavenet}.
These models learn an approximation of the input data distribution in different ways, and then sample from this distribution to generate previously unseen but plausible outputs.

To the best of our knowledge, no prior work has explored using adversarial inputs to attack generative models.
There are two main requirements for such work: describing a plausible scenario in which an attacker might want to attack a generative model; and designing and demonstrating an attack that succeeds against generative models.
We address both of these requirements in this work.

One of the most basic applications of generative models is input reconstruction.
Given an input image, the model first encodes it into a lower-dimensional latent representation, and then uses that representation to generate a reconstruction of the original input image.
Since the latent representation usually has much fewer dimensions than the original input, it can be used as a form of compression.
The latent representation can also be used to remove some types of noise from inputs, even when the network has not been explicitly trained for denoising, due to the lower dimensionality of the latent representation restricting what information the trained network is able to represent.
Many generative models also allow manipulation of the generated output by sampling different latent values or modifying individual dimensions of the latent vectors without needing to pass through the encoding step.

These properties of input reconstruction generative networks suggest a variety of different attacks that would be enabled by effective adversaries against generative networks.
Any attack that targets the compression bottleneck of the latent representation can exploit natural security vulnerabilities in applications built to use that latent representation.
Specifically, if the person doing the encoding step is separated from the person doing the decoding step, the attacker may be able to cause the encoding party to believe they have encoded a particular message for the decoding party, but in reality they have encoded a different message of the attacker's choosing.
We explore this idea in more detail as it applies to the application of compressing images using a VAE or VAE-GAN architecture.

%% file: background.tex
\section{Related work and background}
\label{sec:related-work-background}

This work focuses on adversaries for variational autoencoders (VAEs, proposed in \citet{kingma2013auto}) and VAE-GANs (VAEs composed with a generative adversarial network, proposed in \citet{larsen2015autoencoding}).

\subsection{Related work on adversaries}
\label{sec:related-work-adversarial}

Many adversarial attacks on classification models have been described in existing literature~\citep{goodfellow2014explaining, szegedy2013intriguing}.
These attacks can be untargeted, where the adversary's goal is to cause any misclassification, or the least likely misclassification \citep{goodfellow2014explaining, kurakin2016adv}; or they can be targeted, where the attacker desires a specific misclassification.
\citet{moosavi2015deepfool} gives a recent example of a strong targeted adversarial attack.
Some adversarial attacks allow for a threat model where the adversary does not have access to the target model~\citep{szegedy2013intriguing, papernot2016practical}, but commonly it is assumed that the attacker does have that access, in an online or offline setting~\citep{goodfellow2014explaining, kurakin2016adv}.\footnote{
 See \citet{papernot2015limitations} for an overview of different adversarial threat models.
}

Given a classifier $f(\vx): \vx \in \calX \rightarrow y \in \calY$ and original inputs $\vx \in \calX$, the problem of generating \textit{untargeted} adversarial examples can be expressed as the following optimization: $\argmin_{\vxstar} L(\vx,\vxstar)\ s.t.\ f(\vxstar) \ne f(\vx)$, where $L(\cdot)$ is a chosen distance measure between examples from the input space (e.g., the $L_2$ norm).
Similarly, generating a \textit{targeted} adversarial attack on a classifier can be expressed as $\argmin_{\vxstar} L(\vx,\vxstar)\ s.t.\ f(\vxstar) = y_t$, where $y_t \in \calY$ is some target label chosen by the attacker.

These optimization problems can often be solved with optimizers like L-BFGS or Adam \citep{kingma2014adam}, as done in \citet{szegedy2013intriguing} and \citet{carlini2016towards}.
They can also be approximated with single-step gradient-based techniques like fast gradient sign \citep{goodfellow2014explaining}, fast gradient $L_2$ \citep{huang2015adv}, or fast least likely class \citep{kurakin2016adv}; or they can be approximated with iterative variants of those and other gradient-based techniques \citep{kurakin2016adv, moosavi2015deepfool}.

An interesting variation of this type of attack can be found in \citet{sabour2015adv}.
In that work, they attack the hidden state of the target network directly by taking an input image $\vx$ and a target image $\vx_t$ and searching for a perturbed variant of $\vx$ that generates similar hidden state at layer $l$ of the target network to the hidden state at the same layer generated by $\vx_t$.
This approach can also be applied directly to attacking the latent vector of a generative model.

A variant of this attack has also been applied to VAE models in the concurrent work of \citet{tabacof2016advvae}\footnote{
 This work was made public shortly after we published our early drafts.
}, which uses the KL divergence between the latent representation of the source and target images to generate the adversarial example.
However in their paper, the authors mention that they tried attacking the output directly and that this only managed to make the reconstructions more blurry.
While they do not explain the exact experimental setting, the attack sounds similar to our $\calL_{\VAE}$ attack, which we find very successful.
Also, in their paper the authors do not consider the more advanced VAE-GAN models and more complex datasets like CelebA.

\subsection{Background on VAEs and VAE-GANs}
\label{sec:problem-definition-vae}

The general architecture of a variational autoencoder consists of three components, as shown in Figure~\ref{fig:vae-architecture}.
The \textbf{encoder $\fenc(\vx)$} is a neural network mapping a high-dimensional input representation $\vx$ into a lower-dimensional (compressed) \textbf{latent representation $\vz$}.
All possible values of $\vz$ form a latent space.
Similar values in the latent space should produce similar outputs from the decoder in a well-trained VAE.
And finally, the \textbf{decoder/generator $\fdec(\vz)$}, which is a neural network mapping the compressed latent representation back to a high-dimensional output $\vxh$.
Composing these networks allows basic input reconstruction $\vxh = \fdec(\fenc(\vx))$.
This composed architecture is used during training to backpropagate errors from the loss function.

The variational autoencoder's loss function $\calL_{\VAE}$ enables the network to learn a latent representation that approximates the intractable posterior distribution $p(\vz|\vx)$:
\begin{equation}
\label{eqn:vae-loss}
\calL_{\VAE} = -D_{\KL}[q(\vz|\vx)||p(\vz)] + E_q[\log p(\vx|\vz)] \mathrm{.}
\end{equation}
$q(\vz|\vx)$ is the learned approximation of the posterior distribution $p(\vz|\vx)$.
$p(\vz)$ is the prior distribution of the latent representation $\vz$.
$D_{\KL}$ denotes the Kullback--Leibler divergence.
$E_q[\log p(\vx|\vz)]$ is the variational lower bound, which in the case of input reconstruction is the cross-entropy $H[\vx, \vxh]$ between the inputs $\vx$ and their reconstructions $\vxh$.
In order to generate $\vxh$ the VAE needs to sample $q(\vz|\vx)$ and then compute $\fdec(\vz)$.

For the VAE to be fully differentiable while sampling from $q(\vz|\vx)$, the reparametrization trick \citep{kingma2013auto} extracts the random sampling step from the network and turns it into an input, $\vvarepsilon$.
VAEs are often parameterized with Gaussian distributions.
In this case, $\fenc(\vx)$ outputs the distribution parameters $\vmu$ and $\vsigma^2$.
That distribution is then sampled by computing $\vz = \vmu + \vvarepsilon \sqrt{\vsigma^2}$ where $\vvarepsilon \sim N(0, 1)$ is the input random sample, which does not depend on any parameters of $\fenc$, and thus does not impact differentiation of the network.

The VAE-GAN architecture of \citet{larsen2015autoencoding} has the same $\fenc$ and $\fdec$ pair as in the VAE.
It also adds a discriminator $\fdisc$ that is used during training, as in standard generative adversarial networks \citep{goodfellow2014gan}.
The loss function of $\fdec$ uses the disciminator loss instead of cross-entropy for estimating the reconstruction error.


%% file: problem-definition.tex
\section{Problem definition}

We provide a motivating attack scenario for adversaries against generative models, as well as a formal definition of an adversary in the generative setting.

\subsection{Motivating attack scenario}
\label{sec:example-attack-scenario}

\begin{figure}
\begin{center}
\includegraphics[scale=1.0]{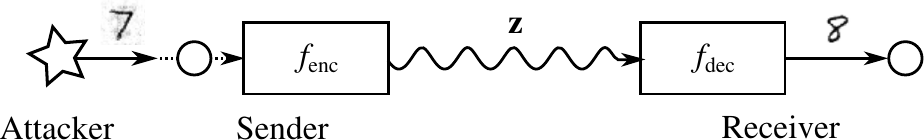}
\end{center}
\caption{Depiction of the attack scenario. The VAE is used as a compression scheme to transmit a latent representation of the image from the sender (left) to the receiver (right). The attacker convinces the sender to compress a particular image into its latent vector, which is sent to the receiver, where the decoder reconstructs the latent vector into some other image chosen by the attacker.}
\label{fig:vae-attack-scenario}
\end{figure}

To motivate the attacks presented below, we describe the attack scenario depicted in Figure~\ref{fig:vae-attack-scenario}.
In this scenario, there are two parties, the sender and the receiver, who wish to share images with each other over a computer network.
In order to conserve bandwidth, they share a VAE trained on the input distribution of interest, which will allow them to send only latent vectors $\vz$.

The attacker's goal is to convince the sender to send an image of the attacker's choosing to the receiver, but the attacker has no direct control over the bytes sent between the two parties.
However, the attacker has a copy of the shared VAE.
The attacker presents an image $\vxstar$ to the sender which resembles an image $\vx$ that the sender wants to share with the receiver.
For example, the sender wants to share pictures of kittens with the receiver, so the attacker presents a web page to the sender with a picture of a kitten, which is $\vxstar$.
The sender chooses $\vxstar$ and sends its corresponding $\vz$ to the receiver, who reconstructs it.
 %
However, because the attacker controlled the chosen image, when the receiver reconstructs it, instead of getting a faithful reproduction $\vxh$ of $\vx$ (e.g., a kitten), the receiver sees some other image of the attacker's choosing, $\vxhadv$, which has a different meaning from $\vx$ (e.g., a request to send money to the attacker's bank account).

There are other attacks of this general form, where the sender and the receiver may be separated by distance, as in this example, or by time, in the case of storing compressed images to disk for later retrieval.
In the time-separated attack, the sender and the receiver may be the same person or multiple different people.
In either case, if they are using the insecure channel of the VAE's latent space, the messages they share may be under the control of an attacker.
For example, an attacker may be able to fool an automatic surveillance system if the system uses this type of compression to store the video signal before it is processed by other systems.
In this case, the subsequent analysis of the video signal could be on compromised data showing what the attacker wants to show.

While we do not specifically attack their models, viable compression schemes based on deep neural networks have already been proposed in the literature, showing promising results~\cite{toderici2015variable,toderici2016full}.

\begin{figure}
\begin{center}
\includegraphics[scale=0.7]{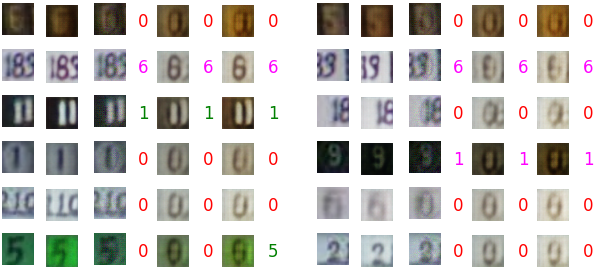}
\end{center}
\caption{Results for the $L_2$ optimization latent attack (see Section~\ref{sec:attack-direct}) on the VAE-GAN, targeting a specific image from the class $0$.
Shown are the first 12 non-zero images from the test SVHN data set.
The columns are, in order: the original image, the reconstruction of the original image, the adversarial example, the predicted class of the adversarial example, the reconstruction of the adversarial example, the predicted class of the reconstructed adversarial example, the reconstruction of the reconstructed adversarial example (see Section~\ref{sec:measuring-effectiveness}), and the predicted class of that reconstruction.}
\label{fig:svhn-baseline-targeted-reconstructions-0}
\end{figure}

\subsection{Defining adversarial examples against generative models}
\label{sec:problem-definition-adversarial}

We make the following assumptions about generating adversarial examples on a target generative model, $\Gtarg(\vx) = \fdec(\fenc(\vx))$.
$\Gtarg$ is trained on inputs $\calX$ that can naturally be labeled with semantically meaningful classes $\calY$, although there may be no such labels at training time, or the labels may not have been used during training.
$\Gtarg$ normally succeeds at generating an output $\vxh = \Gtarg(\vx)$ in class $y$ when presented with an input $\vx$ from class $y$.
In other words, whatever target output class the attacker is interested in, we assume that $\Gtarg$ successfully captures it in the latent representation such that it can generate examples of that class from the decoder.
This target output class does not need to be from the most salient classes in the training dataset.
For example, on models trained on MNIST, the attacker may not care about generating different target digits (which are the most salient classes).
The attacker may prefer to generate the same input digits in a different style (perhaps to aid forgery).
We also assume that the attacker has access to $\Gtarg$.
Finally, the attacker has access to a set of examples from the same distribution as $\calX$ that have the target label $y_t$ the attacker wants to generate.
This does not mean that the attacker needs access to the labeled training dataset (which may not exist), or to an appropriate labeled dataset with large numbers of examples labeled for each class $y \in \calY$ (which may be hard or expensive to collect).
The attacks described here may be successful with only a small amount of data labeled for a single target class of interest.

One way to generate such adversaries is by solving the optimization problem $\argmin_{\vxstar} L(\vx,\vxstar)\ s.t.\ \Oracle(\Gtarg(\vxstar)) = y_t$, where $\Oracle$ reliably discriminates between inputs of class $y_t$ and inputs of other classes.
In practice, a classifier trained by the attacker may server as $\Oracle$.
Other types of adversaries from Section~\ref{sec:related-work-adversarial} can also be used to approximate this optimization in natural ways, some of which we describe in Section~\ref{sec:attack-methodology}.


If the attacker only needs to generate one successful attack, the problem of determining if an attack is successful can be solved by manually reviewing the $\vxstar$ and $\vxhadv$ pairs and choosing whichever the attacker considers best.
However, if the attacker wants to generate many successful attacks, an automated method of evaluating the success of an attack is necessary.
We show in Section~\ref{sec:measuring-effectiveness} how to measure the effectiveness of an attack automatically using a classifier trained on $\vz = \fenc(\vx)$.

%% file: attack-methodology.tex
\section{Attack methodology}
\label{sec:attack-methodology}


The attacker would like to construct an adversarially-perturbed input to influence the latent representation in a way that will cause the reconstruction process to reconstruct an output for a different class.
We propose three approaches to attacking generative models: a classifier-based attack, where we train a new classifier on top of the latent space $\vz$ and use that classifier to find adversarial examples in the latent space; an attack using $\calL_{\VAE}$ to target the output directly; and an attack on the latent space, $\vz$.
All three methods are technically applicable to any generative architecture that relies on a learned latent representation $\vz$.
Without loss of generality, we focus on the VAE-GAN architecture.

\subsection{Classifier attack}
\label{sec:attack-indirect}

\begin{figure}[t]
\begin{center}
\includegraphics[scale=1.0]{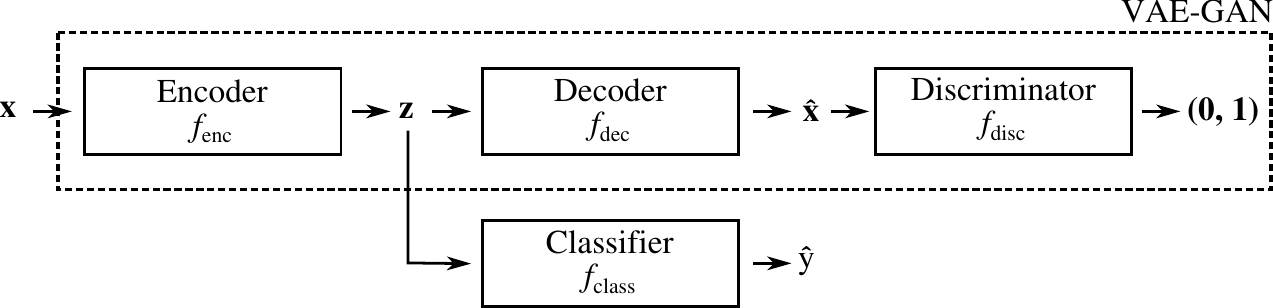}
\end{center}
\caption{The VAE-GAN classifier architecture used to generate classifier-based adversarial examples on the VAE-GAN.
The VAE-GAN in the dashed box is the target network and is frozen while training the classifier.
The path $\vx \rightarrow \fenc \rightarrow \vz \rightarrow \fclass \rightarrow \yh$ is used to generate adversarial examples in $\vz$, which can then be reconstructed by $\fdec$.}
\label{fig:attack-architecture-vae-gan}
\end{figure}

By adding a classifier $\fclass$ to the pre-trained generative model\footnote{
 This is similar to the process of semi-supervised learning in~\cite{kingma2014semi}, although the goal is different.
}, we can turn the problem of generating adversaries for generative models back into the previously solved problem of generating adversarial examples for classifiers.
This approach allows us to apply all of the existing attacks on classifiers in the literature.
However, as discussed below, using this classifier tends to produce lower-quality reconstructions from the adversarial examples than the other two attacks due to the inaccuracies of the classifier.

\paragraph{Step 1.} The weights of the target generative model are frozen, and a new classifier $\fclass(\vz) \rightarrow \yh$ is trained on top of $\fenc$ using a standard classification loss $\calL_{\classifier}$ such as cross-entropy, as shown in Figure~\ref{fig:attack-architecture-vae-gan}.
This process is independent of how the original model is trained, but it requires a training corpus pulled from approximately the same input distribution as was used to train $\Gtarg$, with ground truth labels for at least two classes: $y_t$ and $y_{\ttilde}$, the negative class.

\paragraph{Step 2.} With the trained classifier, the attacker finds adversarial examples $\vxstar$ using the methods described in Section~\ref{sec:methods-adversarial}.

Using $\fclass$ to generate adversarial examples does not always result in high-quality reconstructions, as can be seen in the middle column of Figure~\ref{fig:direct-targeted-reconstructions-0} and in Figure~\ref{fig:classifier-targeted-reconstructions-4}.
This appears to be due to the fact that $\fclass$ adds additional noise to the process.
For example, $\fclass$ sometimes confidently misclassifies latent vectors $\vz$ that represent inputs that are far from the training data distribution, resulting in $\fdec$ failing to reconstruct a plausible output from the adversarial example.

\subsection{$\calL_{\VAE}$ attack}
\label{sec:attack-direct-vae-loss}

Our second approach generates adversarial perturbations using the VAE loss function.
The attacker chooses two inputs, $\vx_s$ (the source) and $\vx_t$ (the target), and uses one of the standard adversarial methods to perturb $\vx_s$ into $\vxstar$ such that its reconstruction $\vxhstar$ matches the reconstruction of $\vx_t$, using the methods described in Section~\ref{sec:methods-adversarial}.

The adversary precomputes the reconstruction $\vxh_t$ by evaluating $\fdec(\fenc(\vx_t))$ once before performing optimization.
In order to use $\calL_{\VAE}$ in an attack, the second term (the reconstruction loss) of $\calL_{\VAE}$ (see Equation~\ref{eqn:vae-loss}) is changed so that instead of computing the reconstruction loss between $\vx$ and $\vxh$, the loss is computed between $\vxhstar$ and $\vxh_t$.
This means that during each optimization iteration, the adversary needs to compute $\vxhstar$, which requires the full $\fdec(\fenc(\vxstar))$ to be evaluated.

\subsection{Latent attack}
\label{sec:attack-direct}

Our third approach attacks the latent space of the generative model.

\paragraph{Single latent vector target.}
This attack is similar to the work of \citet{sabour2015adv}, in which they use a pair of source image $\vx_s$ and target image $\vx_t$ to generate $\vxstar$ that induces the target network to produce similar activations at some hidden layer $l$ as are produced by $\vx_t$, while maintaining similarity between $\vx_s$ and $\vxstar$.

For this attack to work on latent generative models, it is sufficient to compute $\vz_t = \fenc(\vx_t)$ and then use the following loss function to generate adversarial examples from different source images $\vx_s$, using the methods described in Section~\ref{sec:methods-adversarial}:
\begin{equation}
\label{eqn:direct-attack-loss}
 \calL_{\latent} = L(\vz_t, \fenc(\vxstar)) \mathrm{.}
\end{equation}
$L(\cdot)$ is a distance measure between two vectors.
We use the $L_2$ norm, under the assumption that the latent space is approximately euclidean.

We also explored a variation on the single latent vector target attack, which we describe in Section~\ref{sec:attack-mean-latent-vector} in the Appendix.

\subsection{Methods for solving the adversarial optimization problem}
\label{sec:methods-adversarial}

We can use a number of different methods to generate the adversarial examples.
We initially evaluated both the fast gradient sign~\cite{goodfellow2014explaining} method and an $L_2$ optimization method.
As the latter produces much better results we focus on the $L_2$ optimization method, while we include some FGS results in the Appendix.
The attack can be used either in targeted mode (where we want a specific class, $y_t$, to be reconstructed) or untargeted mode (where we just want an incorrect class to be reconstructed).
In this paper, we focus on the targeted mode of the attacks.

 %

\paragraph{$L_2$ optimization.}
The optimization-based approach, explored in~\citet{szegedy2013intriguing} and \citet{carlini2016towards}, poses the adversarial generation problem as the following optimization problem:
\begin{equation}
\label{eqn:optimization}
\argmin_{\vxstar} \lambda L(\vx, \vxstar) + \calL(\vxstar, y_t) \mathrm{.}
\end{equation}
As above, $L(\cdot)$ is a distance measure, and $\calL$ is one of $\calL_{\classifier}$, $\calL_{\VAE}$, or $\calL_{\latent}$.
The constant $\lambda$ is used to balance the two loss contributions.
For the $\calL_{\VAE}$ attack, the optimizer must do a full reconstruction at each step of the optimizer.
The other two attacks do not need to do reconstructions while the optimizer is running, so they generate adversarial examples much more quickly, as shown in Table~\ref{tab:adversarial-example-distances}.

%% file: measuring-effectiveness.tex
\subsection{Measuring attack effectiveness}
\label{sec:measuring-effectiveness}

To generate a large number of adversarial examples automatically against a generative model, the attacker needs a way to judge the quality of the adversarial examples.
We leverage $\fclass$ to estimate whether a particular attack was successful.\footnote{
 Note that $\fclass$ here is being used in a different manner than when we use it to generate adversarial examples.
 However, the network itself is identical, so we don't distinguish between the two uses in the notation.
}

\paragraph{Reconstruction feedback loop.}
The architecture is the same as shown in Figure~\ref{fig:attack-architecture-vae-gan}.
We use the generative model to reconstruct the attempted adversarial inputs $\vxstar$ by computing:
\begin{equation}
  \vxhstar = \fdec(\fenc(\vxstar)) \mathrm{.}
\end{equation}
Then, $\fclass$ is used to compute:
\begin{equation}
  \yh = \fclass(\fenc(\vxhstar)) \mathrm{.}
\end{equation}
The input adversarial examples $\vxstar$ are not classified directly, but are first fed to the generative model for reconstruction.
This reconstruction loop improves the accuracy of the classifier by $60\%$ on average against the adversarial attacks we examined.
The predicted label $\yh$ after the reconstruction feedback loop is compared with the attack target $y_t$ to determine if the adversarial example successfully reconstructed to the target class.
If the precision and recall of $\fclass$ are sufficiently high on $y_t$, $\fclass$ can be used to filter out most of the failed adversarial examples while keeping most of the good ones.

We derive two metrics from classifier predictions after one reconstruction feedback loop.
The first metric is $\ASuntargeted$, the attack success rate ignoring targeting, i.e., without requiring the output class of the adversarial example to match the target class:
\begin{equation}
  \ASuntargeted = \frac{1}{N} \sum_{i=1}^{N} \textbf{1}_{\yh^i \ne y^i}
\end{equation}
$N$ is the total number of reconstructed adversarial examples; $\textbf{1}_{\yh^i \ne y^i}$ is $1$ when $\yh^i$, the classification of the reconstruction for image $i$, does not equal $y^i$, the ground truth classification of the original image, and $0$ otherwise.
The second metric is $\AStargeted$, the attack success rate including targeting (i.e., requiring the output class of the adversarial example to match the target class), which we define similarly as:
\begin{equation}
  \AStargeted = \frac{1}{N} \sum_{i=1}^{N} \textbf{1}_{\yh^i = y_t^i} \mathrm{.}
\end{equation}
Both metrics are expected to be higher for more successful attacks.
Note that $\AStargeted \le \ASuntargeted$.
When computing these metrics, we exclude input examples that have the same ground truth class as the target class.

%% file: evaluation.tex
\section{Evaluation}
\label{sec:evaluation}

We evaluate the three attacks on MNIST~\citep{mnist}, SVHN~\citep{svhn} and CelebA~\citep{celeba}, using the standard training and validation set splits.
The VAE and VAE-GAN architectures are implemented in TensorFlow~\citep{tensorflow2015-whitepaper}.
We optimized using Adam with learning rate $0.001$ and other parameters set to default values for both the generative model and the classifier.
For the VAE, we use two architectures: a simple architecture with a single fully-connected hidden layer with 512 units and ReLU activation function; and a convolutional architecture taken from the original VAE-GAN paper \citet{larsen2015autoencoding} (but trained with only the VAE loss).
We use the same architecture trained with the additional GAN loss for the VAE-GAN model, as described in that work.
For both VAE and VAE-GAN we use a 50-dimensional latent representation on MNIST, a 1024-dimensional latent representation on SVHN and 2048-dimensional latent representation on CelebA.

In this section we only show results where no sampling from latent space has been performed.
Instead we use the mean vector $\vmu$ as the latent representation $\vz$.
As sampling can have an effect on the resulting reconstructions, we evaluated it separately.
We show the results with different number of samples in Figure~\ref{fig:faces-sampling} in the Appendix.
On most examples, the visible change is small and in general the attack is still successful.

\subsection{MNIST}

Both VAE and VAE-GAN by themselves reconstruct the original inputs well as show in Figure~\ref{fig:original-reconstructions}, although the quality from the VAE-GAN is noticeably better.
As a control, we also generate random noise of the same magnitude as used for the adversarial examples (see Figure~\ref{fig:random-noise-reconstructions}), to show that random noise does not cause the reconstructed noisy images to change in any significant way.
Although we ran experiments on both VAEs and VAE-GANs, we only show results for the VAE-GAN as it generates much higher quality reconstructions than the corresponding VAE.

\subsubsection{Classifier attack}
\label{sec:evaluation-indirect}

We use a simple classifier architecture to help generate attacks on the VAE and VAE-GAN models.
The classifier consists of two fully-connected hidden layers with 512 units each, using the ReLU activation function.
The output layer is a 10 dimensional softmax.
The input to the classifier is the 50 dimensional latent representation produced by the VAE/VAE-GAN encoder.
The classifier achieves $98.05\%$ accuracy on the validation set after training for 100 epochs.

To see if there are differences between classes, we generate targeted adversarial examples for each MNIST class and present the results per-class.
For the targeted attacks we used the optimization method with lambda $0.001$, where Adam-based optimization was performed for $1000$ epochs with a learning rate of $0.1$.
The mean $L_2$ norm of the difference between original images and generated adversarial examples using the classifier attack is $3.36$, while the mean RMSD is $0.120$.

\begin{figure}
\begin{center}
\includegraphics[scale=0.5]{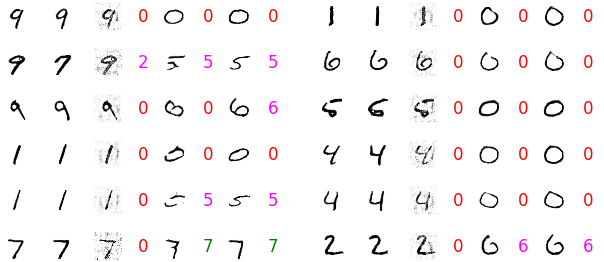}
\end{center}
\caption{Results for the $L_2$ optimization latent attack on the VAE-GAN, targeting the mean latent vector for $0$.
Shown are the first 12 non-zero images from the test MNIST data set.
The columns are, in order: the original image, the reconstruction of the original image, the adversarial example, the predicted class of the adversarial example, the reconstruction of the adversarial example, the predicted class of the reconstructed adversarial example, the reconstruction of the reconstructed adversarial example (see Section~\ref{sec:measuring-effectiveness}), and the predicted class of that reconstruction.}
\label{fig:baseline-targeted-reconstructions-0}
\end{figure}

\begin{figure}
\begin{center}
\includegraphics[scale=0.3]{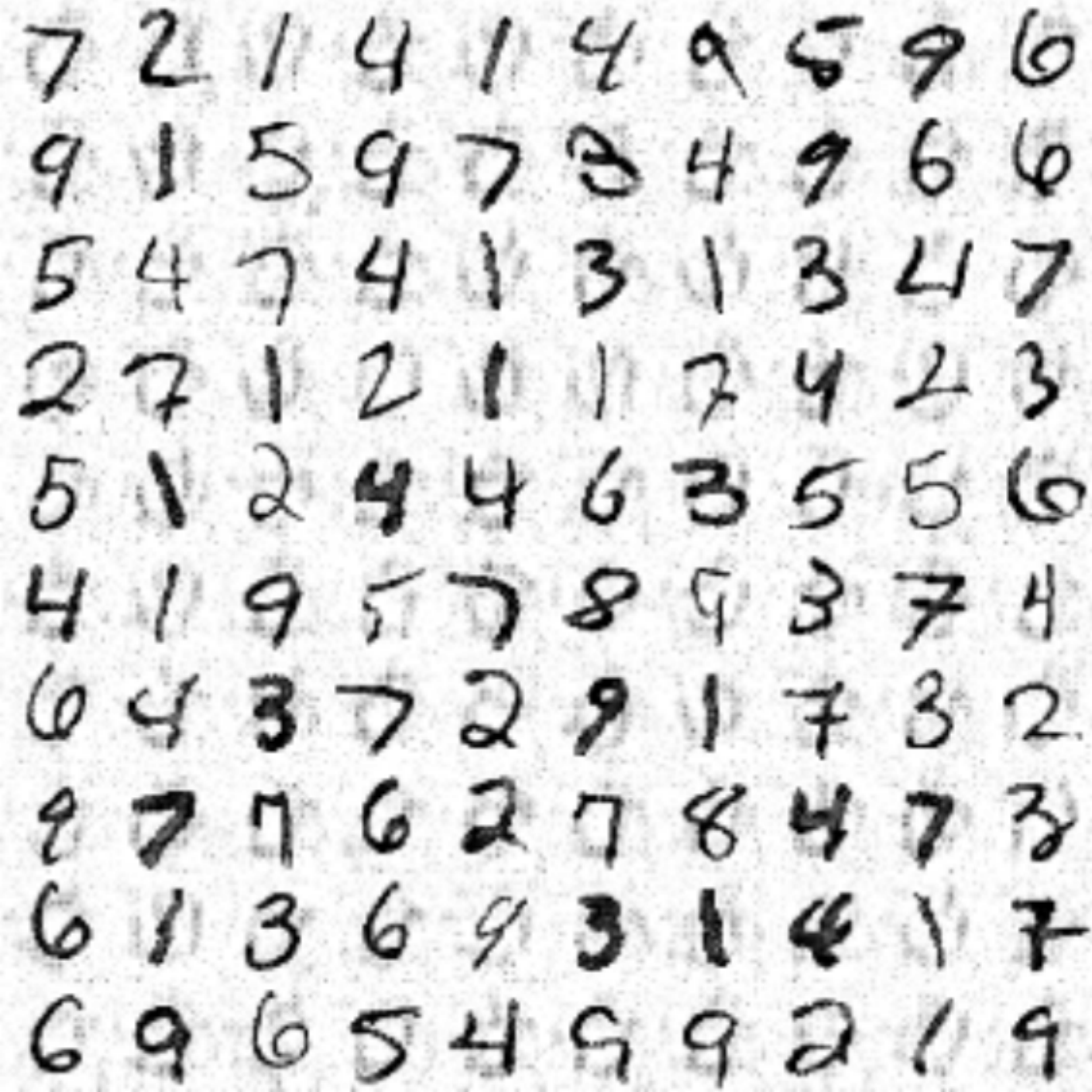}
\rule{0.4pt}{4.3cm}
\includegraphics[scale=0.3]{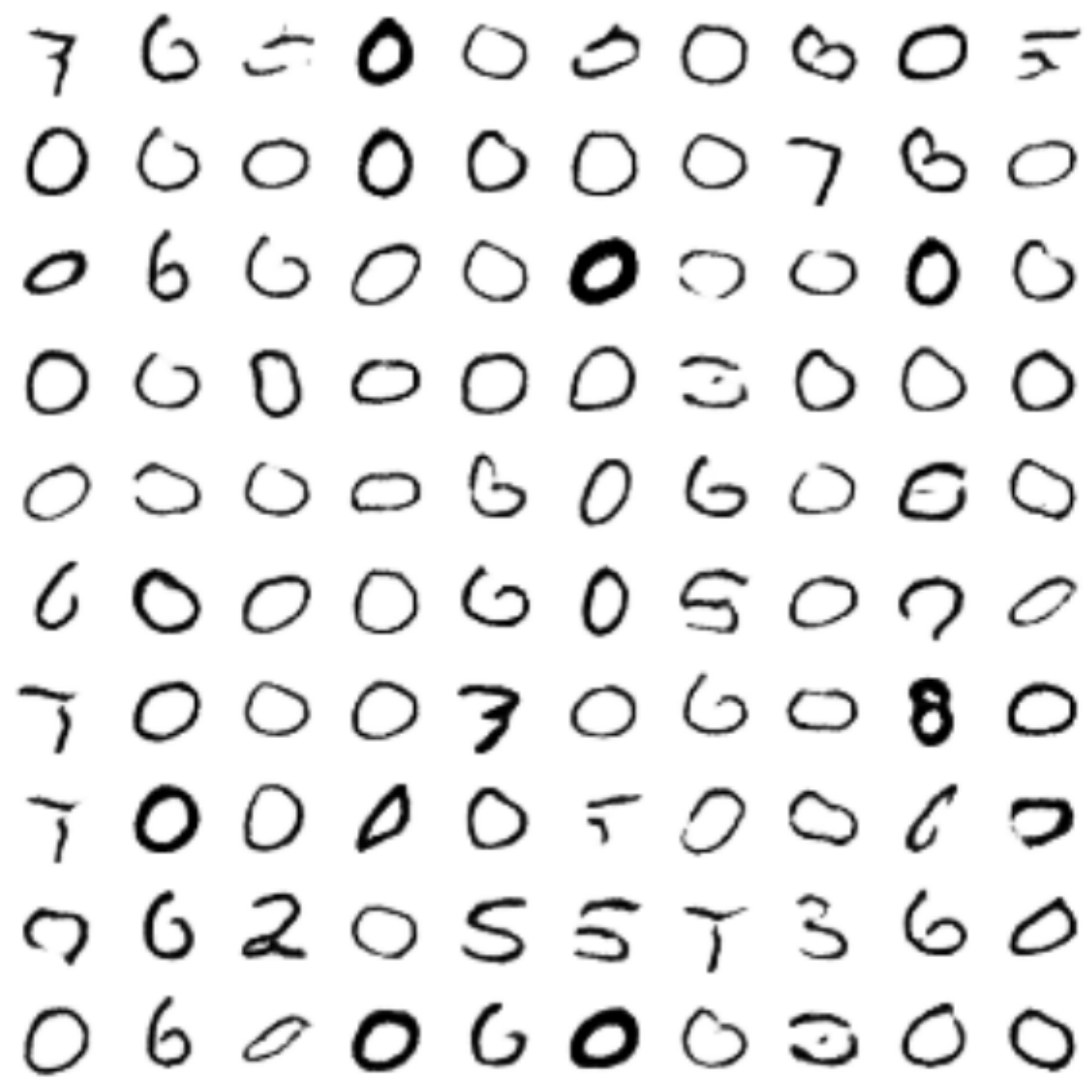}
\rule{0.4pt}{4.3cm}
\includegraphics[scale=0.3]{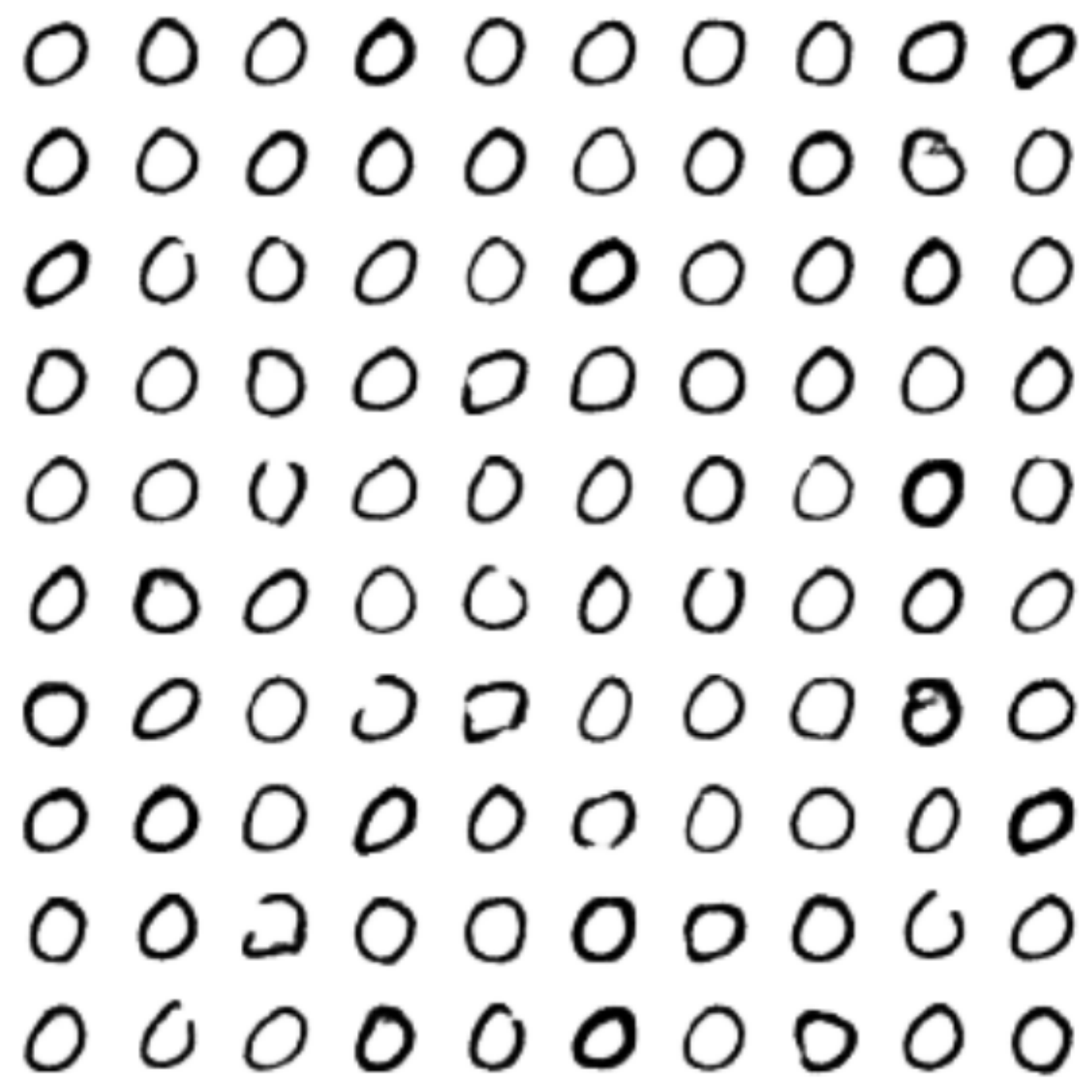}
\end{center}
\caption{%
\textit{Left:} representative adversarial examples with a target class of $0$ on the first $100$ non-zero images from the MNIST validation set.
These were produced using the $L_2$ optimization latent attack (Section~\ref{sec:attack-direct}).
\textit{Middle:} VAE-GAN reconstructions from adversarial examples produced using the $L_2$ optimization classifier attack on the same set of $100$ validation images (those adversaries are not shown, but are qualitatively similiar, see Section~\ref{sec:attack-indirect}).
\textit{Right:} VAE-GAN reconstructions from the adversarial examples in the left column.
Many of the classifier adversarial examples fail to reconstruct as zeros, whereas almost every adversarial example from the latent attack reconstructs as zero.
}
\label{fig:direct-targeted-reconstructions-0}
\end{figure}

Numerical results in Table~\ref{tab:baseline-targeted-results} show that the targeted classifier attack successfully fools the classifier.
Classifier accuracy is reduced to $0\%$, while the matching rate (the ratio between the number of predictions matching the target class and the number of incorrectly classified images) is $100\%$, which means that all incorrect predictions match the target class.
However, what we are interested in (as per the attack definition from Section~\ref{sec:problem-definition-adversarial}) is how the generative model reconstructs the adversarial examples.
If we look at the images generated by the VAE-GAN for class $0$, shown in Figure~\ref{fig:baseline-targeted-reconstructions-0}, the targeted attack is successful on some reconstructed images (e.g. one, four, five, six and nine are reconstructed as zeroes).
But even when the classifier accuracy is $0\%$ and matching rate is $100\%$, an incorrect classification does not always result in a reconstruction to the target class, which shows that the classifier is fooled by an adversarial example more easily than the generative model.

\paragraph{Reconstruction feedback loop.}
The reconstruction feedback loop described in Section~\ref{sec:measuring-effectiveness} can be used to measure how well a targeted attack succeeds in making the generative model change the reconstructed classes.
Table~\ref{tab:indirect-distribution-classification-accuracies} in the Appendix shows $\ASuntargeted$ and $\AStargeted$ for all source and target class pairs.
A higher value signifies a more successful attack for that pair of classes.
It is interesting to observe that attacking some source/target pairs is much easier than others (e.g. pair $(4,0)$ vs. $(0,8)$) and that the results are not symmetric over source/target pairs.
Also, some pairs do well in $\ASuntargeted$, but do poorly in $\AStargeted$ (e.g., all source digits when targeting $4$).
As can be seen in Figure~\ref{fig:classifier-targeted-reconstructions-4}, the classifier adversarial examples targeting $4$ consistently fail to reconstruct into something easily recognizable as a $4$.
Most of the reconstructions look like $5$, but the adversarial example reconstructions of source $5$s instead look like $0$ or $3$.

\subsubsection{$\calL_{\VAE}$ attack}

For generating adversarial examples using the $\calL_{\VAE}$ attack, we used the optimization method with $\lambda = 1.0$, where Adam-based optimization was performed for $1000$ epochs with a learning rate of $0.1$.
The mean $L_2$ norm of the difference between original images and generated adversarial examples with this approach is $3.68$, while the mean RMSD is $0.131$.

We show $\ASuntargeted$ and $\AStargeted$ of the $\calL_{\VAE}$ attack in Table~\ref{tab:direct-lvae-distribution-classification-accuracies} in the Appendix.
Comparing with the numerical evaluation results of the latent attack (below), we can see that both methods achieve similar results on MNIST.

\subsubsection{Latent attack}

To generate adversarial examples using the latent attack, we used the optimization method with $\lambda = 1.0$, where Adam-based optimization was performed for $1000$ epochs with a learning rate of $0.1$.
The mean $L_2$ norm of the difference between original images and generated adversarial examples using this approach is $2.96$, while the mean RMSD is $0.105$.

Table~\ref{tab:direct-targeted-random-numerical} shows $\ASuntargeted$ and $\AStargeted$ for all source and target class pairs.
Comparing with the numerical evaluation results of the classifier attack we can see that the latent attack performs much better.
This result remains true when visually comparing the reconstructed images, shown in Figure~\ref{fig:direct-targeted-reconstructions-0}.

We also tried an untargeted version of the latent attack, where we change Equation~\ref{eqn:direct-attack-loss} to maximize the distance in latent space between the encoding of the original image and the encoding of the adversarial example.
In this case the loss we are trying to minimize is unbounded, since the $L_2$ distance can always grow larger, so the attack normally fails to generate a reasonable adversarial example.

Additionally, we also experimented with targeting latent representations of specific images from the training set instead of taking the mean, as described in Section~\ref{sec:attack-direct}.
We show the numerical results in Table~\ref{tab:direct-targeted-random-numerical} and the generated reconstructions in Figure~\ref{fig:direct-targeted-random-reconstructions} (in the Appendix).
It is also interesting to compare the results with $\calL_{\VAE}$, by choosing the same image as the target.
Results for $\calL_{\VAE}$ for the same target images as in Table~\ref{tab:direct-targeted-random-numerical} are shown in Table~\ref{tab:direct-lvae-targeted-random-numerical} in the Appendix.
The results are identical between the two attacks, which is expected as the target image is the same~-- only the loss function differs between the methods.

\subsection{SVHN}

\begin{figure}
\begin{center}
\includegraphics[scale=0.4]{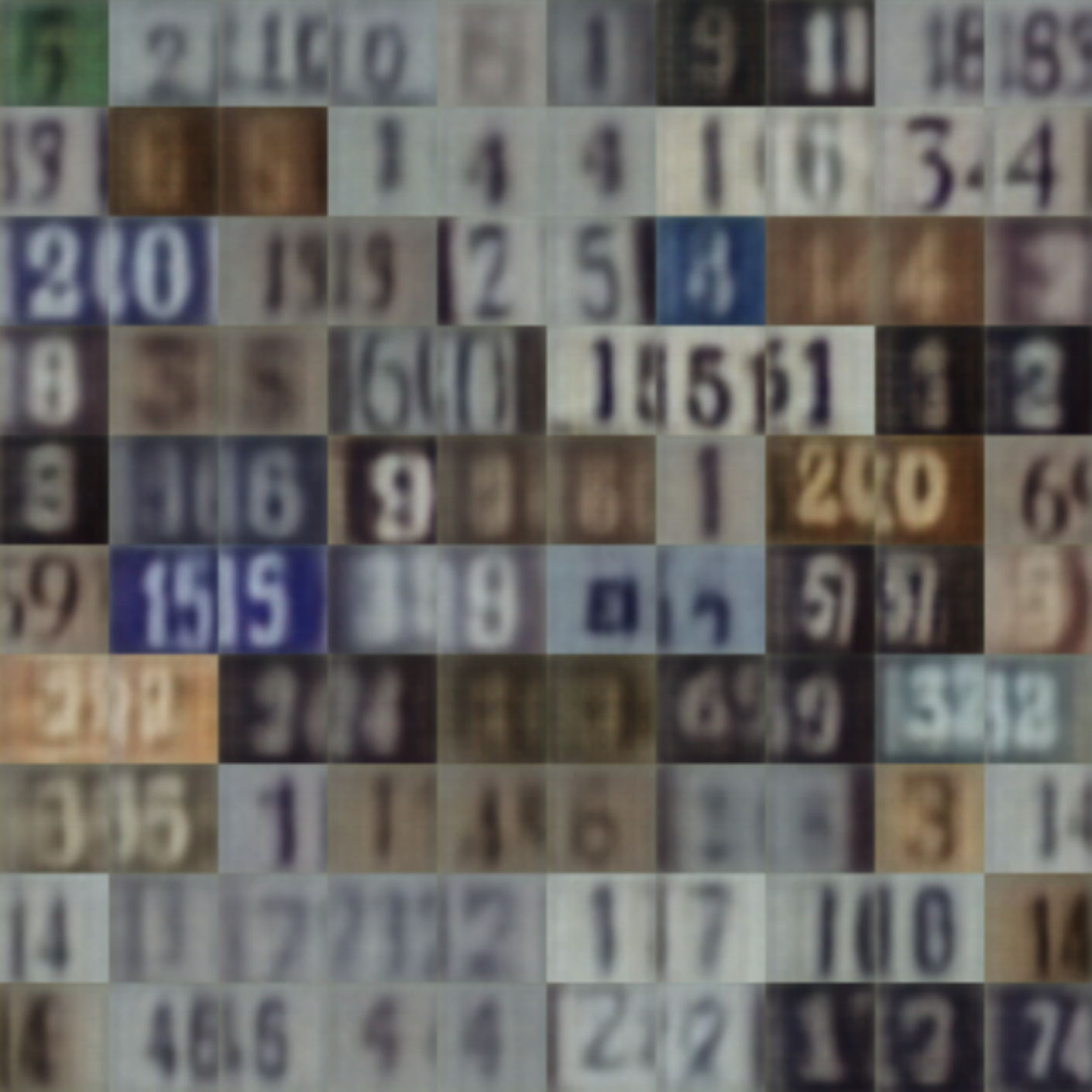}
\rule{0.4pt}{5.7cm}
\includegraphics[scale=0.4]{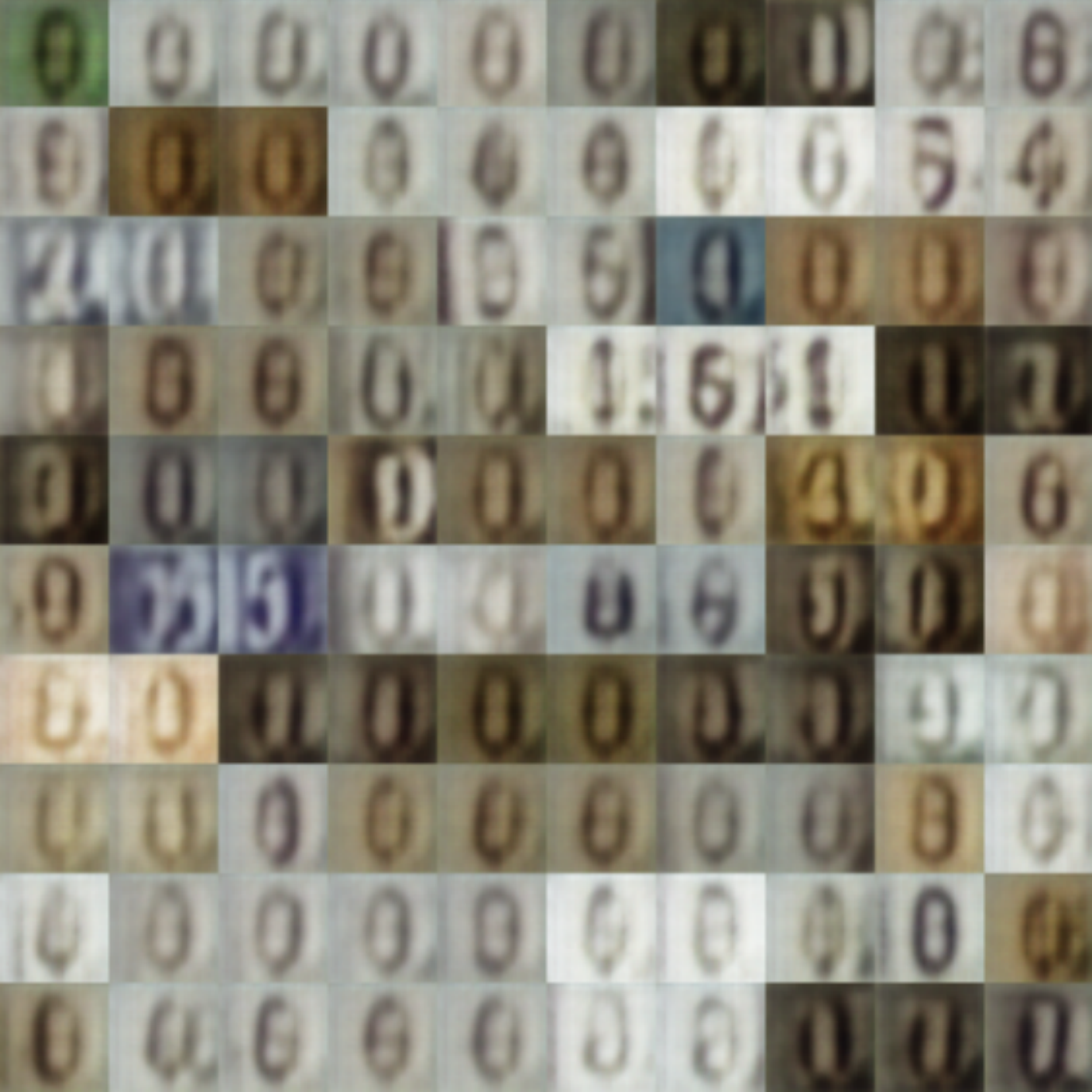}
\end{center}
\caption{%
\textit{Left:} VAE-GAN reconstructions of adversarial examples generated using the $L_2$ optimization $\calL_{\VAE}$ attack (single image target).
\textit{Right:} VAE-GAN reconstructions of adversarial examples generated using the $L_2$ optimization latent attack (single image target).
Approximately $85$ out of $100$ images are convincing zeros for the $L_2$ latent attack, whereas only about $5$ out of $100$ could be mistaken for zeros with the $\calL_{\VAE}$ attack.
}
\label{fig:svhn-latent-vs-lvae}
\end{figure}

The SVHN dataset consists of cropped street number images and is much less clean than MNIST.
Due to the way the images have been processed, each image may contain more than one digit; the target digit is roughly in the center.
VAE-GAN produces high-quality reconstructions of the original images as shown in Figure~\ref{fig:svhn-original-reconstructions} in the Appendix.

For the classifier attack, we set $\lambda = 10^{-5}$ after testing a range of values, although we were unable to find an effective value for this attack against SVHN.
For the latent and $\calL_{\VAE}$ attacks we set $\lambda = 10$.

In Table~\ref{tab:svhn-direct-targeted-random-numerical} we show $\ASuntargeted$ and $\AStargeted$ for the $L_2$ optimization latent attack.
The evaluation metrics are less strong on SVHN than on MNIST, but it is still straightforward for an attacker to find a successful attack for almost all source/target pairs.
Figure~\ref{fig:svhn-baseline-targeted-reconstructions-0} supports this evaluation.
Visual inspection shows that $11$ out of the $12$ adversarial examples reconstructed as $0$, the target digit.
It is worth noting that $2$ out of the $12$ adversarial examples look like zeros (rows $1$ and $11$), and two others look like both the original digit and zero, depending on whether the viewer focuses on the light or dark areas of the image (rows $4$ and $7$).
The $L_2$ optimization latent attack achieves much better results than the $\calL_{\VAE}$ attack (see Table~\ref{tab:svhn-lvae-targeted-random-numerical} and Figure~\ref{fig:svhn-latent-vs-lvae}) on SVHN, while both attacks work equally well on MNIST.

\subsection{CelebA}

The CelebA dataset consists of more than 200,000 cropped faces of celebrities, each annotated with 40 different attributes.
For our experiments, we further scale the images to 64x64 and ignore the attribute annotations.
VAE-GAN reconstructions of original images after training are shown in Figure~\ref{fig:faces-original-reconstructions} in the Appendix.

Since faces don't have natural classes, we only evaluated the latent and $\calL_{\VAE}$ attacks.
We tried lambdas ranging from $0.1$ to $0.75$ for both attacks.
Figure~\ref{fig:faces-latent-single-vector-target} shows adversarial examples generated using the latent attack and a lambda value of $0.5$ ($L_2$ norm between original images and generated adversarial examples $9.78$, RMSD $0.088$) and the corresponding VAE-GAN reconstructions.
Most of the reconstructions reflect the target image very well.
We get even better results with the $\calL_{\VAE}$ attack, using a lambda value of $0.75$ ($L_2$ norm between original images and generated adversarial examples $8.98$, RMSD $0.081$) as shown in Figure~\ref{fig:faces-lvae-single-image-target}.

\begin{figure}[h]
\begin{center}
\raisebox{-0.5\height}{\includegraphics[scale=0.5]{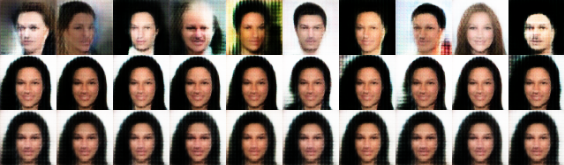}}
~
\raisebox{-0.5\height}{\includegraphics[scale=0.1]{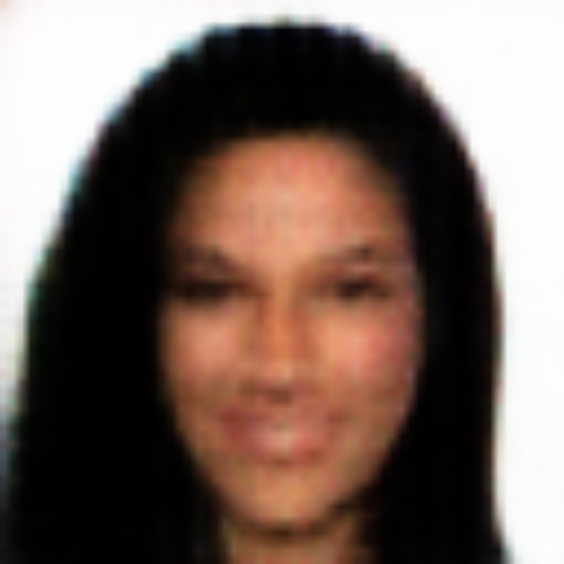}}
\end{center}
\caption{Summary of different attacks on CelebA dataset: reconstructions of original images (top), reconstructions of adversarial examples generated using the latent attack (middle) and $\calL_{\VAE}$ attack (bottom). Target reconstruction is shown on the right. Full results are in the Appendix.}
\label{fig:faces-summary}
\end{figure}

\subsection{Summary of different attack methods}

\begin{table}
{
\tiny
\begin{center}
\begin{tabular}{|c|c|c|c|c|c|c|}
\hline
 & \multicolumn{3}{c|}{\textbf{MNIST}} & \multicolumn{3}{c|}{\textbf{SVHN}} \\
\textbf{Method} & \multicolumn{1}{r}{\textbf{Mean $L_2$}} & \multicolumn{1}{r}{\textbf{Mean RMSD}} & \textbf{Time to attack} & \multicolumn{1}{r}{\textbf{Mean $L_2$}} & \multicolumn{1}{r}{\textbf{Mean RMSD}} & \textbf{Time to attack}  \\
\hline
$L_2$ Optimization Classifier Attack & $3.36$ & $0.120$ & $277$ & $1.77$ & $0.032$ & $274$ \\
\hline
$L_2$ Optimization $\calL_{\VAE}$ Attack & $3.68$ & $0.131$ & $734$ & $2.36$ & $0.043$ & $895$ \\
\hline
$L_2$ Optimization Latent Attack & $2.96$ & $0.105$ & $236$ & $2.80$ & $0.051$ & $242$ \\
\hline
\end{tabular}
\end{center}
}
\caption{Comparison of mean $L_2$ norm and RMSD between the original images and the generated adversarial examples for the different attacks.
\textit{Time to attack} is the mean number of seconds it takes to generate $1000$ adversarial examples using the given attack method (with the same number of optimization iterations for each attack).}
\label{tab:adversarial-example-distances}
\end{table}

Table~\ref{tab:adversarial-example-distances} shows a comparison of the mean distances between original images and generated adversarial examples for the three different attack methods.
The larger the distance between the original image and the adversarial perturbation, the more noticeable the perturbation will tend to be, and the more likely a human observer will no longer recognize the original input, so effective attacks keep these distances small while still achieving their goal.
The latent attack consistently gives the best results in our experiments, and the classifier attack performs the worst.

We also measure the time it takes to generate $1000$ adversarial examples using the given attack method.
The $\calL_{\VAE}$ attack is by far the slowest of the three, due to the fact that it requires computing full reconstructions at each step of the optimizer when generating the adversarial examples.
The other two attacks do not need to run the reconstruction step during optimization of the adversarial examples.

%% file: conclusion.tex
\section{Conclusion}

We explored generating adversarial examples against generative models such as VAEs and VAE-GANs.
These models are also vulnerable to adversaries that convince them to turn inputs into surprisingly different outputs.
We have also motivated why an attacker might want to attack generative models.
Our work adds further support to the hypothesis that adversarial examples are a general phenomenon for current neural network architectures, given our successful application of adversarial attacks to popular generative models.
In this work, we are helping to lay the foundations for understanding how to build more robust networks.
Future work will explore defense and robustification in greater depth as well as attacks on generative models trained using natural image datasets such as CIFAR-10 and ImageNet.

%% file: appendix.tex
\appendix
\section{Appendix}

\subsection{Mean latent vector targeted attack}
\label{sec:attack-mean-latent-vector}

A variant of the single latent vector targeted attack described in Section~\ref{sec:attack-direct}, that was not explored in previous work to our knowledge is to take the mean latent vector of many target images and use that vector as $\vx_t$.
This variant is more flexible, in that the attacker can choose different latent properties to target without needing to find the ideal input.
For example, in MNIST, the attacker may wish to have a particular line thickness or slant in the reconstructed digit, but may not have such an image available.
In that case, by choosing some images of the target class with thinner lines or less slant, and some with thicker lines or more slant, the attacker can find a target latent vector that closely matches the desired properties.

In this case, the attack starts by using $\fenc$ to produce the target latent vector, $\vz_t$, from the chosen target images, $\vx_{(t)}$.
\begin{equation}
  \vz_t = \frac{1}{|\vx_{(t)}|} \sum_{i = 0}^{|\vx_{(t)}|} \fenc(\vx_{(t)}^i) \mathrm{.}
\end{equation}

In this work, we choose to reconstruct ``ideal'' MNIST digits by taking the mean latent vector of all of the training digits of each class, and using those vectors as $\vx_t$.
Given a target class $y_t$, a set of examples $\calX$ and their corresponding ground truth labels $\vy$, we create a subset $\vx_{(t)}$ as follows:
\begin{equation}
  \vx_{(t)} = \left\{ \vx_i | \vx_i \in \calX \wedge y_i = y_t \right\} \mathrm{.}
\end{equation}

\begin{figure}[h]
\begin{center}
\includegraphics[scale=0.4]{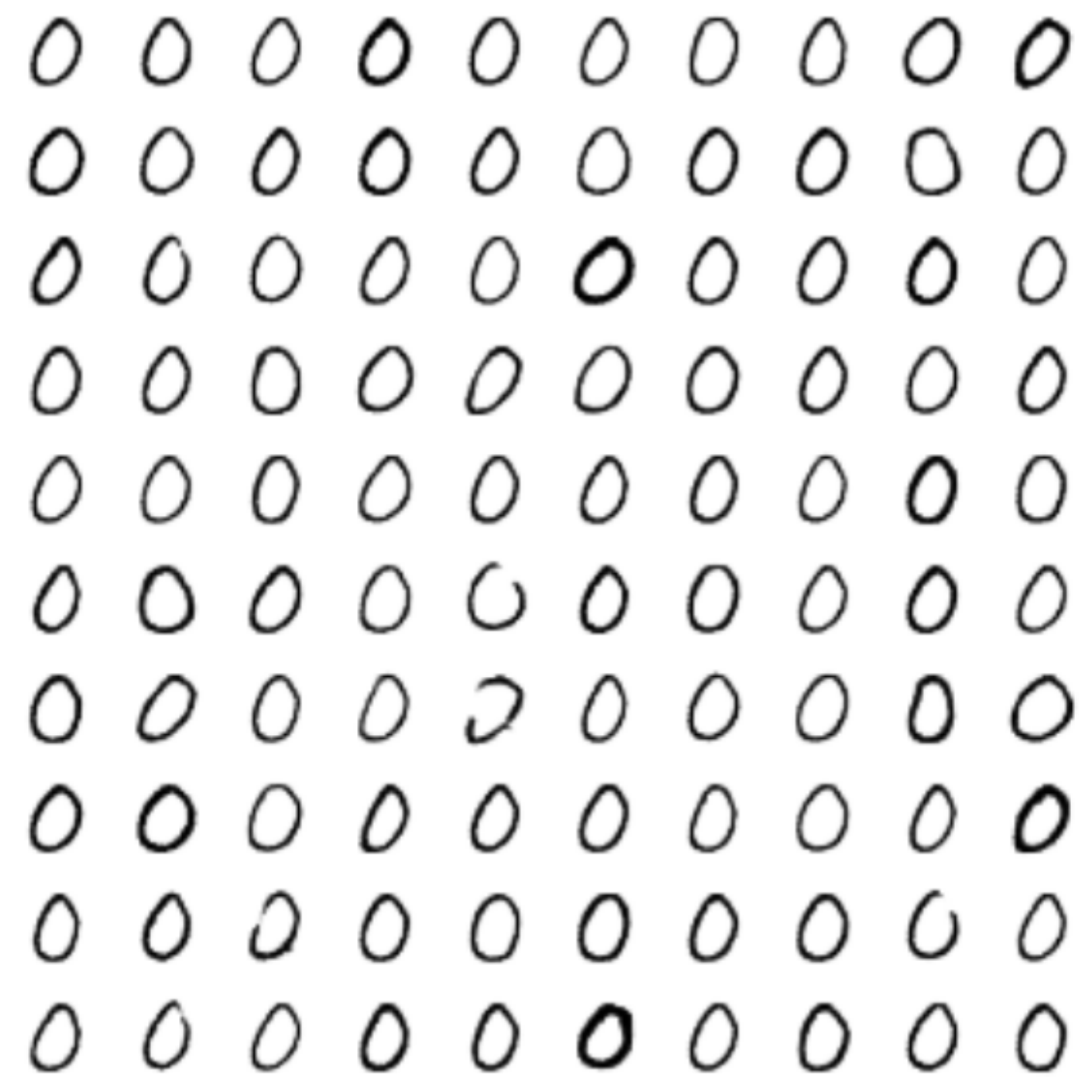}
\includegraphics[scale=0.4]{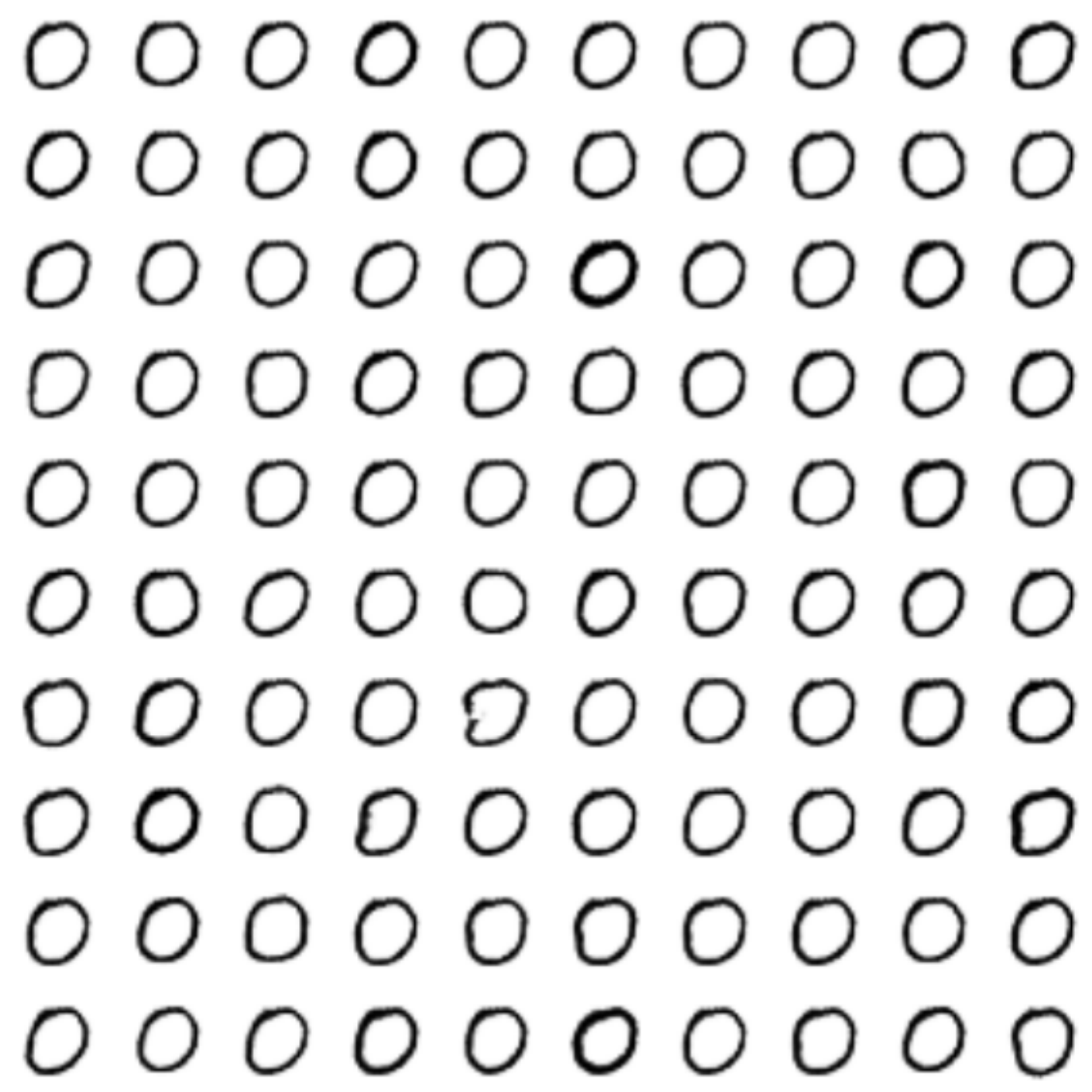} \\
\includegraphics[scale=0.4]{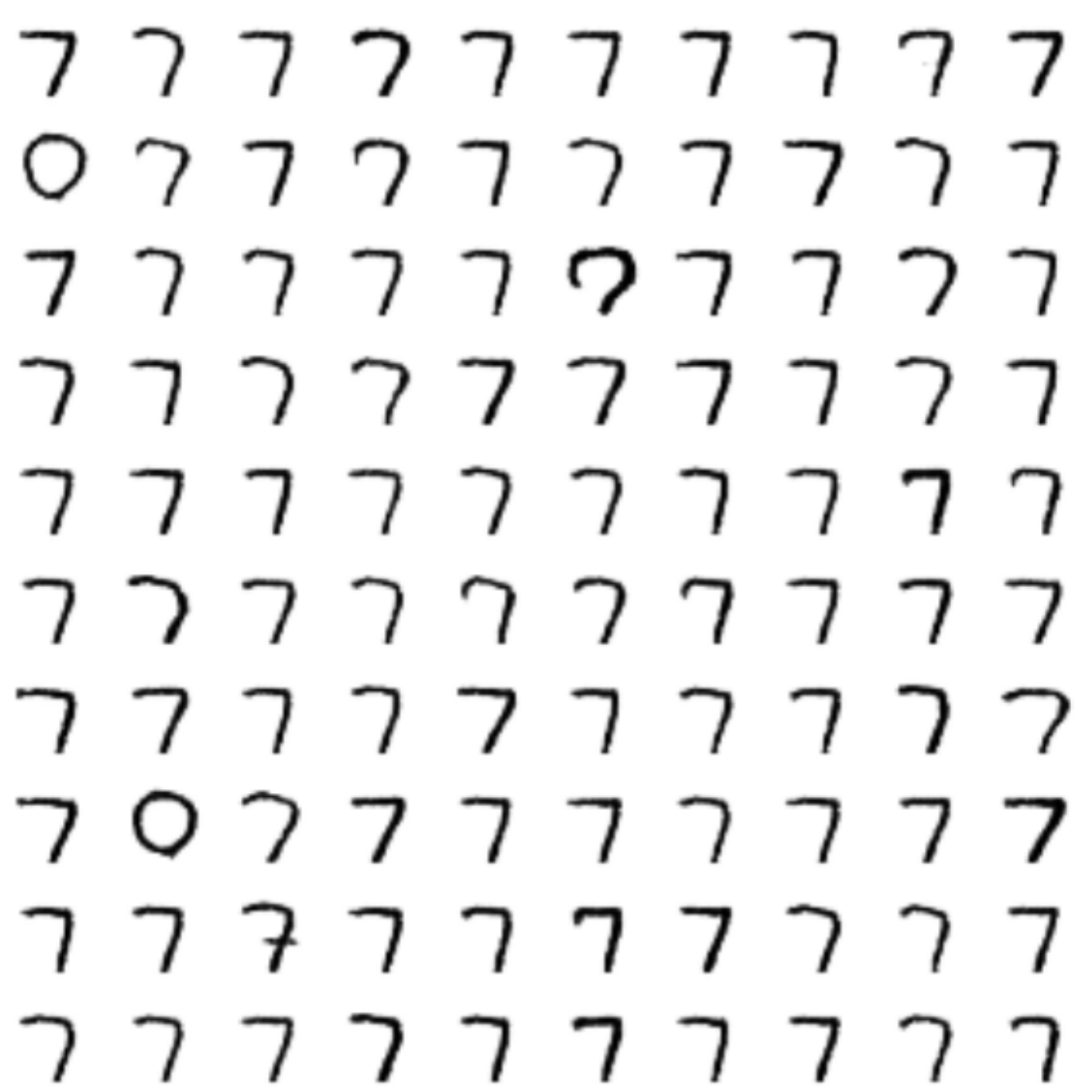}
\includegraphics[scale=0.4]{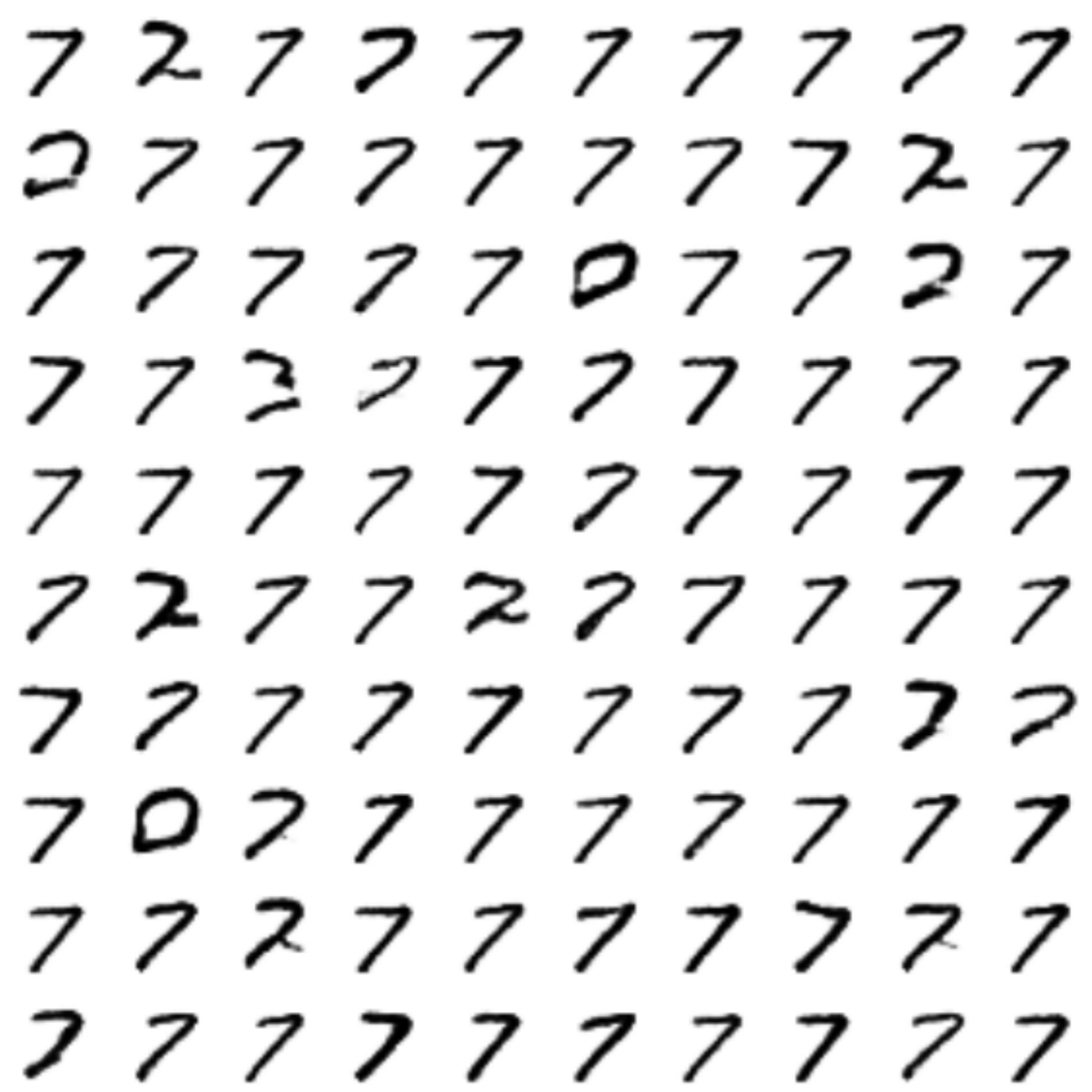}
\end{center}
\caption{\textbf{$L_2$ Optimization Latent Attack (single latent vector target):} VAE-GAN reconstructions of adversarial examples generated using the latent attack with target classes $0$ and $7$ using two random targets in latent space per target class.
Original examples which already belong to the target class are excluded.
The stylistic differences in the reconstructions are clearly visible.}
\label{fig:direct-targeted-random-reconstructions}
\end{figure}

Both variants of this attack appear to be similarly effective, as shown in Figure~\ref{fig:direct-targeted-random-reconstructions} and Figure~\ref{fig:direct-targeted-reconstructions-0}.
The trade-off between the two in these experiments is between the simplicity of the first attack and the flexibility of the second attack.

\subsection{Evaluation results}

\begin{figure}
\begin{center}
\includegraphics[scale=1.0]{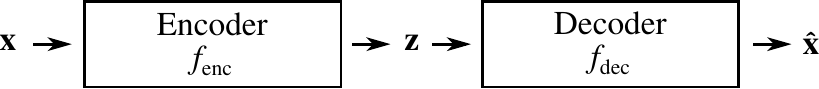}
\end{center}
\caption{Variational autoencoder architecture.}
\label{fig:vae-architecture}
\end{figure}

\begin{figure}[h]
\begin{center}
\includegraphics[scale=0.4]{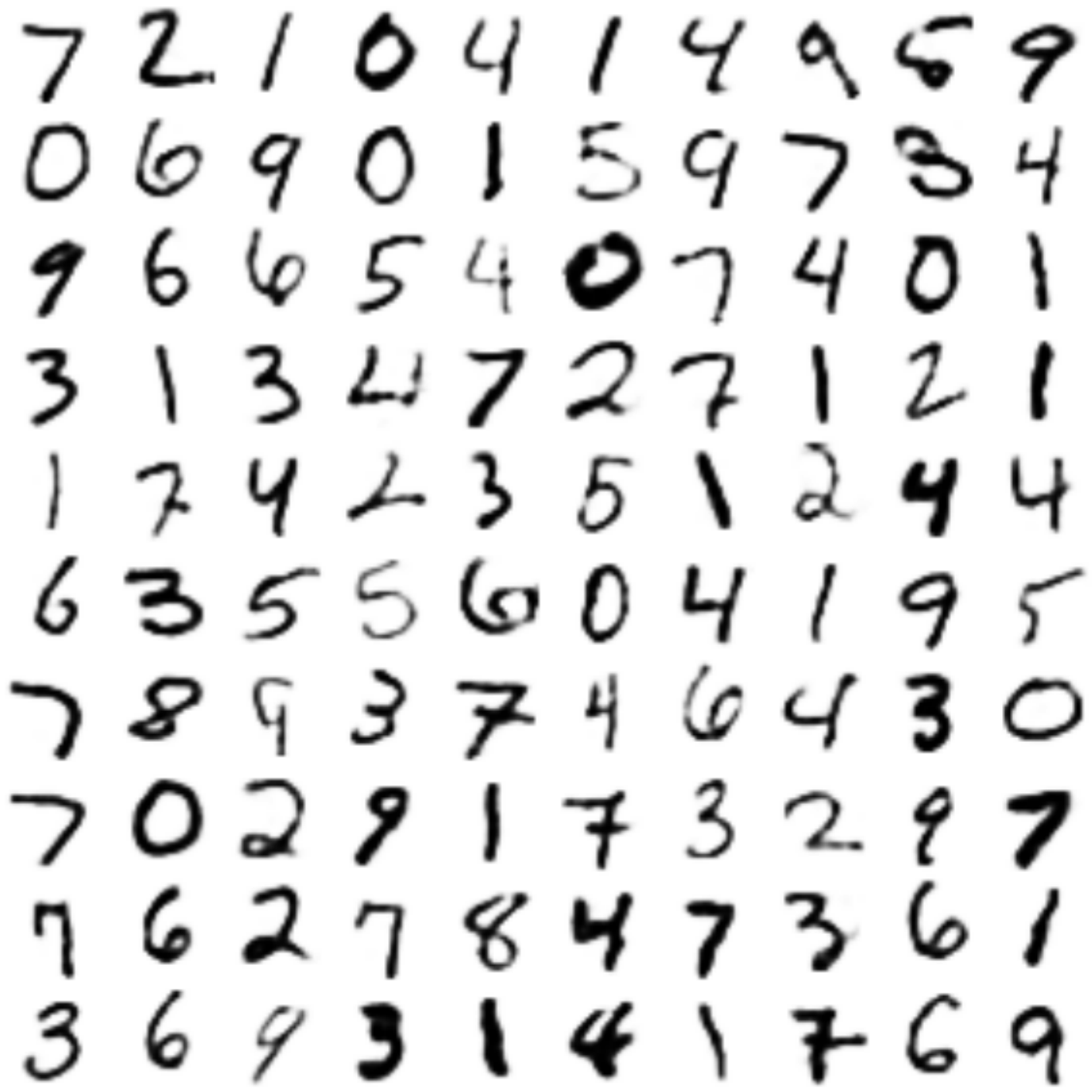}
\rule{0.4pt}{5.7cm}
\includegraphics[scale=0.4]{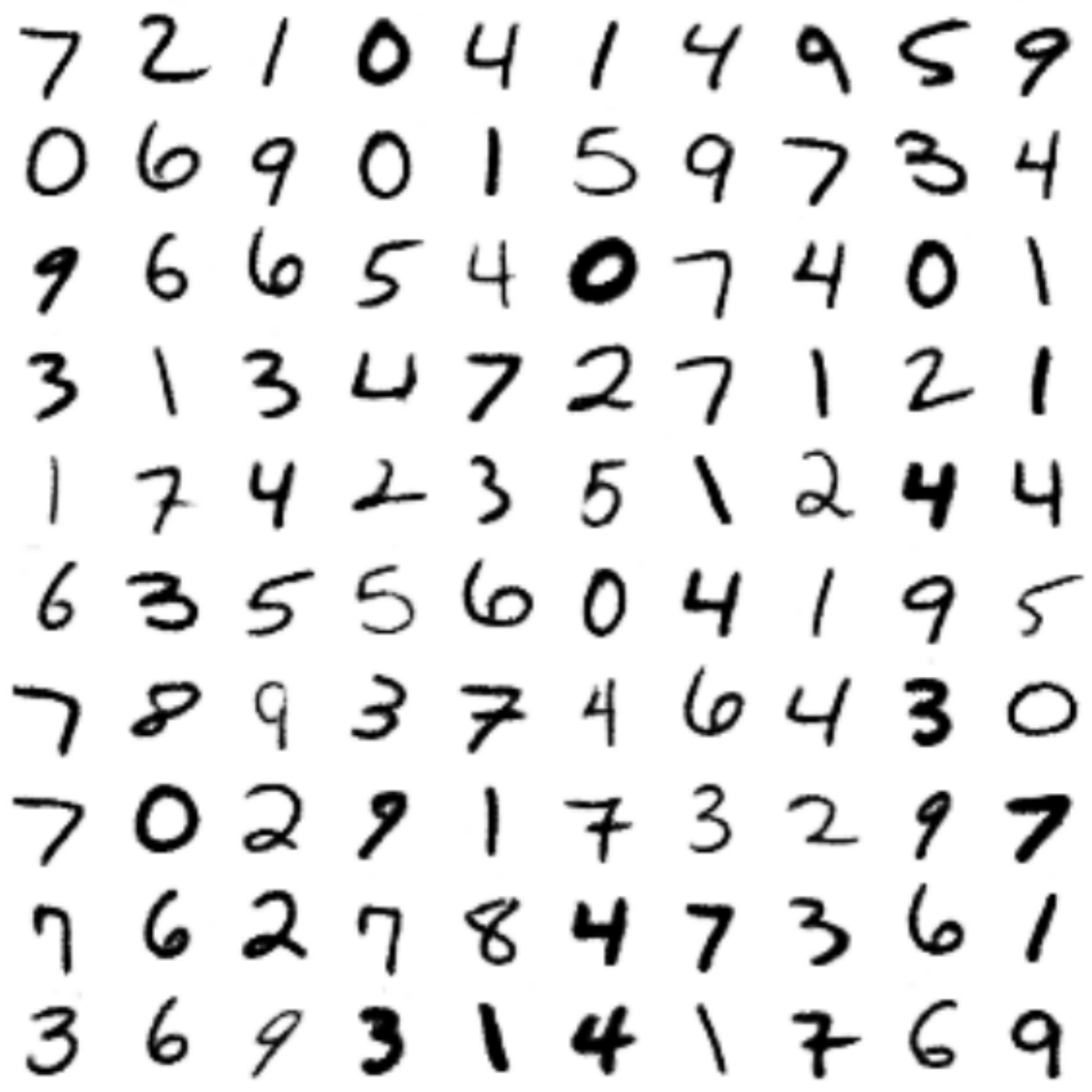}
\end{center}
\caption{\textbf{Original Inputs and Reconstructions:} The first $100$ images from the validation set reconstructed by the VAE (left) and the VAE-GAN (right).}
\label{fig:original-reconstructions}
\end{figure}

\begin{figure}[h]
\begin{center}
\includegraphics[scale=0.4]{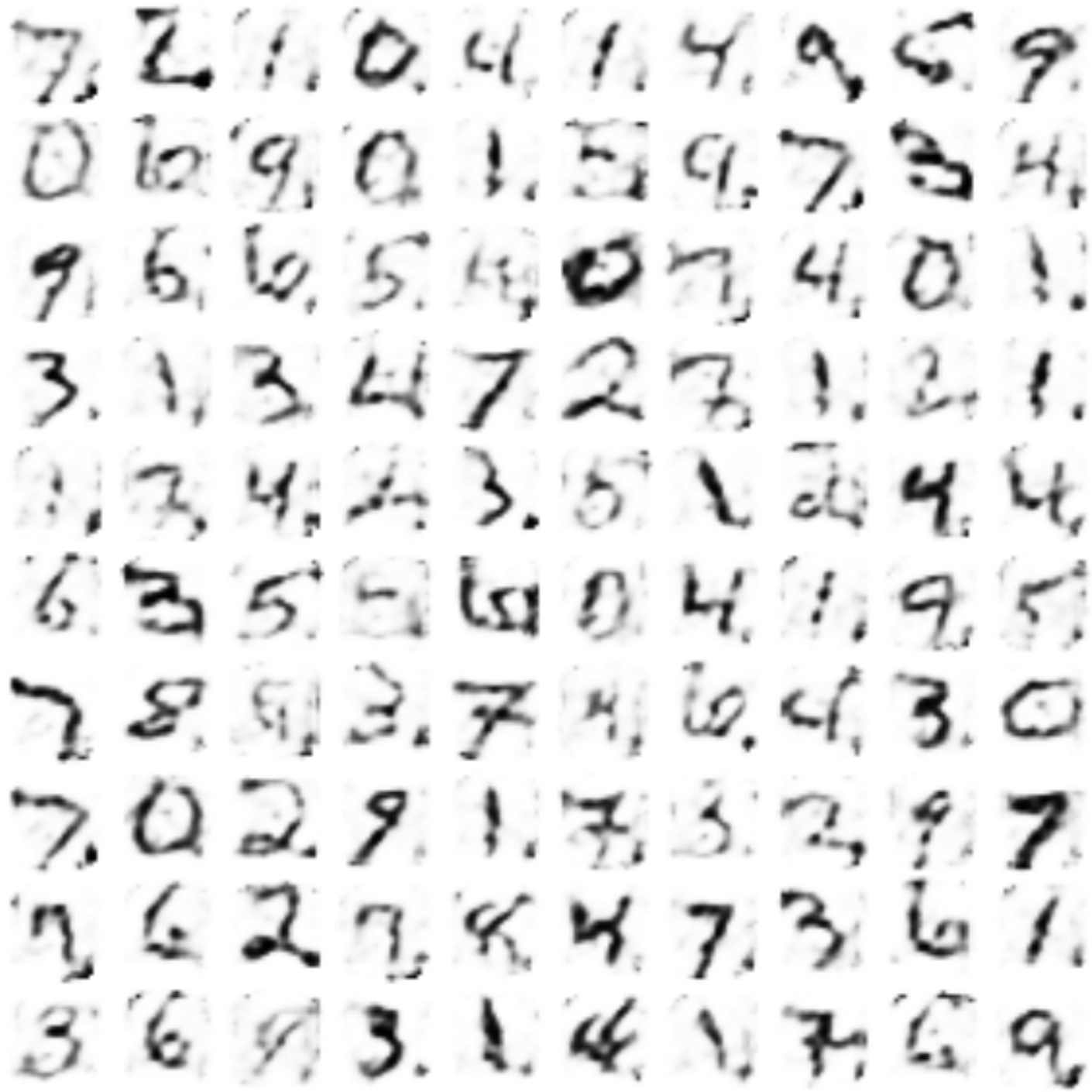}
\rule{0.4pt}{5.7cm}
\includegraphics[scale=0.4]{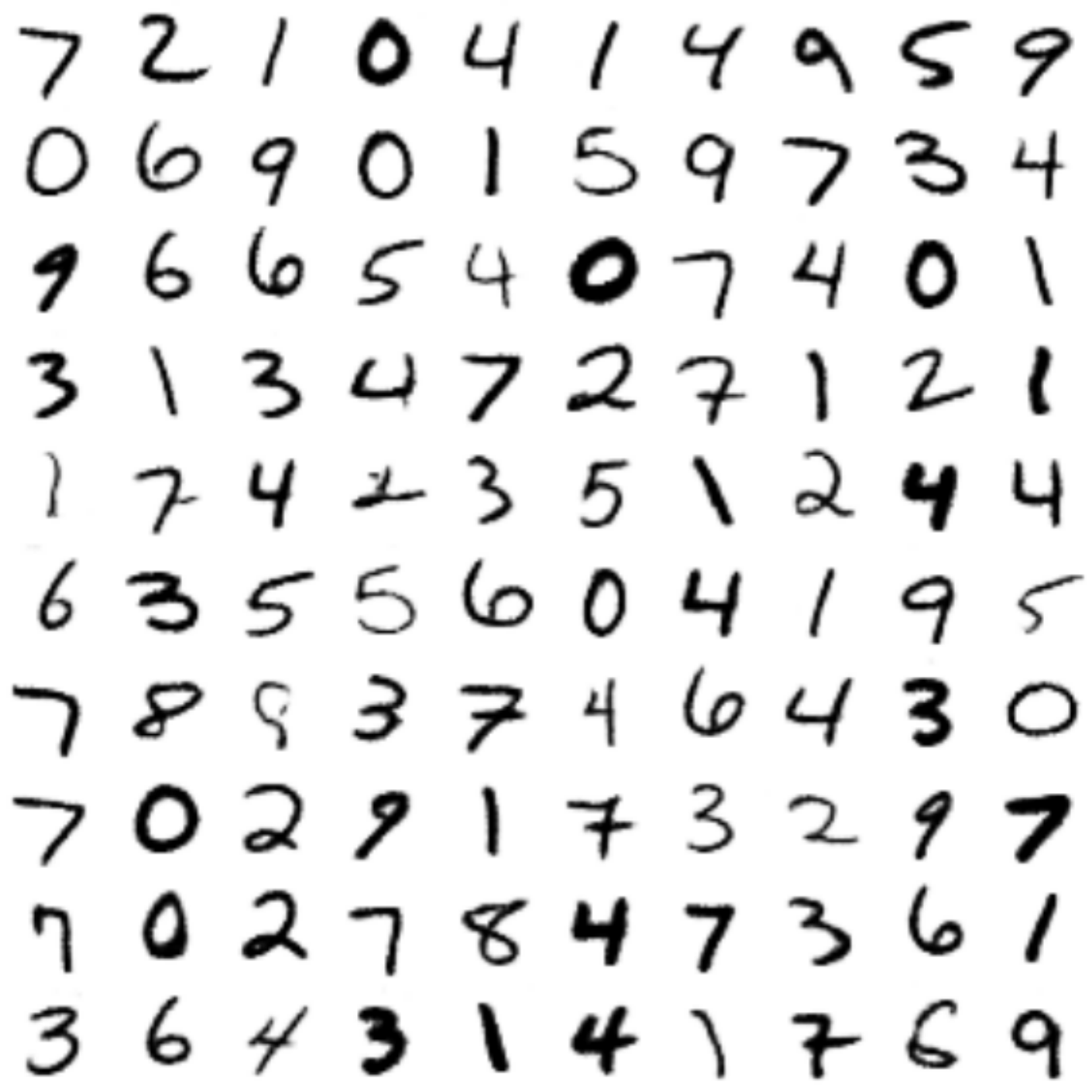}
\end{center}
\caption{\textbf{Untargeted FGS $\calL_{\VAE}$ Attack:} VAE reconstructions (left) and VAE-GAN reconstructions (right).
Note the difference in reconstructions compared to Figure~\ref{fig:original-reconstructions}.
Careful visual inspection reveals that none of the VAE reconstructions change class, and only two of the VAE-GAN reconstructions change class (a $6$ to a $0$ in the next-to-last row, and a $9$ to a $4$ in the last row).
Combining FGS with $\calL_{\VAE}$ does not seem to give an effective attack.}
\label{fig:vae-loss-reconstructions}
\end{figure}


\begin{table}
{
\tiny
\begin{center}
\begin{tabular}{|@{~~}c@{~~}|@{~~}c@{~~}|@{~~}c@{~~}|@{~~}c@{~~}|@{~~}c@{~~}|@{~~}c@{~~}|@{~~}c@{~~}|@{~~}c@{~~}|@{~~}c@{~~}|@{~~}c@{~~}|@{~~}c@{~~}|}
\hline
\textbf{Target} & \textbf{0} & \textbf{1} & \textbf{2} & \textbf{3} & \textbf{4} & \textbf{5} & \textbf{6} & \textbf{7} & \textbf{8} & \textbf{9} \\
\hline
\textbf{Classifier accuracy} & $1.98\%$ & $0.00\%$ & $0.00\%$ & $0.00\%$ & $0.00\%$ & $0.00\%$ & $0.00\%$ & $0.00\%$ & $0.00\%$ & $0.00\%$ \\
\hline
\textbf{Matching rate} & $95.06\%$ & $100.00\%$ & $100.00\%$ & $100.00\%$ & $100.00\%$ & $100.00\%$ & $100.00\%$ & $100.00\%$ & $100.00\%$ & $99.89\%$ \\
\hline
\end{tabular}
\end{center}
}
\caption{\textbf{$L_2$ Optimization Classifier Attack on MNIST:} $\fclass$ accuracy on adversarial examples against the VAE-GAN for each target class (middle row) and the matching rate between the predictions $\fclass$ made and the adversarial target class (bottom row).
The adversarial examples successfully fool $\fclass$ into predicting the target class almost $100\%$ of the time, which makes this attack seem like a strong attack, but the attack actually fails to generate good reconstructions in many cases.
Reconstructions for target classes $0$ and $4$ can be seen in Figure~\ref{fig:baseline-targeted-reconstructions-0} and Figure~\ref{fig:classifier-targeted-reconstructions-4}.}
\label{tab:baseline-targeted-results}
\end{table}

\begin{table}
{
\tiny
\begin{center}
\begin{tabular}{|c|c|c|c|c|c|c|c|c|c|c|}
\hline
\textbf{Source} & \textbf{Target 0} & \textbf{Target 1} & \textbf{Target 2} & \textbf{Target 3} & \textbf{Target 4} & \textbf{Target 5} & \textbf{Target 6} & \textbf{Target 7} & \textbf{Target 8} & \textbf{Target 9} \\
\hline
\textbf{0} & - & \begin{tabular}{@{}c@{}}85.54\% \\ (34.94\%)\end{tabular} & \begin{tabular}{@{}c@{}}100.00\% \\ (100.00\%)\end{tabular} & \begin{tabular}{@{}c@{}}100.00\% \\ (13.25\%)\end{tabular} & \begin{tabular}{@{}c@{}}75.90\% \\ (75.90\%)\end{tabular} & \begin{tabular}{@{}c@{}}96.39\% \\ (92.77\%)\end{tabular} & \begin{tabular}{@{}c@{}}100.00\% \\ (100.00\%)\end{tabular} & \begin{tabular}{@{}c@{}}96.39\% \\ (91.57\%)\end{tabular} & \begin{tabular}{@{}c@{}}0.00\% \\ (0.00\%)\end{tabular} & \begin{tabular}{@{}c@{}}100.00\% \\ (83.13\%)\end{tabular} \\
\hline
\textbf{1} & \begin{tabular}{@{}c@{}}100.00\% \\ (100.00\%)\end{tabular} & - & \begin{tabular}{@{}c@{}}100.00\% \\ (100.00\%)\end{tabular} & \begin{tabular}{@{}c@{}}100.00\% \\ (0.00\%)\end{tabular} & \begin{tabular}{@{}c@{}}100.00\% \\ (93.60\%)\end{tabular} & \begin{tabular}{@{}c@{}}100.00\% \\ (100.00\%)\end{tabular} & \begin{tabular}{@{}c@{}}100.00\% \\ (100.00\%)\end{tabular} & \begin{tabular}{@{}c@{}}100.00\% \\ (100.00\%)\end{tabular} & \begin{tabular}{@{}c@{}}100.00\% \\ (0.00\%)\end{tabular} & \begin{tabular}{@{}c@{}}100.00\% \\ (98.40\%)\end{tabular} \\
\hline
\textbf{2} & \begin{tabular}{@{}c@{}}100.00\% \\ (100.00\%)\end{tabular} & \begin{tabular}{@{}c@{}}97.37\% \\ (55.26\%)\end{tabular} & - & \begin{tabular}{@{}c@{}}100.00\% \\ (55.26\%)\end{tabular} & \begin{tabular}{@{}c@{}}97.37\% \\ (88.60\%)\end{tabular} & \begin{tabular}{@{}c@{}}95.61\% \\ (74.56\%)\end{tabular} & \begin{tabular}{@{}c@{}}100.00\% \\ (100.00\%)\end{tabular} & \begin{tabular}{@{}c@{}}99.12\% \\ (94.74\%)\end{tabular} & \begin{tabular}{@{}c@{}}100.00\% \\ (0.00\%)\end{tabular} & \begin{tabular}{@{}c@{}}100.00\% \\ (92.98\%)\end{tabular} \\
\hline
\textbf{3} & \begin{tabular}{@{}c@{}}100.00\% \\ (100.00\%)\end{tabular} & \begin{tabular}{@{}c@{}}90.65\% \\ (89.72\%)\end{tabular} & \begin{tabular}{@{}c@{}}100.00\% \\ (100.00\%)\end{tabular} & - & \begin{tabular}{@{}c@{}}100.00\% \\ (91.59\%)\end{tabular} & \begin{tabular}{@{}c@{}}94.39\% \\ (94.39\%)\end{tabular} & \begin{tabular}{@{}c@{}}100.00\% \\ (100.00\%)\end{tabular} & \begin{tabular}{@{}c@{}}85.05\% \\ (84.11\%)\end{tabular} & \begin{tabular}{@{}c@{}}100.00\% \\ (0.00\%)\end{tabular} & \begin{tabular}{@{}c@{}}90.65\% \\ (88.79\%)\end{tabular} \\
\hline
\textbf{4} & \begin{tabular}{@{}c@{}}100.00\% \\ (100.00\%)\end{tabular} & \begin{tabular}{@{}c@{}}97.27\% \\ (67.27\%)\end{tabular} & \begin{tabular}{@{}c@{}}100.00\% \\ (100.00\%)\end{tabular} & \begin{tabular}{@{}c@{}}100.00\% \\ (18.18\%)\end{tabular} & - & \begin{tabular}{@{}c@{}}100.00\% \\ (100.00\%)\end{tabular} & \begin{tabular}{@{}c@{}}100.00\% \\ (100.00\%)\end{tabular} & \begin{tabular}{@{}c@{}}100.00\% \\ (100.00\%)\end{tabular} & \begin{tabular}{@{}c@{}}100.00\% \\ (0.00\%)\end{tabular} & \begin{tabular}{@{}c@{}}100.00\% \\ (100.00\%)\end{tabular} \\
\hline
\textbf{5} & \begin{tabular}{@{}c@{}}100.00\% \\ (100.00\%)\end{tabular} & \begin{tabular}{@{}c@{}}96.55\% \\ (80.46\%)\end{tabular} & \begin{tabular}{@{}c@{}}100.00\% \\ (100.00\%)\end{tabular} & \begin{tabular}{@{}c@{}}2.30\% \\ (2.30\%)\end{tabular} & \begin{tabular}{@{}c@{}}100.00\% \\ (96.55\%)\end{tabular} & - & \begin{tabular}{@{}c@{}}100.00\% \\ (100.00\%)\end{tabular} & \begin{tabular}{@{}c@{}}98.85\% \\ (89.66\%)\end{tabular} & \begin{tabular}{@{}c@{}}100.00\% \\ (0.00\%)\end{tabular} & \begin{tabular}{@{}c@{}}95.40\% \\ (94.25\%)\end{tabular} \\
\hline
\textbf{6} & \begin{tabular}{@{}c@{}}100.00\% \\ (100.00\%)\end{tabular} & \begin{tabular}{@{}c@{}}87.36\% \\ (80.46\%)\end{tabular} & \begin{tabular}{@{}c@{}}100.00\% \\ (100.00\%)\end{tabular} & \begin{tabular}{@{}c@{}}100.00\% \\ (11.49\%)\end{tabular} & \begin{tabular}{@{}c@{}}100.00\% \\ (97.70\%)\end{tabular} & \begin{tabular}{@{}c@{}}100.00\% \\ (100.00\%)\end{tabular} & - & \begin{tabular}{@{}c@{}}100.00\% \\ (98.85\%)\end{tabular} & \begin{tabular}{@{}c@{}}100.00\% \\ (0.00\%)\end{tabular} & \begin{tabular}{@{}c@{}}100.00\% \\ (96.55\%)\end{tabular} \\
\hline
\textbf{7} & \begin{tabular}{@{}c@{}}100.00\% \\ (100.00\%)\end{tabular} & \begin{tabular}{@{}c@{}}90.91\% \\ (82.83\%)\end{tabular} & \begin{tabular}{@{}c@{}}100.00\% \\ (100.00\%)\end{tabular} & \begin{tabular}{@{}c@{}}100.00\% \\ (16.16\%)\end{tabular} & \begin{tabular}{@{}c@{}}100.00\% \\ (79.80\%)\end{tabular} & \begin{tabular}{@{}c@{}}100.00\% \\ (98.99\%)\end{tabular} & \begin{tabular}{@{}c@{}}100.00\% \\ (100.00\%)\end{tabular} & - & \begin{tabular}{@{}c@{}}100.00\% \\ (0.00\%)\end{tabular} & \begin{tabular}{@{}c@{}}100.00\% \\ (100.00\%)\end{tabular} \\
\hline
\textbf{8} & \begin{tabular}{@{}c@{}}100.00\% \\ (100.00\%)\end{tabular} & \begin{tabular}{@{}c@{}}89.77\% \\ (71.59\%)\end{tabular} & \begin{tabular}{@{}c@{}}100.00\% \\ (100.00\%)\end{tabular} & \begin{tabular}{@{}c@{}}100.00\% \\ (35.23\%)\end{tabular} & \begin{tabular}{@{}c@{}}100.00\% \\ (97.73\%)\end{tabular} & \begin{tabular}{@{}c@{}}89.77\% \\ (62.50\%)\end{tabular} & \begin{tabular}{@{}c@{}}100.00\% \\ (100.00\%)\end{tabular} & \begin{tabular}{@{}c@{}}98.86\% \\ (92.05\%)\end{tabular} & - & \begin{tabular}{@{}c@{}}98.86\% \\ (96.59\%)\end{tabular} \\
\hline
\textbf{9} & \begin{tabular}{@{}c@{}}100.00\% \\ (100.00\%)\end{tabular} & \begin{tabular}{@{}c@{}}95.65\% \\ (75.00\%)\end{tabular} & \begin{tabular}{@{}c@{}}100.00\% \\ (100.00\%)\end{tabular} & \begin{tabular}{@{}c@{}}100.00\% \\ (18.48\%)\end{tabular} & \begin{tabular}{@{}c@{}}100.00\% \\ (97.83\%)\end{tabular} & \begin{tabular}{@{}c@{}}100.00\% \\ (95.65\%)\end{tabular} & \begin{tabular}{@{}c@{}}100.00\% \\ (100.00\%)\end{tabular} & \begin{tabular}{@{}c@{}}100.00\% \\ (100.00\%)\end{tabular} & \begin{tabular}{@{}c@{}}100.00\% \\ (0.00\%)\end{tabular} & - \\
\hline
\end{tabular}
\end{center}
}
\caption{\textbf{$L_2$ Optimization Latent Attack on MNIST (single latent vector target):} $\ASuntargeted$\ ($\AStargeted$\ in parentheses) after one reconstruction loop for different source and target class pairs on the VAE-GAN model.
The latent representation of a random image from the target class is used to generate the target latent vector.
Higher values indicate more successful attacks against the generative model.}
\label{tab:direct-targeted-random-numerical}
\end{table}

\begin{table}[h]
{
\tiny
\begin{center}
\begin{tabular}{|c|c|c|c|c|c|c|c|c|c|c|}
\hline
\textbf{Source} & \textbf{Target 0} & \textbf{Target 1} & \textbf{Target 2} & \textbf{Target 3} & \textbf{Target 4} & \textbf{Target 5} & \textbf{Target 6} & \textbf{Target 7} & \textbf{Target 8} & \textbf{Target 9} \\
\hline
\textbf{0} & - & \begin{tabular}{@{}c@{}}40.96\% \\ (1.20\%)\end{tabular} & \begin{tabular}{@{}c@{}}6.02\% \\ (4.82\%)\end{tabular} & \begin{tabular}{@{}c@{}}10.84\% \\ (7.23\%)\end{tabular} & \begin{tabular}{@{}c@{}}75.90\% \\ (0.00\%)\end{tabular} & \begin{tabular}{@{}c@{}}6.02\% \\ (3.61\%)\end{tabular} & \begin{tabular}{@{}c@{}}28.92\% \\ (28.92\%)\end{tabular} & \begin{tabular}{@{}c@{}}37.35\% \\ (20.48\%)\end{tabular} & \begin{tabular}{@{}c@{}}6.02\% \\ (1.20\%)\end{tabular} & \begin{tabular}{@{}c@{}}10.84\% \\ (3.61\%)\end{tabular} \\
\hline
\textbf{1} & \begin{tabular}{@{}c@{}}99.20\% \\ (77.60\%)\end{tabular} & - & \begin{tabular}{@{}c@{}}7.20\% \\ (5.60\%)\end{tabular} & \begin{tabular}{@{}c@{}}1.60\% \\ (1.60\%)\end{tabular} & \begin{tabular}{@{}c@{}}85.60\% \\ (0.00\%)\end{tabular} & \begin{tabular}{@{}c@{}}8.00\% \\ (5.60\%)\end{tabular} & \begin{tabular}{@{}c@{}}28.80\% \\ (28.00\%)\end{tabular} & \begin{tabular}{@{}c@{}}8.80\% \\ (7.20\%)\end{tabular} & \begin{tabular}{@{}c@{}}3.20\% \\ (1.60\%)\end{tabular} & \begin{tabular}{@{}c@{}}69.60\% \\ (0.80\%)\end{tabular} \\
\hline
\textbf{2} & \begin{tabular}{@{}c@{}}85.96\% \\ (80.70\%)\end{tabular} & \begin{tabular}{@{}c@{}}3.51\% \\ (2.63\%)\end{tabular} & - & \begin{tabular}{@{}c@{}}29.82\% \\ (23.68\%)\end{tabular} & \begin{tabular}{@{}c@{}}78.95\% \\ (0.00\%)\end{tabular} & \begin{tabular}{@{}c@{}}72.81\% \\ (20.18\%)\end{tabular} & \begin{tabular}{@{}c@{}}72.81\% \\ (46.49\%)\end{tabular} & \begin{tabular}{@{}c@{}}35.09\% \\ (8.77\%)\end{tabular} & \begin{tabular}{@{}c@{}}41.23\% \\ (12.28\%)\end{tabular} & \begin{tabular}{@{}c@{}}68.42\% \\ (2.63\%)\end{tabular} \\
\hline
\textbf{3} & \begin{tabular}{@{}c@{}}93.46\% \\ (83.18\%)\end{tabular} & \begin{tabular}{@{}c@{}}26.17\% \\ (12.15\%)\end{tabular} & \begin{tabular}{@{}c@{}}27.10\% \\ (16.82\%)\end{tabular} & - & \begin{tabular}{@{}c@{}}67.29\% \\ (0.00\%)\end{tabular} & \begin{tabular}{@{}c@{}}66.36\% \\ (62.62\%)\end{tabular} & \begin{tabular}{@{}c@{}}87.85\% \\ (22.43\%)\end{tabular} & \begin{tabular}{@{}c@{}}50.47\% \\ (27.10\%)\end{tabular} & \begin{tabular}{@{}c@{}}23.36\% \\ (8.41\%)\end{tabular} & \begin{tabular}{@{}c@{}}33.64\% \\ (8.41\%)\end{tabular} \\
\hline
\textbf{4} & \begin{tabular}{@{}c@{}}100.00\% \\ (82.73\%)\end{tabular} & \begin{tabular}{@{}c@{}}70.00\% \\ (48.18\%)\end{tabular} & \begin{tabular}{@{}c@{}}28.18\% \\ (10.91\%)\end{tabular} & \begin{tabular}{@{}c@{}}84.55\% \\ (17.27\%)\end{tabular} & - & \begin{tabular}{@{}c@{}}66.36\% \\ (31.82\%)\end{tabular} & \begin{tabular}{@{}c@{}}95.45\% \\ (71.82\%)\end{tabular} & \begin{tabular}{@{}c@{}}62.73\% \\ (37.27\%)\end{tabular} & \begin{tabular}{@{}c@{}}20.91\% \\ (0.91\%)\end{tabular} & \begin{tabular}{@{}c@{}}51.82\% \\ (44.55\%)\end{tabular} \\
\hline
\textbf{5} & \begin{tabular}{@{}c@{}}93.10\% \\ (89.66\%)\end{tabular} & \begin{tabular}{@{}c@{}}21.84\% \\ (1.15\%)\end{tabular} & \begin{tabular}{@{}c@{}}68.97\% \\ (11.49\%)\end{tabular} & \begin{tabular}{@{}c@{}}28.74\% \\ (18.39\%)\end{tabular} & \begin{tabular}{@{}c@{}}3.45\% \\ (0.00\%)\end{tabular} & - & \begin{tabular}{@{}c@{}}20.69\% \\ (19.54\%)\end{tabular} & \begin{tabular}{@{}c@{}}80.46\% \\ (41.38\%)\end{tabular} & \begin{tabular}{@{}c@{}}22.99\% \\ (2.30\%)\end{tabular} & \begin{tabular}{@{}c@{}}44.83\% \\ (12.64\%)\end{tabular} \\
\hline
\textbf{6} & \begin{tabular}{@{}c@{}}29.89\% \\ (28.74\%)\end{tabular} & \begin{tabular}{@{}c@{}}44.83\% \\ (1.15\%)\end{tabular} & \begin{tabular}{@{}c@{}}24.14\% \\ (3.45\%)\end{tabular} & \begin{tabular}{@{}c@{}}59.77\% \\ (11.49\%)\end{tabular} & \begin{tabular}{@{}c@{}}77.01\% \\ (0.00\%)\end{tabular} & \begin{tabular}{@{}c@{}}10.34\% \\ (8.05\%)\end{tabular} & - & \begin{tabular}{@{}c@{}}62.07\% \\ (8.05\%)\end{tabular} & \begin{tabular}{@{}c@{}}8.05\% \\ (0.00\%)\end{tabular} & \begin{tabular}{@{}c@{}}75.86\% \\ (4.60\%)\end{tabular} \\
\hline
\textbf{7} & \begin{tabular}{@{}c@{}}79.80\% \\ (65.66\%)\end{tabular} & \begin{tabular}{@{}c@{}}77.78\% \\ (26.26\%)\end{tabular} & \begin{tabular}{@{}c@{}}20.20\% \\ (8.08\%)\end{tabular} & \begin{tabular}{@{}c@{}}8.08\% \\ (4.04\%)\end{tabular} & \begin{tabular}{@{}c@{}}100.00\% \\ (0.00\%)\end{tabular} & \begin{tabular}{@{}c@{}}56.57\% \\ (23.23\%)\end{tabular} & \begin{tabular}{@{}c@{}}97.98\% \\ (17.17\%)\end{tabular} & - & \begin{tabular}{@{}c@{}}38.38\% \\ (1.01\%)\end{tabular} & \begin{tabular}{@{}c@{}}17.17\% \\ (10.10\%)\end{tabular} \\
\hline
\textbf{8} & \begin{tabular}{@{}c@{}}94.32\% \\ (84.09\%)\end{tabular} & \begin{tabular}{@{}c@{}}96.59\% \\ (18.18\%)\end{tabular} & \begin{tabular}{@{}c@{}}60.23\% \\ (42.05\%)\end{tabular} & \begin{tabular}{@{}c@{}}57.95\% \\ (43.18\%)\end{tabular} & \begin{tabular}{@{}c@{}}100.00\% \\ (0.00\%)\end{tabular} & \begin{tabular}{@{}c@{}}93.18\% \\ (80.68\%)\end{tabular} & \begin{tabular}{@{}c@{}}100.00\% \\ (57.95\%)\end{tabular} & \begin{tabular}{@{}c@{}}100.00\% \\ (34.09\%)\end{tabular} & - & \begin{tabular}{@{}c@{}}87.50\% \\ (26.14\%)\end{tabular} \\
\hline
\textbf{9} & \begin{tabular}{@{}c@{}}98.91\% \\ (79.35\%)\end{tabular} & \begin{tabular}{@{}c@{}}97.83\% \\ (33.70\%)\end{tabular} & \begin{tabular}{@{}c@{}}26.09\% \\ (1.09\%)\end{tabular} & \begin{tabular}{@{}c@{}}17.39\% \\ (2.17\%)\end{tabular} & \begin{tabular}{@{}c@{}}100.00\% \\ (0.00\%)\end{tabular} & \begin{tabular}{@{}c@{}}22.83\% \\ (21.74\%)\end{tabular} & \begin{tabular}{@{}c@{}}100.00\% \\ (30.43\%)\end{tabular} & \begin{tabular}{@{}c@{}}47.83\% \\ (43.48\%)\end{tabular} & \begin{tabular}{@{}c@{}}31.52\% \\ (4.35\%)\end{tabular} & - \\
\hline
\end{tabular}
\end{center}
}
\caption{\textbf{$L_2$ Optimization Classifier Attack on MNIST:} $\ASuntargeted$\ ($\AStargeted$\ in parentheses) for all source and target class pairs using adversarial examples generated on the VAE-GAN model.
Higher values indicate more successful attacks against the generative model.}
\label{tab:indirect-distribution-classification-accuracies}
\end{table}

\begin{figure}[h]
\begin{center}
\includegraphics[scale=0.47]{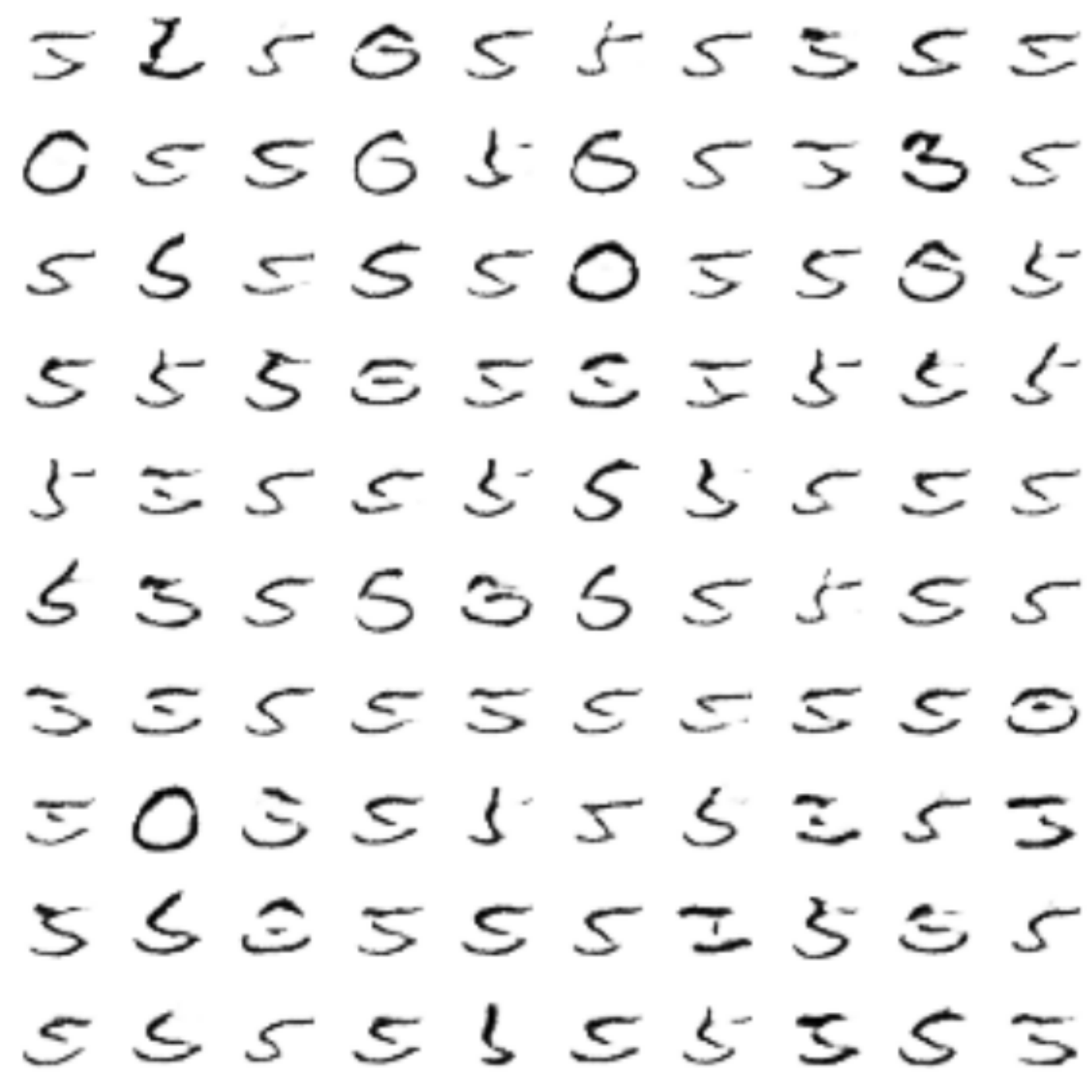}
\end{center}
\caption{\textbf{$L_2$ Optimization Classifier Attack:} Reconstructions of the first $100$ adversarial examples targeting $4$, demonstrating why the $\AStargeted$ metric is $0$ for all source digits.}
\label{fig:classifier-targeted-reconstructions-4}
\end{figure}

\begin{figure}[h]
\begin{center}
\includegraphics[scale=0.47]{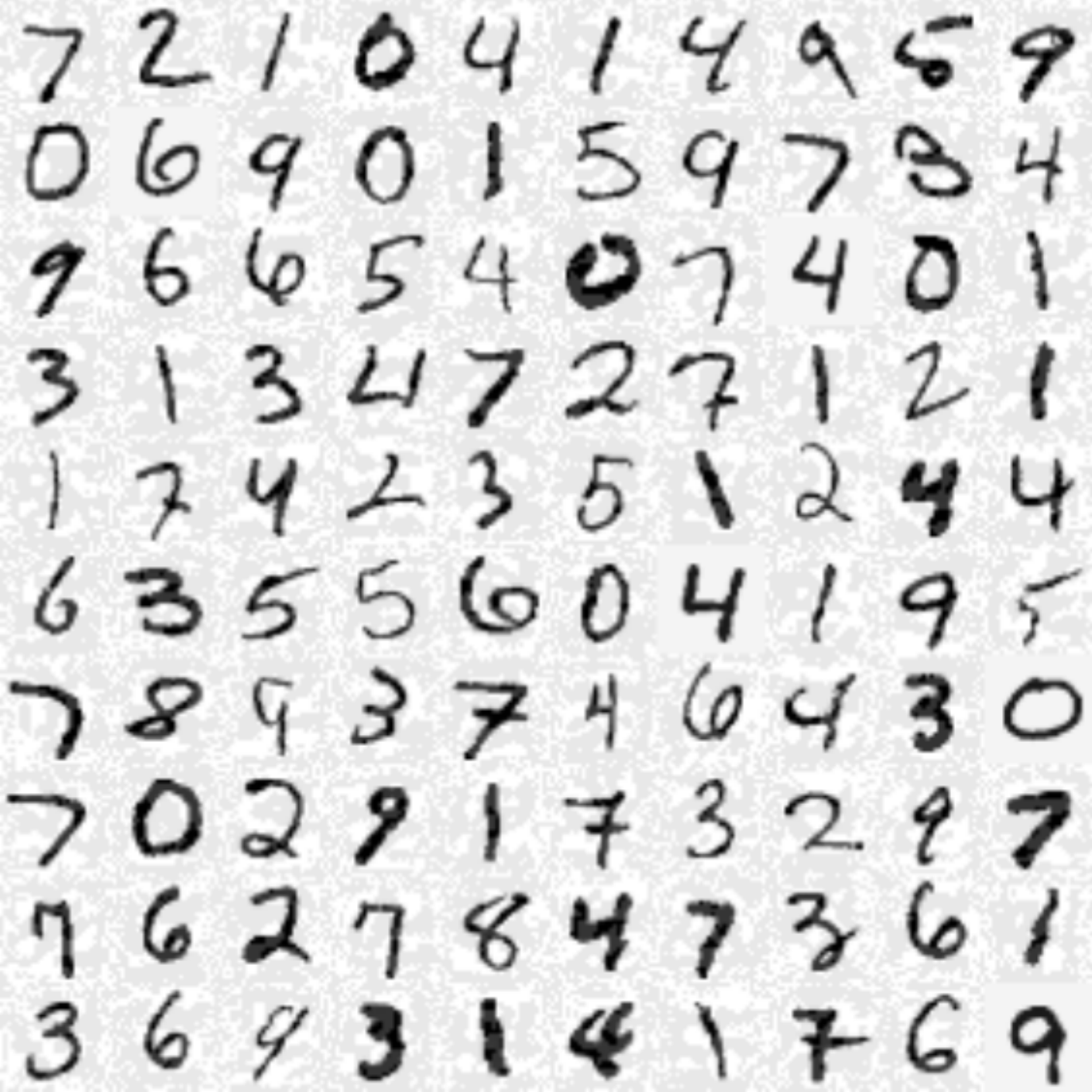} ~
\includegraphics[scale=0.47]{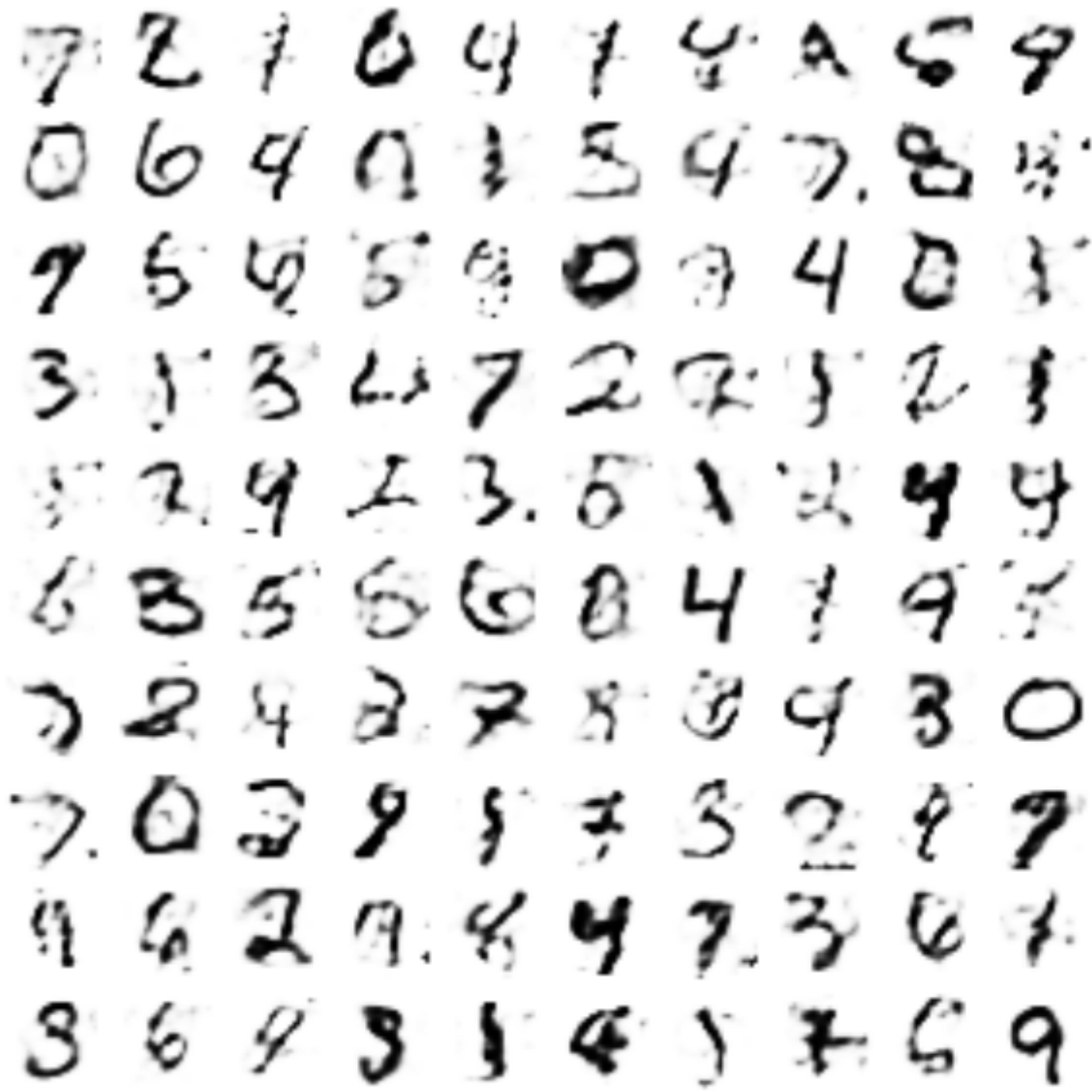} \\ \vspace{0.5cm}
\includegraphics[scale=0.47]{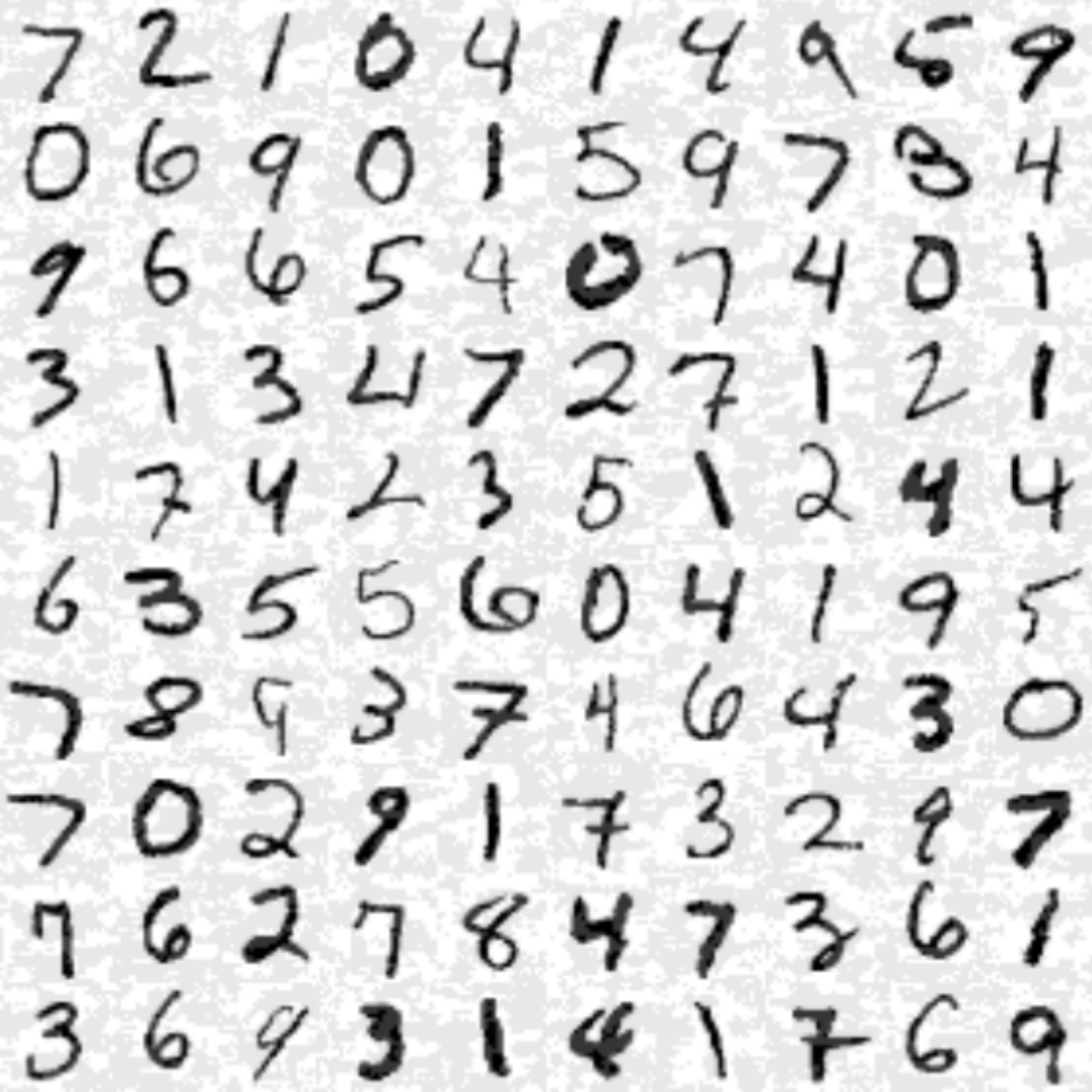} ~
\includegraphics[scale=0.47]{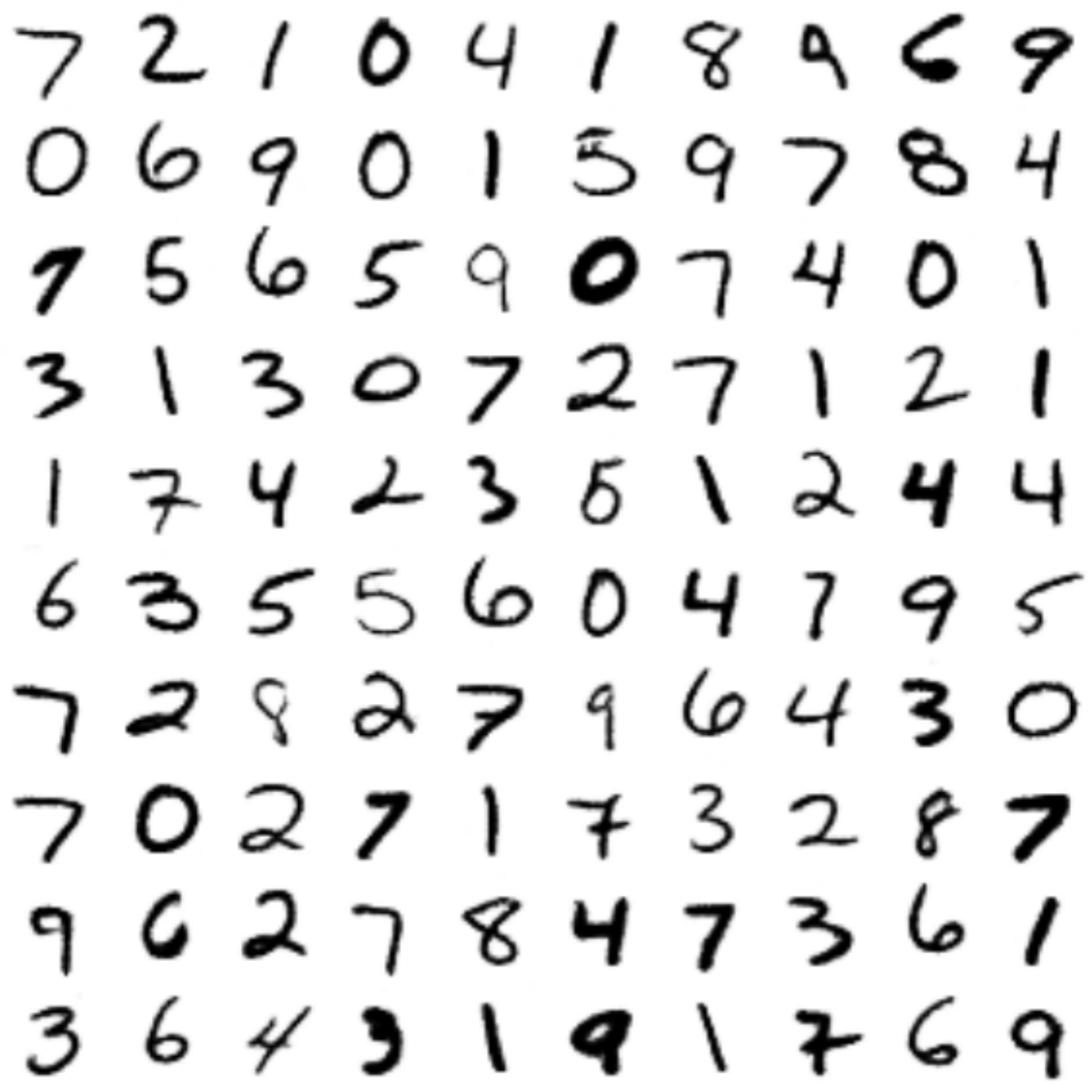}
\end{center}
\caption{\textbf{Untargeted FGS Classifer Attack:} Adversarial examples (left) and their reconstructions by the generative model (right) for the first 100 images from the MNIST validation set.
Top results are for VAE, while bottom results are for VAE-GAN.
Note the difference in quality of the reconstructed adversarial examples.}
\label{fig:reconstructions}
\end{figure}

\begin{figure}[h]
\begin{center}
\includegraphics[scale=0.47]{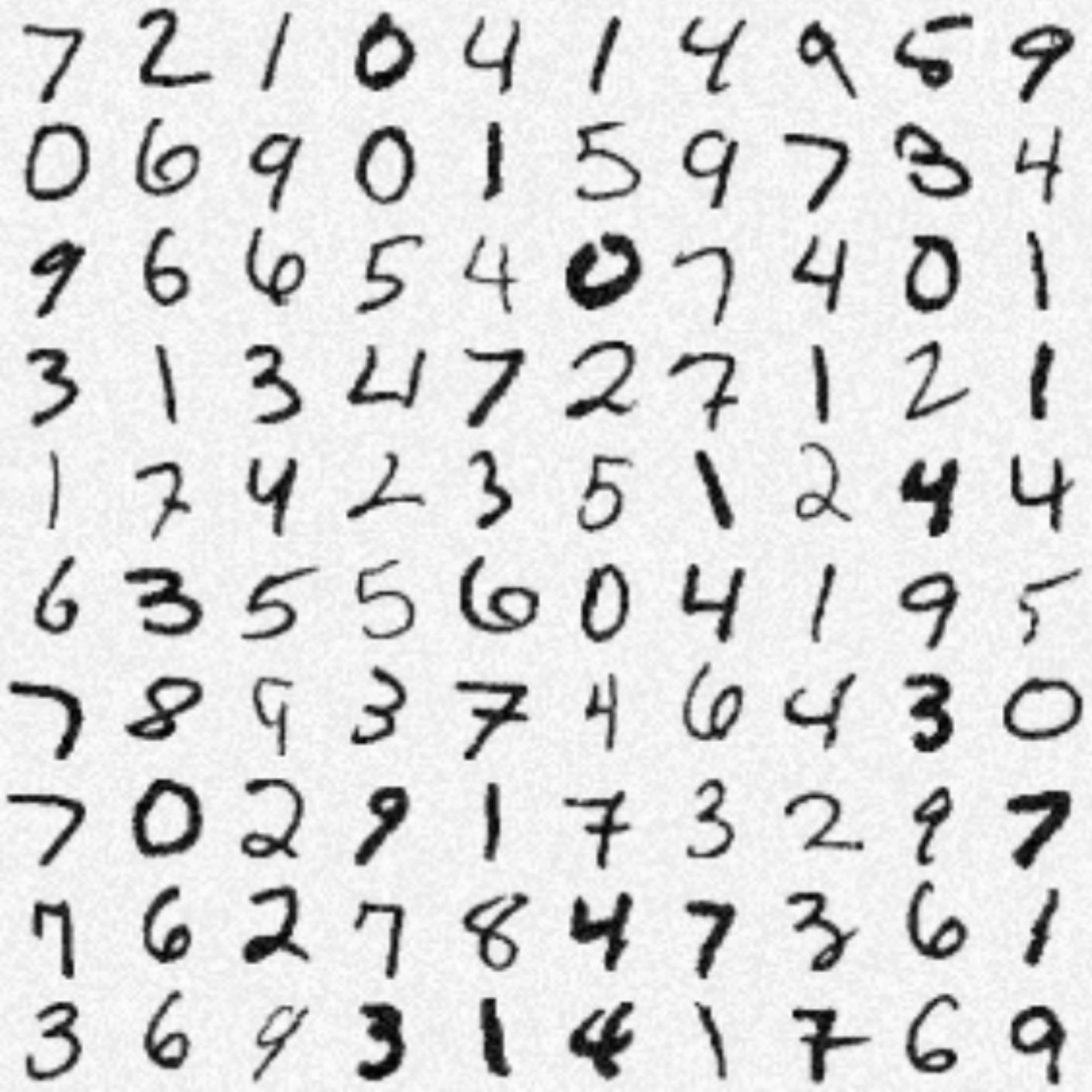} \\ \vspace{0.5cm}
\includegraphics[scale=0.47]{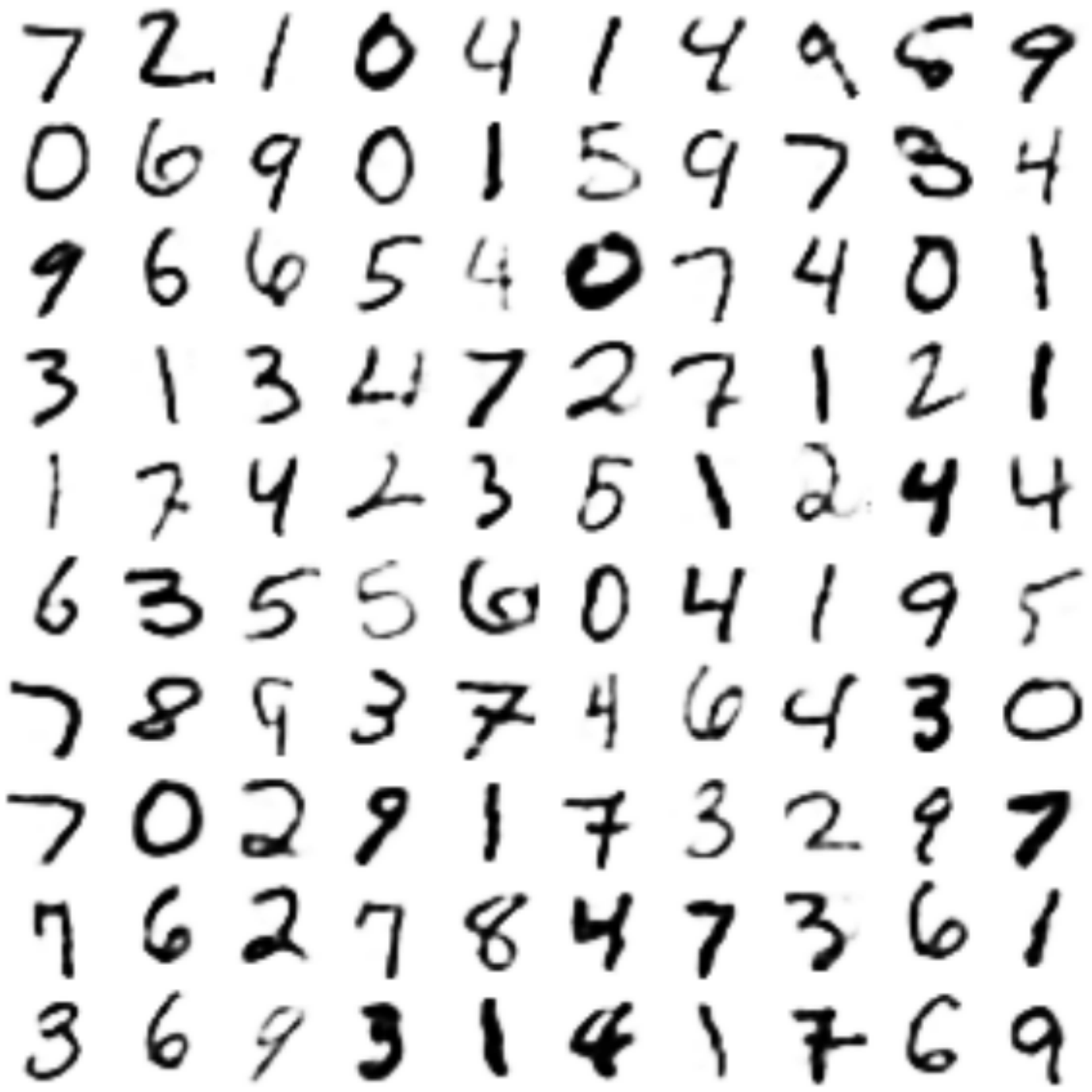} ~
\includegraphics[scale=0.47]{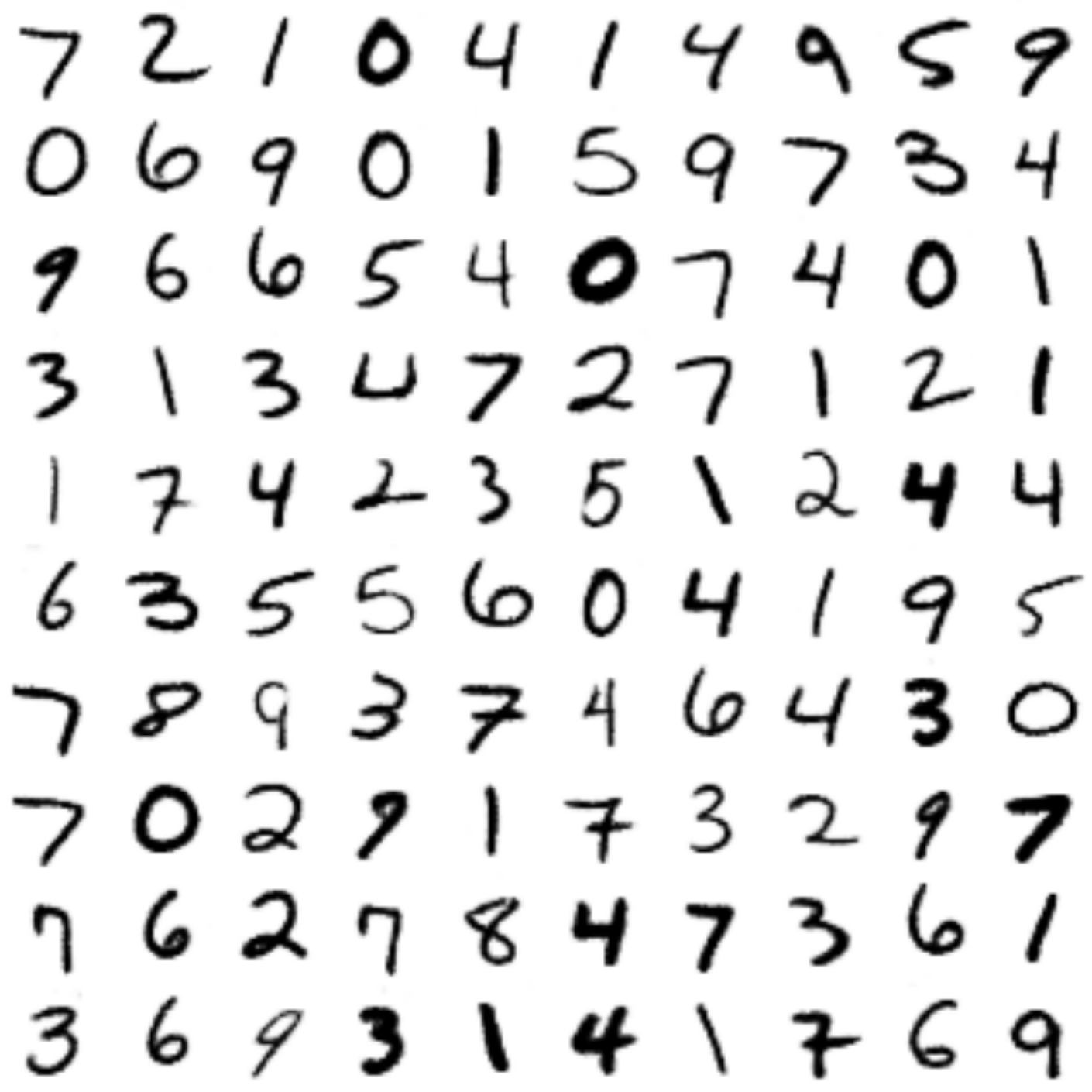}
\end{center}
\caption{Original images with random noise added (top) and their reconstructions by VAE (bottom left) and VAE-GAN (bottom right).
The magnitude of the random noise is the same as for the generated adversarial noise shown in Figure~\ref{fig:reconstructions}.
Random noise does not cause the reconstructed images to change in a significant way.}
\label{fig:random-noise-reconstructions}
\end{figure}

\begin{table}[h]
{
\tiny
\begin{center}
\begin{tabular}{|c|c|c|c|c|c|c|c|c|c|c|}
\hline
\textbf{Source} & \textbf{Target 0} & \textbf{Target 1} & \textbf{Target 2} & \textbf{Target 3} & \textbf{Target 4} & \textbf{Target 5} & \textbf{Target 6} & \textbf{Target 7} & \textbf{Target 8} & \textbf{Target 9} \\
\hline
\textbf{0} & - & \begin{tabular}{@{}c@{}}90.36\% \\ (14.46\%)\end{tabular} & \begin{tabular}{@{}c@{}}100.00\% \\ (100.00\%)\end{tabular} & \begin{tabular}{@{}c@{}}100.00\% \\ (98.80\%)\end{tabular} & \begin{tabular}{@{}c@{}}100.00\% \\ (61.45\%)\end{tabular} & \begin{tabular}{@{}c@{}}91.57\% \\ (90.36\%)\end{tabular} & \begin{tabular}{@{}c@{}}100.00\% \\ (96.39\%)\end{tabular} & \begin{tabular}{@{}c@{}}68.67\% \\ (50.60\%)\end{tabular} & \begin{tabular}{@{}c@{}}100.00\% \\ (91.57\%)\end{tabular} & \begin{tabular}{@{}c@{}}98.80\% \\ (37.35\%)\end{tabular} \\
\hline
\textbf{1} & \begin{tabular}{@{}c@{}}100.00\% \\ (100.00\%)\end{tabular} & - & \begin{tabular}{@{}c@{}}100.00\% \\ (100.00\%)\end{tabular} & \begin{tabular}{@{}c@{}}100.00\% \\ (100.00\%)\end{tabular} & \begin{tabular}{@{}c@{}}100.00\% \\ (99.20\%)\end{tabular} & \begin{tabular}{@{}c@{}}100.00\% \\ (100.00\%)\end{tabular} & \begin{tabular}{@{}c@{}}100.00\% \\ (97.60\%)\end{tabular} & \begin{tabular}{@{}c@{}}100.00\% \\ (96.00\%)\end{tabular} & \begin{tabular}{@{}c@{}}100.00\% \\ (100.00\%)\end{tabular} & \begin{tabular}{@{}c@{}}100.00\% \\ (96.00\%)\end{tabular} \\
\hline
\textbf{2} & \begin{tabular}{@{}c@{}}100.00\% \\ (100.00\%)\end{tabular} & \begin{tabular}{@{}c@{}}84.21\% \\ (60.53\%)\end{tabular} & - & \begin{tabular}{@{}c@{}}100.00\% \\ (100.00\%)\end{tabular} & \begin{tabular}{@{}c@{}}90.35\% \\ (71.93\%)\end{tabular} & \begin{tabular}{@{}c@{}}100.00\% \\ (85.96\%)\end{tabular} & \begin{tabular}{@{}c@{}}88.60\% \\ (88.60\%)\end{tabular} & \begin{tabular}{@{}c@{}}97.37\% \\ (76.32\%)\end{tabular} & \begin{tabular}{@{}c@{}}94.74\% \\ (94.74\%)\end{tabular} & \begin{tabular}{@{}c@{}}97.37\% \\ (35.09\%)\end{tabular} \\
\hline
\textbf{3} & \begin{tabular}{@{}c@{}}100.00\% \\ (100.00\%)\end{tabular} & \begin{tabular}{@{}c@{}}75.70\% \\ (66.36\%)\end{tabular} & \begin{tabular}{@{}c@{}}100.00\% \\ (100.00\%)\end{tabular} & - & \begin{tabular}{@{}c@{}}94.39\% \\ (52.34\%)\end{tabular} & \begin{tabular}{@{}c@{}}99.07\% \\ (99.07\%)\end{tabular} & \begin{tabular}{@{}c@{}}98.13\% \\ (82.24\%)\end{tabular} & \begin{tabular}{@{}c@{}}64.49\% \\ (53.27\%)\end{tabular} & \begin{tabular}{@{}c@{}}100.00\% \\ (96.26\%)\end{tabular} & \begin{tabular}{@{}c@{}}67.29\% \\ (31.78\%)\end{tabular} \\
\hline
\textbf{4} & \begin{tabular}{@{}c@{}}100.00\% \\ (100.00\%)\end{tabular} & \begin{tabular}{@{}c@{}}100.00\% \\ (52.73\%)\end{tabular} & \begin{tabular}{@{}c@{}}100.00\% \\ (100.00\%)\end{tabular} & \begin{tabular}{@{}c@{}}100.00\% \\ (100.00\%)\end{tabular} & - & \begin{tabular}{@{}c@{}}100.00\% \\ (97.27\%)\end{tabular} & \begin{tabular}{@{}c@{}}100.00\% \\ (100.00\%)\end{tabular} & \begin{tabular}{@{}c@{}}100.00\% \\ (99.09\%)\end{tabular} & \begin{tabular}{@{}c@{}}100.00\% \\ (100.00\%)\end{tabular} & \begin{tabular}{@{}c@{}}85.45\% \\ (83.64\%)\end{tabular} \\
\hline
\textbf{5} & \begin{tabular}{@{}c@{}}100.00\% \\ (100.00\%)\end{tabular} & \begin{tabular}{@{}c@{}}96.55\% \\ (40.23\%)\end{tabular} & \begin{tabular}{@{}c@{}}100.00\% \\ (100.00\%)\end{tabular} & \begin{tabular}{@{}c@{}}100.00\% \\ (100.00\%)\end{tabular} & \begin{tabular}{@{}c@{}}93.10\% \\ (59.77\%)\end{tabular} & - & \begin{tabular}{@{}c@{}}100.00\% \\ (95.40\%)\end{tabular} & \begin{tabular}{@{}c@{}}93.10\% \\ (71.26\%)\end{tabular} & \begin{tabular}{@{}c@{}}96.55\% \\ (96.55\%)\end{tabular} & \begin{tabular}{@{}c@{}}83.91\% \\ (51.72\%)\end{tabular} \\
\hline
\textbf{6} & \begin{tabular}{@{}c@{}}100.00\% \\ (100.00\%)\end{tabular} & \begin{tabular}{@{}c@{}}97.70\% \\ (70.11\%)\end{tabular} & \begin{tabular}{@{}c@{}}100.00\% \\ (100.00\%)\end{tabular} & \begin{tabular}{@{}c@{}}100.00\% \\ (100.00\%)\end{tabular} & \begin{tabular}{@{}c@{}}100.00\% \\ (91.95\%)\end{tabular} & \begin{tabular}{@{}c@{}}100.00\% \\ (100.00\%)\end{tabular} & - & \begin{tabular}{@{}c@{}}97.70\% \\ (67.82\%)\end{tabular} & \begin{tabular}{@{}c@{}}100.00\% \\ (98.85\%)\end{tabular} & \begin{tabular}{@{}c@{}}95.40\% \\ (50.57\%)\end{tabular} \\
\hline
\textbf{7} & \begin{tabular}{@{}c@{}}100.00\% \\ (100.00\%)\end{tabular} & \begin{tabular}{@{}c@{}}85.86\% \\ (58.59\%)\end{tabular} & \begin{tabular}{@{}c@{}}100.00\% \\ (100.00\%)\end{tabular} & \begin{tabular}{@{}c@{}}100.00\% \\ (100.00\%)\end{tabular} & \begin{tabular}{@{}c@{}}100.00\% \\ (98.99\%)\end{tabular} & \begin{tabular}{@{}c@{}}100.00\% \\ (97.98\%)\end{tabular} & \begin{tabular}{@{}c@{}}100.00\% \\ (79.80\%)\end{tabular} & - & \begin{tabular}{@{}c@{}}100.00\% \\ (98.99\%)\end{tabular} & \begin{tabular}{@{}c@{}}100.00\% \\ (96.97\%)\end{tabular} \\
\hline
\textbf{8} & \begin{tabular}{@{}c@{}}100.00\% \\ (100.00\%)\end{tabular} & \begin{tabular}{@{}c@{}}69.32\% \\ (44.32\%)\end{tabular} & \begin{tabular}{@{}c@{}}100.00\% \\ (100.00\%)\end{tabular} & \begin{tabular}{@{}c@{}}100.00\% \\ (100.00\%)\end{tabular} & \begin{tabular}{@{}c@{}}54.55\% \\ (53.41\%)\end{tabular} & \begin{tabular}{@{}c@{}}96.59\% \\ (96.59\%)\end{tabular} & \begin{tabular}{@{}c@{}}95.45\% \\ (92.05\%)\end{tabular} & \begin{tabular}{@{}c@{}}73.86\% \\ (52.27\%)\end{tabular} & - & \begin{tabular}{@{}c@{}}42.05\% \\ (29.55\%)\end{tabular} \\
\hline
\textbf{9} & \begin{tabular}{@{}c@{}}100.00\% \\ (100.00\%)\end{tabular} & \begin{tabular}{@{}c@{}}100.00\% \\ (44.57\%)\end{tabular} & \begin{tabular}{@{}c@{}}100.00\% \\ (100.00\%)\end{tabular} & \begin{tabular}{@{}c@{}}100.00\% \\ (100.00\%)\end{tabular} & \begin{tabular}{@{}c@{}}96.74\% \\ (95.65\%)\end{tabular} & \begin{tabular}{@{}c@{}}100.00\% \\ (97.83\%)\end{tabular} & \begin{tabular}{@{}c@{}}100.00\% \\ (100.00\%)\end{tabular} & \begin{tabular}{@{}c@{}}100.00\% \\ (97.83\%)\end{tabular} & \begin{tabular}{@{}c@{}}100.00\% \\ (100.00\%)\end{tabular} & - \\
\hline
\end{tabular}
\end{center}
}
\caption{\textbf{$L_2$ Optimization $\calL_{\VAE}$ Attack on MNIST (single image target):} $\ASuntargeted$\ ($\AStargeted$\ in parentheses) for different source and target class pairs using adversarial examples generated on the VAE-GAN model.
Higher values indicate more successful attacks against the generative model.}
\label{tab:direct-lvae-distribution-classification-accuracies}
\end{table}

\begin{table}[h]
{
\tiny
\begin{center}
\begin{tabular}{|c|c|c|c|c|c|c|c|c|c|c|}
\hline
\textbf{Source} & \textbf{Target 0} & \textbf{Target 1} & \textbf{Target 2} & \textbf{Target 3} & \textbf{Target 4} & \textbf{Target 5} & \textbf{Target 6} & \textbf{Target 7} & \textbf{Target 8} & \textbf{Target 9} \\
\hline
\textbf{0} & - & \begin{tabular}{@{}c@{}}85.54\% \\ (34.94\%)\end{tabular} & \begin{tabular}{@{}c@{}}100.00\% \\ (100.00\%)\end{tabular} & \begin{tabular}{@{}c@{}}100.00\% \\ (13.25\%)\end{tabular} & \begin{tabular}{@{}c@{}}75.90\% \\ (75.90\%)\end{tabular} & \begin{tabular}{@{}c@{}}96.39\% \\ (92.77\%)\end{tabular} & \begin{tabular}{@{}c@{}}100.00\% \\ (100.00\%)\end{tabular} & \begin{tabular}{@{}c@{}}96.39\% \\ (91.57\%)\end{tabular} & \begin{tabular}{@{}c@{}}0.00\% \\ (0.00\%)\end{tabular} & \begin{tabular}{@{}c@{}}100.00\% \\ (83.13\%)\end{tabular} \\
\hline
\textbf{1} & \begin{tabular}{@{}c@{}}100.00\% \\ (100.00\%)\end{tabular} & - & \begin{tabular}{@{}c@{}}100.00\% \\ (100.00\%)\end{tabular} & \begin{tabular}{@{}c@{}}100.00\% \\ (0.00\%)\end{tabular} & \begin{tabular}{@{}c@{}}100.00\% \\ (93.60\%)\end{tabular} & \begin{tabular}{@{}c@{}}100.00\% \\ (100.00\%)\end{tabular} & \begin{tabular}{@{}c@{}}100.00\% \\ (100.00\%)\end{tabular} & \begin{tabular}{@{}c@{}}100.00\% \\ (100.00\%)\end{tabular} & \begin{tabular}{@{}c@{}}100.00\% \\ (0.00\%)\end{tabular} & \begin{tabular}{@{}c@{}}100.00\% \\ (98.40\%)\end{tabular} \\
\hline
\textbf{2} & \begin{tabular}{@{}c@{}}100.00\% \\ (100.00\%)\end{tabular} & \begin{tabular}{@{}c@{}}97.37\% \\ (55.26\%)\end{tabular} & - & \begin{tabular}{@{}c@{}}100.00\% \\ (55.26\%)\end{tabular} & \begin{tabular}{@{}c@{}}97.37\% \\ (88.60\%)\end{tabular} & \begin{tabular}{@{}c@{}}95.61\% \\ (74.56\%)\end{tabular} & \begin{tabular}{@{}c@{}}100.00\% \\ (100.00\%)\end{tabular} & \begin{tabular}{@{}c@{}}99.12\% \\ (94.74\%)\end{tabular} & \begin{tabular}{@{}c@{}}100.00\% \\ (0.00\%)\end{tabular} & \begin{tabular}{@{}c@{}}100.00\% \\ (92.98\%)\end{tabular} \\
\hline
\textbf{3} & \begin{tabular}{@{}c@{}}100.00\% \\ (100.00\%)\end{tabular} & \begin{tabular}{@{}c@{}}90.65\% \\ (89.72\%)\end{tabular} & \begin{tabular}{@{}c@{}}100.00\% \\ (100.00\%)\end{tabular} & - & \begin{tabular}{@{}c@{}}100.00\% \\ (91.59\%)\end{tabular} & \begin{tabular}{@{}c@{}}94.39\% \\ (94.39\%)\end{tabular} & \begin{tabular}{@{}c@{}}100.00\% \\ (100.00\%)\end{tabular} & \begin{tabular}{@{}c@{}}85.05\% \\ (84.11\%)\end{tabular} & \begin{tabular}{@{}c@{}}100.00\% \\ (0.00\%)\end{tabular} & \begin{tabular}{@{}c@{}}90.65\% \\ (88.79\%)\end{tabular} \\
\hline
\textbf{4} & \begin{tabular}{@{}c@{}}100.00\% \\ (100.00\%)\end{tabular} & \begin{tabular}{@{}c@{}}97.27\% \\ (67.27\%)\end{tabular} & \begin{tabular}{@{}c@{}}100.00\% \\ (100.00\%)\end{tabular} & \begin{tabular}{@{}c@{}}100.00\% \\ (18.18\%)\end{tabular} & - & \begin{tabular}{@{}c@{}}100.00\% \\ (100.00\%)\end{tabular} & \begin{tabular}{@{}c@{}}100.00\% \\ (100.00\%)\end{tabular} & \begin{tabular}{@{}c@{}}100.00\% \\ (100.00\%)\end{tabular} & \begin{tabular}{@{}c@{}}100.00\% \\ (0.00\%)\end{tabular} & \begin{tabular}{@{}c@{}}100.00\% \\ (100.00\%)\end{tabular} \\
\hline
\textbf{5} & \begin{tabular}{@{}c@{}}100.00\% \\ (100.00\%)\end{tabular} & \begin{tabular}{@{}c@{}}96.55\% \\ (80.46\%)\end{tabular} & \begin{tabular}{@{}c@{}}100.00\% \\ (100.00\%)\end{tabular} & \begin{tabular}{@{}c@{}}2.30\% \\ (2.30\%)\end{tabular} & \begin{tabular}{@{}c@{}}100.00\% \\ (96.55\%)\end{tabular} & - & \begin{tabular}{@{}c@{}}100.00\% \\ (100.00\%)\end{tabular} & \begin{tabular}{@{}c@{}}98.85\% \\ (89.66\%)\end{tabular} & \begin{tabular}{@{}c@{}}100.00\% \\ (0.00\%)\end{tabular} & \begin{tabular}{@{}c@{}}95.40\% \\ (94.25\%)\end{tabular} \\
\hline
\textbf{6} & \begin{tabular}{@{}c@{}}100.00\% \\ (100.00\%)\end{tabular} & \begin{tabular}{@{}c@{}}87.36\% \\ (80.46\%)\end{tabular} & \begin{tabular}{@{}c@{}}100.00\% \\ (100.00\%)\end{tabular} & \begin{tabular}{@{}c@{}}100.00\% \\ (11.49\%)\end{tabular} & \begin{tabular}{@{}c@{}}100.00\% \\ (97.70\%)\end{tabular} & \begin{tabular}{@{}c@{}}100.00\% \\ (100.00\%)\end{tabular} & - & \begin{tabular}{@{}c@{}}100.00\% \\ (98.85\%)\end{tabular} & \begin{tabular}{@{}c@{}}100.00\% \\ (0.00\%)\end{tabular} & \begin{tabular}{@{}c@{}}100.00\% \\ (96.55\%)\end{tabular} \\
\hline
\textbf{7} & \begin{tabular}{@{}c@{}}100.00\% \\ (100.00\%)\end{tabular} & \begin{tabular}{@{}c@{}}90.91\% \\ (82.83\%)\end{tabular} & \begin{tabular}{@{}c@{}}100.00\% \\ (100.00\%)\end{tabular} & \begin{tabular}{@{}c@{}}100.00\% \\ (16.16\%)\end{tabular} & \begin{tabular}{@{}c@{}}100.00\% \\ (79.80\%)\end{tabular} & \begin{tabular}{@{}c@{}}100.00\% \\ (98.99\%)\end{tabular} & \begin{tabular}{@{}c@{}}100.00\% \\ (100.00\%)\end{tabular} & - & \begin{tabular}{@{}c@{}}100.00\% \\ (0.00\%)\end{tabular} & \begin{tabular}{@{}c@{}}100.00\% \\ (100.00\%)\end{tabular} \\
\hline
\textbf{8} & \begin{tabular}{@{}c@{}}100.00\% \\ (100.00\%)\end{tabular} & \begin{tabular}{@{}c@{}}89.77\% \\ (71.59\%)\end{tabular} & \begin{tabular}{@{}c@{}}100.00\% \\ (100.00\%)\end{tabular} & \begin{tabular}{@{}c@{}}100.00\% \\ (35.23\%)\end{tabular} & \begin{tabular}{@{}c@{}}100.00\% \\ (97.73\%)\end{tabular} & \begin{tabular}{@{}c@{}}89.77\% \\ (62.50\%)\end{tabular} & \begin{tabular}{@{}c@{}}100.00\% \\ (100.00\%)\end{tabular} & \begin{tabular}{@{}c@{}}98.86\% \\ (92.05\%)\end{tabular} & - & \begin{tabular}{@{}c@{}}98.86\% \\ (96.59\%)\end{tabular} \\
\hline
\textbf{9} & \begin{tabular}{@{}c@{}}100.00\% \\ (100.00\%)\end{tabular} & \begin{tabular}{@{}c@{}}95.65\% \\ (75.00\%)\end{tabular} & \begin{tabular}{@{}c@{}}100.00\% \\ (100.00\%)\end{tabular} & \begin{tabular}{@{}c@{}}100.00\% \\ (18.48\%)\end{tabular} & \begin{tabular}{@{}c@{}}100.00\% \\ (97.83\%)\end{tabular} & \begin{tabular}{@{}c@{}}100.00\% \\ (95.65\%)\end{tabular} & \begin{tabular}{@{}c@{}}100.00\% \\ (100.00\%)\end{tabular} & \begin{tabular}{@{}c@{}}100.00\% \\ (100.00\%)\end{tabular} & \begin{tabular}{@{}c@{}}100.00\% \\ (0.00\%)\end{tabular} & - \\
\hline
\end{tabular}
\end{center}
}
\caption{\textbf{$L_2$ Optimization $\calL_{\VAE}$ Attack (mean reconstruction target):} $\ASuntargeted$\ ($\AStargeted$\ in parentheses) for all source and target class pairs using adversarial examples generated on the VAE-GAN model.
The mean reconstruction image for each target class (over all of the images of that class in the training set) is used as the target reconstruction.
Higher values indicate more successful attacks against the generative model.}
\label{tab:direct-lvae-targeted-random-numerical}
\end{table}

\begin{table}[h]
{
\tiny
\begin{center}
\begin{tabular}{|c|c|c|c|c|c|c|c|c|c|c|}
\hline
\textbf{Source} & \textbf{Target 0} & \textbf{Target 1} & \textbf{Target 2} & \textbf{Target 3} & \textbf{Target 4} & \textbf{Target 5} & \textbf{Target 6} & \textbf{Target 7} & \textbf{Target 8} & \textbf{Target 9} \\
\hline
\textbf{0} & - & \begin{tabular}{@{}c@{}}40.96\% \\ (10.84\%)\end{tabular} & \begin{tabular}{@{}c@{}}65.06\% \\ (65.06\%)\end{tabular} & \begin{tabular}{@{}c@{}}53.01\% \\ (46.99\%)\end{tabular} & \begin{tabular}{@{}c@{}}62.65\% \\ (54.22\%)\end{tabular} & \begin{tabular}{@{}c@{}}36.14\% \\ (36.14\%)\end{tabular} & \begin{tabular}{@{}c@{}}59.04\% \\ (59.04\%)\end{tabular} & \begin{tabular}{@{}c@{}}46.99\% \\ (46.99\%)\end{tabular} & \begin{tabular}{@{}c@{}}13.25\% \\ (12.05\%)\end{tabular} & \begin{tabular}{@{}c@{}}44.58\% \\ (27.71\%)\end{tabular} \\
\hline
\textbf{1} & \begin{tabular}{@{}c@{}}100.00\% \\ (100.00\%)\end{tabular} & - & \begin{tabular}{@{}c@{}}100.00\% \\ (100.00\%)\end{tabular} & \begin{tabular}{@{}c@{}}100.00\% \\ (100.00\%)\end{tabular} & \begin{tabular}{@{}c@{}}100.00\% \\ (100.00\%)\end{tabular} & \begin{tabular}{@{}c@{}}100.00\% \\ (100.00\%)\end{tabular} & \begin{tabular}{@{}c@{}}100.00\% \\ (100.00\%)\end{tabular} & \begin{tabular}{@{}c@{}}100.00\% \\ (100.00\%)\end{tabular} & \begin{tabular}{@{}c@{}}100.00\% \\ (100.00\%)\end{tabular} & \begin{tabular}{@{}c@{}}100.00\% \\ (96.80\%)\end{tabular} \\
\hline
\textbf{2} & \begin{tabular}{@{}c@{}}96.49\% \\ (96.49\%)\end{tabular} & \begin{tabular}{@{}c@{}}60.53\% \\ (59.65\%)\end{tabular} & - & \begin{tabular}{@{}c@{}}95.61\% \\ (95.61\%)\end{tabular} & \begin{tabular}{@{}c@{}}78.07\% \\ (75.44\%)\end{tabular} & \begin{tabular}{@{}c@{}}98.25\% \\ (71.05\%)\end{tabular} & \begin{tabular}{@{}c@{}}94.74\% \\ (90.35\%)\end{tabular} & \begin{tabular}{@{}c@{}}71.05\% \\ (69.30\%)\end{tabular} & \begin{tabular}{@{}c@{}}52.63\% \\ (50.88\%)\end{tabular} & \begin{tabular}{@{}c@{}}75.44\% \\ (42.98\%)\end{tabular} \\
\hline
\textbf{3} & \begin{tabular}{@{}c@{}}100.00\% \\ (100.00\%)\end{tabular} & \begin{tabular}{@{}c@{}}87.85\% \\ (66.36\%)\end{tabular} & \begin{tabular}{@{}c@{}}90.65\% \\ (90.65\%)\end{tabular} & - & \begin{tabular}{@{}c@{}}85.98\% \\ (73.83\%)\end{tabular} & \begin{tabular}{@{}c@{}}95.33\% \\ (95.33\%)\end{tabular} & \begin{tabular}{@{}c@{}}79.44\% \\ (53.27\%)\end{tabular} & \begin{tabular}{@{}c@{}}65.42\% \\ (64.49\%)\end{tabular} & \begin{tabular}{@{}c@{}}59.81\% \\ (46.73\%)\end{tabular} & \begin{tabular}{@{}c@{}}70.09\% \\ (58.88\%)\end{tabular} \\
\hline
\textbf{4} & \begin{tabular}{@{}c@{}}99.09\% \\ (99.09\%)\end{tabular} & \begin{tabular}{@{}c@{}}67.27\% \\ (66.36\%)\end{tabular} & \begin{tabular}{@{}c@{}}96.36\% \\ (96.36\%)\end{tabular} & \begin{tabular}{@{}c@{}}100.00\% \\ (81.82\%)\end{tabular} & - & \begin{tabular}{@{}c@{}}100.00\% \\ (98.18\%)\end{tabular} & \begin{tabular}{@{}c@{}}93.64\% \\ (93.64\%)\end{tabular} & \begin{tabular}{@{}c@{}}98.18\% \\ (95.45\%)\end{tabular} & \begin{tabular}{@{}c@{}}97.27\% \\ (92.73\%)\end{tabular} & \begin{tabular}{@{}c@{}}39.09\% \\ (39.09\%)\end{tabular} \\
\hline
\textbf{5} & \begin{tabular}{@{}c@{}}100.00\% \\ (100.00\%)\end{tabular} & \begin{tabular}{@{}c@{}}79.31\% \\ (51.72\%)\end{tabular} & \begin{tabular}{@{}c@{}}100.00\% \\ (83.91\%)\end{tabular} & \begin{tabular}{@{}c@{}}70.11\% \\ (70.11\%)\end{tabular} & \begin{tabular}{@{}c@{}}80.46\% \\ (72.41\%)\end{tabular} & - & \begin{tabular}{@{}c@{}}73.56\% \\ (73.56\%)\end{tabular} & \begin{tabular}{@{}c@{}}87.36\% \\ (73.56\%)\end{tabular} & \begin{tabular}{@{}c@{}}55.17\% \\ (52.87\%)\end{tabular} & \begin{tabular}{@{}c@{}}75.86\% \\ (65.52\%)\end{tabular} \\
\hline
\textbf{6} & \begin{tabular}{@{}c@{}}97.70\% \\ (97.70\%)\end{tabular} & \begin{tabular}{@{}c@{}}68.97\% \\ (50.57\%)\end{tabular} & \begin{tabular}{@{}c@{}}96.55\% \\ (96.55\%)\end{tabular} & \begin{tabular}{@{}c@{}}95.40\% \\ (71.26\%)\end{tabular} & \begin{tabular}{@{}c@{}}73.56\% \\ (73.56\%)\end{tabular} & \begin{tabular}{@{}c@{}}87.36\% \\ (77.01\%)\end{tabular} & - & \begin{tabular}{@{}c@{}}88.51\% \\ (72.41\%)\end{tabular} & \begin{tabular}{@{}c@{}}90.80\% \\ (55.17\%)\end{tabular} & \begin{tabular}{@{}c@{}}91.95\% \\ (35.63\%)\end{tabular} \\
\hline
\textbf{7} & \begin{tabular}{@{}c@{}}100.00\% \\ (97.98\%)\end{tabular} & \begin{tabular}{@{}c@{}}83.84\% \\ (83.84\%)\end{tabular} & \begin{tabular}{@{}c@{}}100.00\% \\ (100.00\%)\end{tabular} & \begin{tabular}{@{}c@{}}100.00\% \\ (100.00\%)\end{tabular} & \begin{tabular}{@{}c@{}}93.94\% \\ (90.91\%)\end{tabular} & \begin{tabular}{@{}c@{}}98.99\% \\ (96.97\%)\end{tabular} & \begin{tabular}{@{}c@{}}88.89\% \\ (81.82\%)\end{tabular} & - & \begin{tabular}{@{}c@{}}100.00\% \\ (86.87\%)\end{tabular} & \begin{tabular}{@{}c@{}}50.51\% \\ (50.51\%)\end{tabular} \\
\hline
\textbf{8} & \begin{tabular}{@{}c@{}}100.00\% \\ (100.00\%)\end{tabular} & \begin{tabular}{@{}c@{}}96.59\% \\ (78.41\%)\end{tabular} & \begin{tabular}{@{}c@{}}100.00\% \\ (100.00\%)\end{tabular} & \begin{tabular}{@{}c@{}}98.86\% \\ (95.45\%)\end{tabular} & \begin{tabular}{@{}c@{}}94.32\% \\ (86.36\%)\end{tabular} & \begin{tabular}{@{}c@{}}98.86\% \\ (98.86\%)\end{tabular} & \begin{tabular}{@{}c@{}}98.86\% \\ (93.18\%)\end{tabular} & \begin{tabular}{@{}c@{}}98.86\% \\ (73.86\%)\end{tabular} & - & \begin{tabular}{@{}c@{}}87.50\% \\ (78.41\%)\end{tabular} \\
\hline
\textbf{9} & \begin{tabular}{@{}c@{}}100.00\% \\ (100.00\%)\end{tabular} & \begin{tabular}{@{}c@{}}100.00\% \\ (76.09\%)\end{tabular} & \begin{tabular}{@{}c@{}}100.00\% \\ (100.00\%)\end{tabular} & \begin{tabular}{@{}c@{}}98.91\% \\ (96.74\%)\end{tabular} & \begin{tabular}{@{}c@{}}100.00\% \\ (100.00\%)\end{tabular} & \begin{tabular}{@{}c@{}}100.00\% \\ (98.91\%)\end{tabular} & \begin{tabular}{@{}c@{}}97.83\% \\ (97.83\%)\end{tabular} & \begin{tabular}{@{}c@{}}98.91\% \\ (98.91\%)\end{tabular} & \begin{tabular}{@{}c@{}}97.83\% \\ (94.57\%)\end{tabular} & - \\
\hline
\end{tabular}
\end{center}
}
\caption{\textbf{$L_2$ Optimization Latent Attack (mean latent vector target):} $\ASuntargeted$\ ($\AStargeted$\ in parentheses) for all source and target class pairs using adversarial examples generated on the VAE-GAN model.
The mean latent vector for each target class (over all of the images of that class in the training set) is used as the target latent vector.
Higher values indicate more successful attacks against the generative model.}
\label{tab:direct-distribution-classification-accuracies}
\end{table}


\begin{figure}[h]
\begin{center}
\includegraphics[scale=0.3]{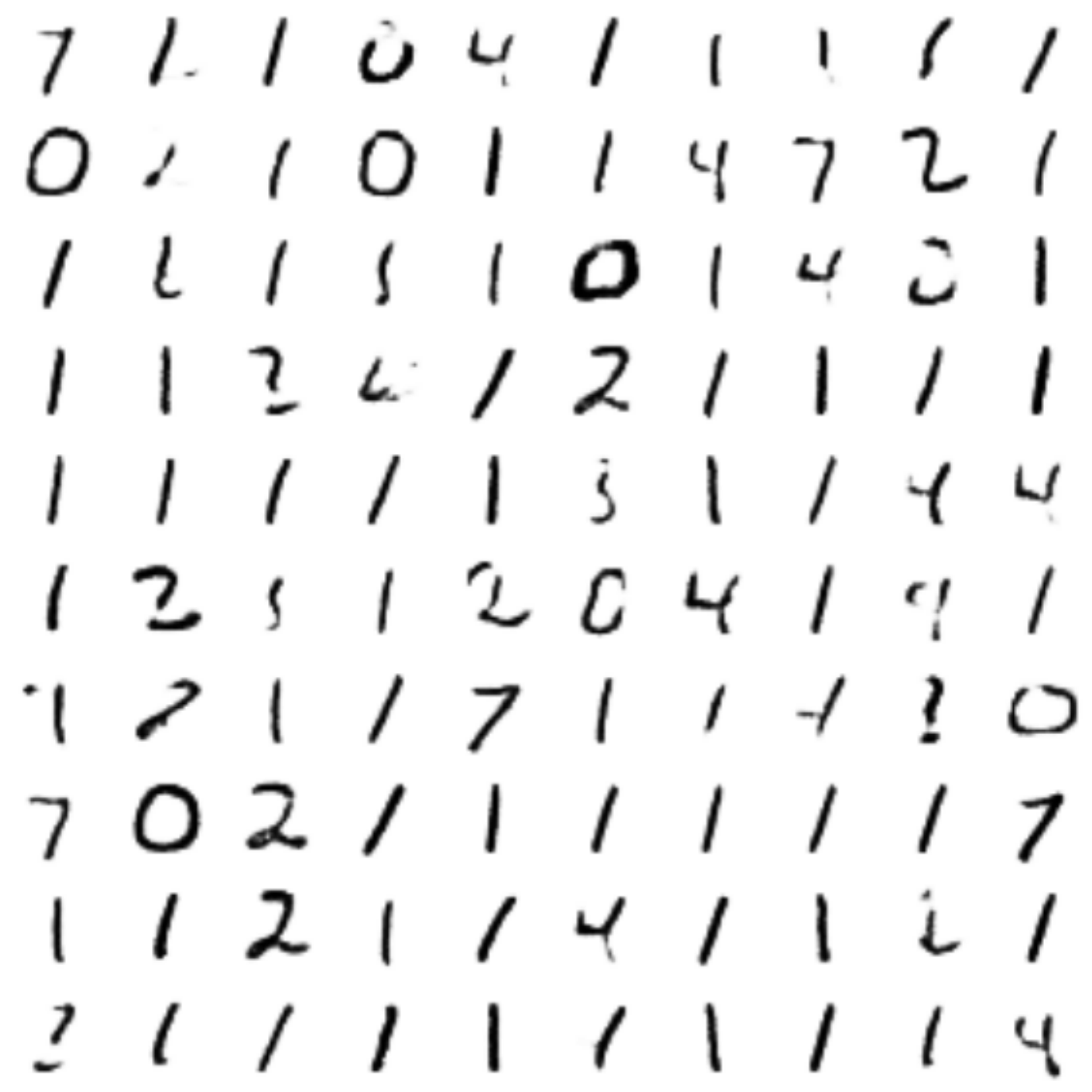}
\includegraphics[scale=0.3]{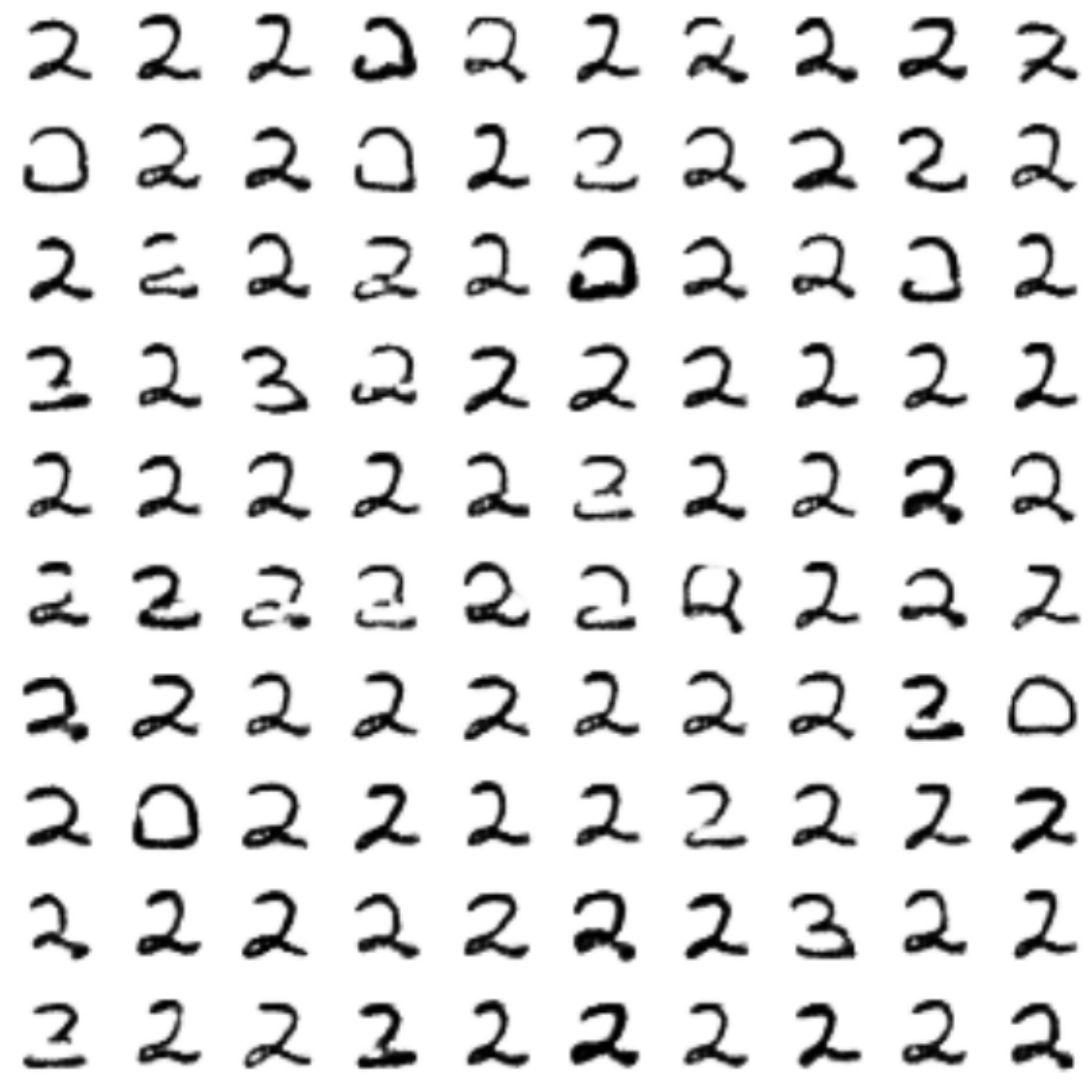}
\includegraphics[scale=0.3]{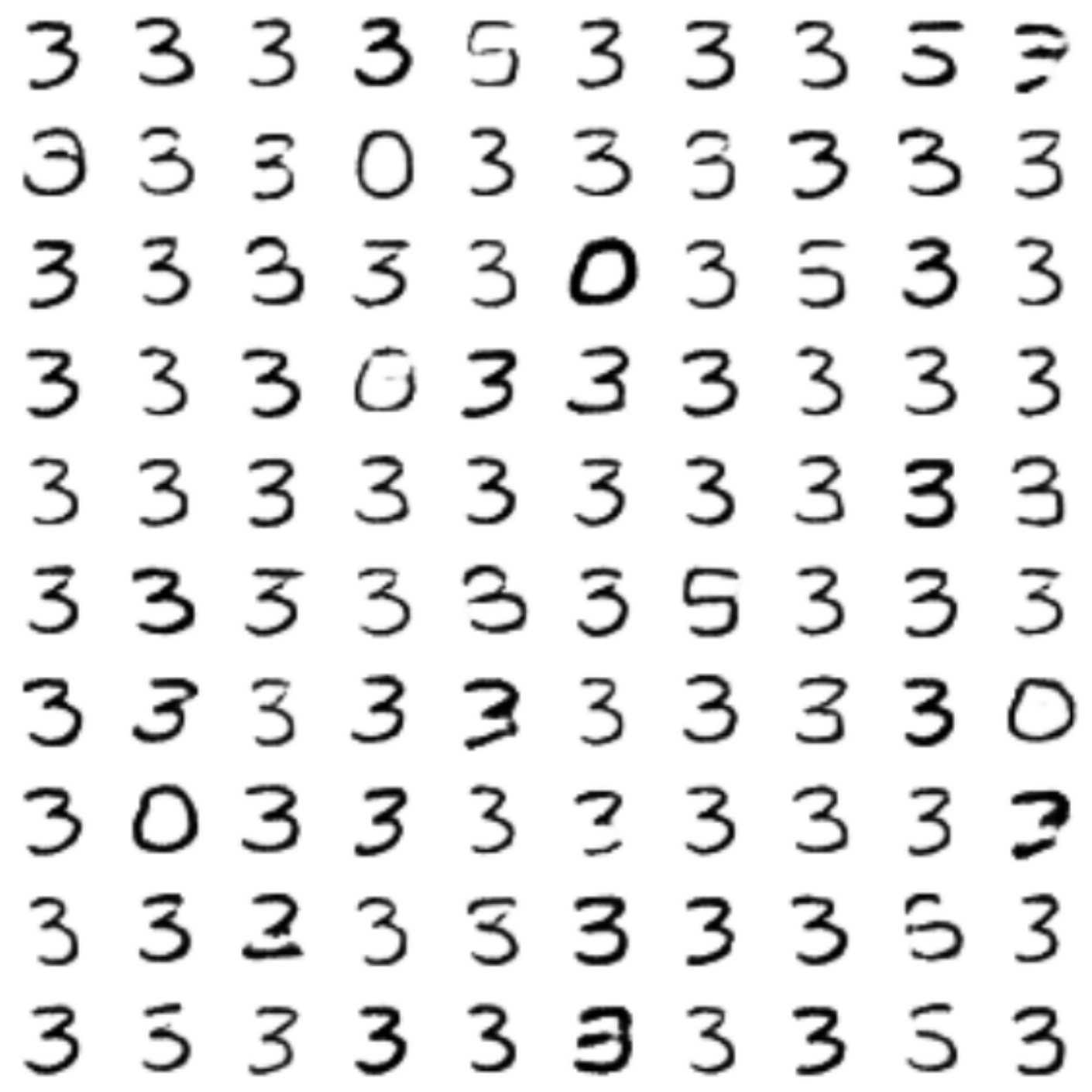} \\
\includegraphics[scale=0.3]{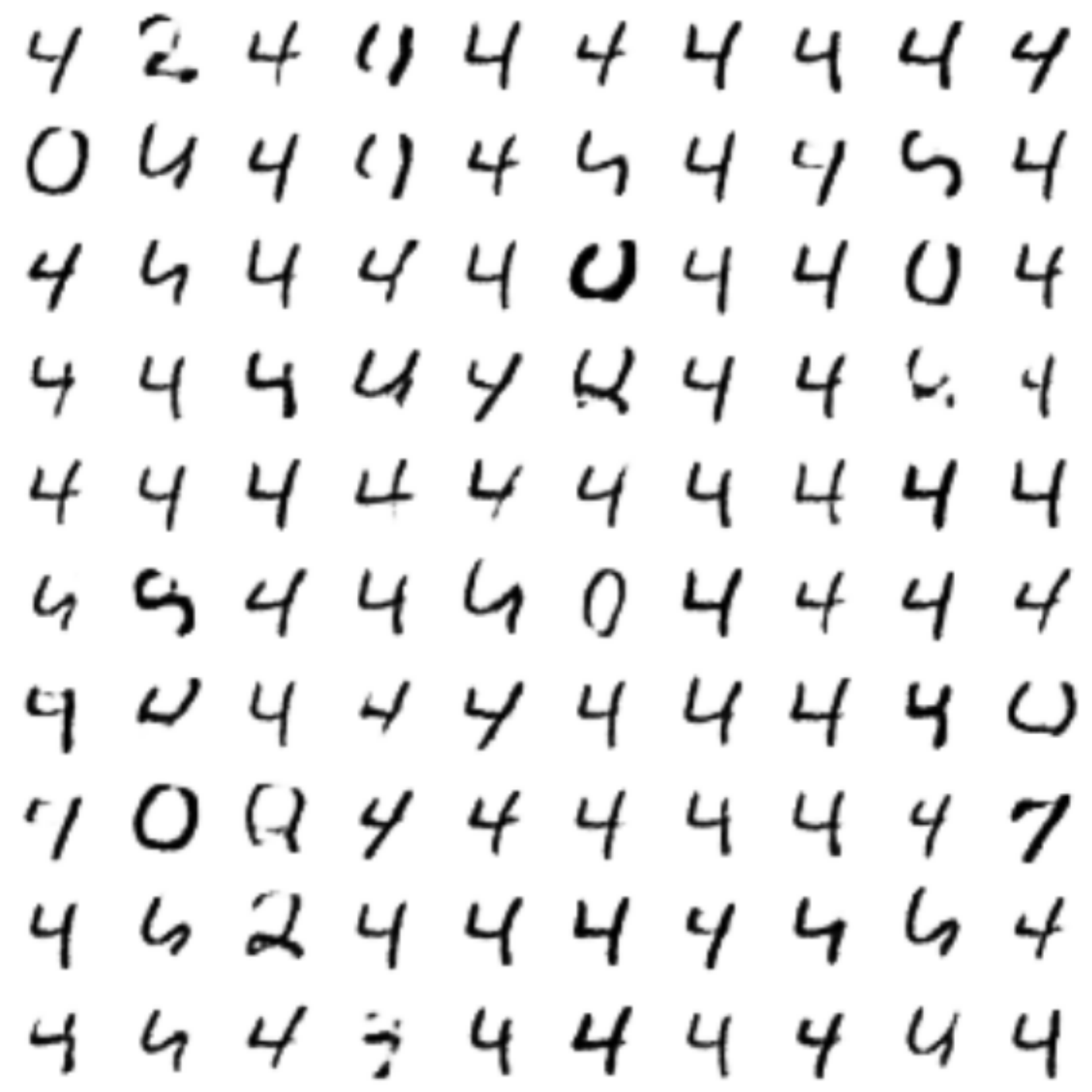}
\includegraphics[scale=0.3]{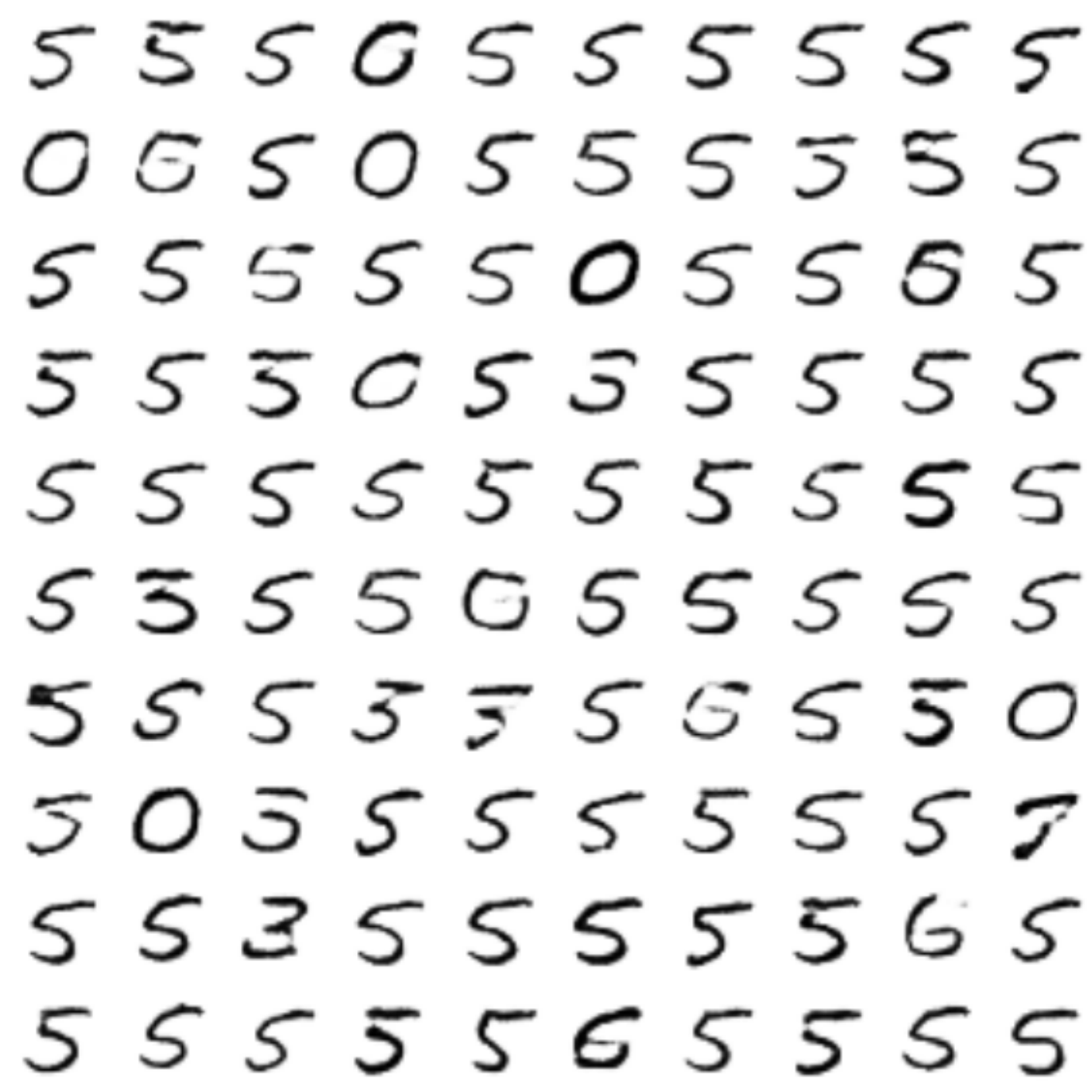}
\includegraphics[scale=0.3]{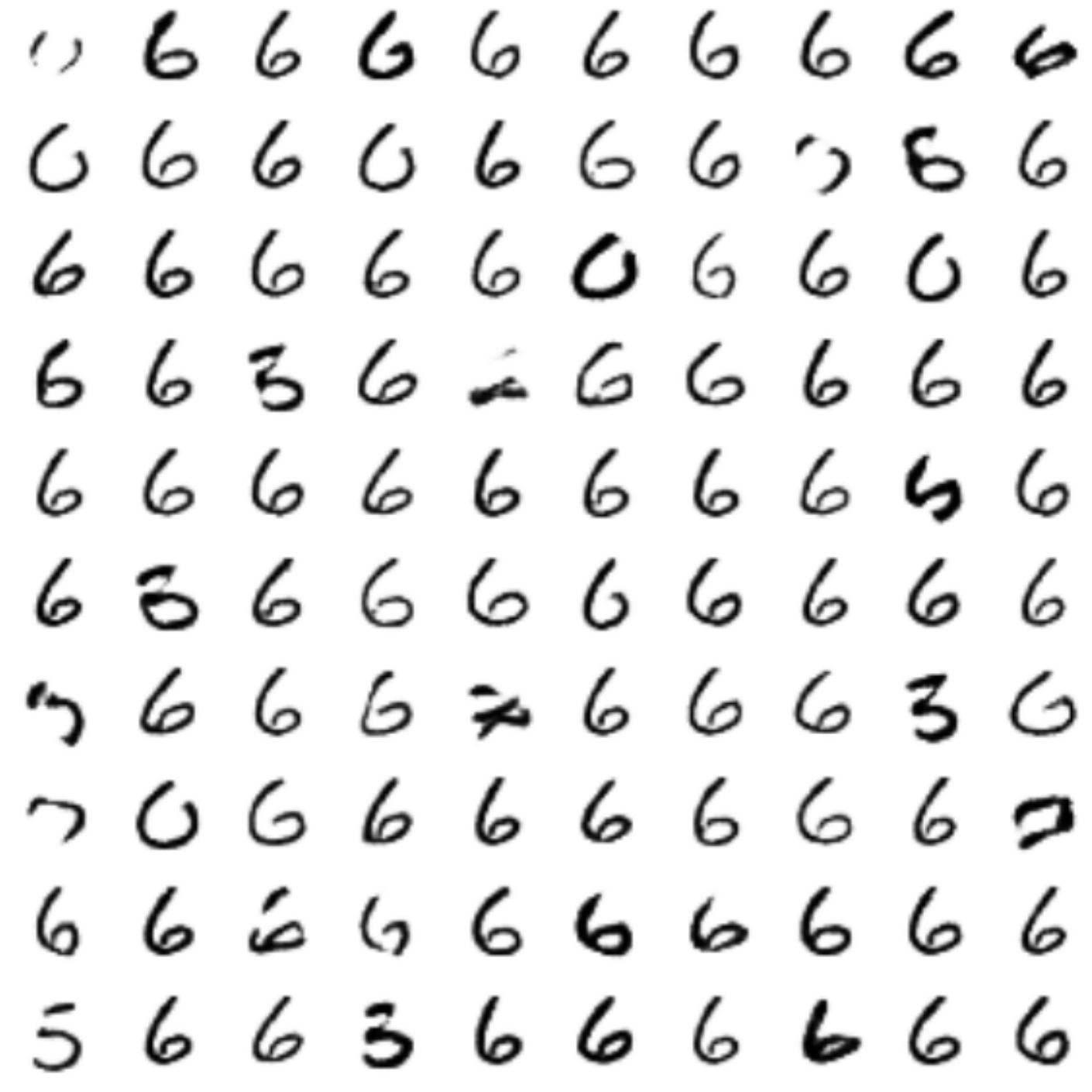} \\
\includegraphics[scale=0.3]{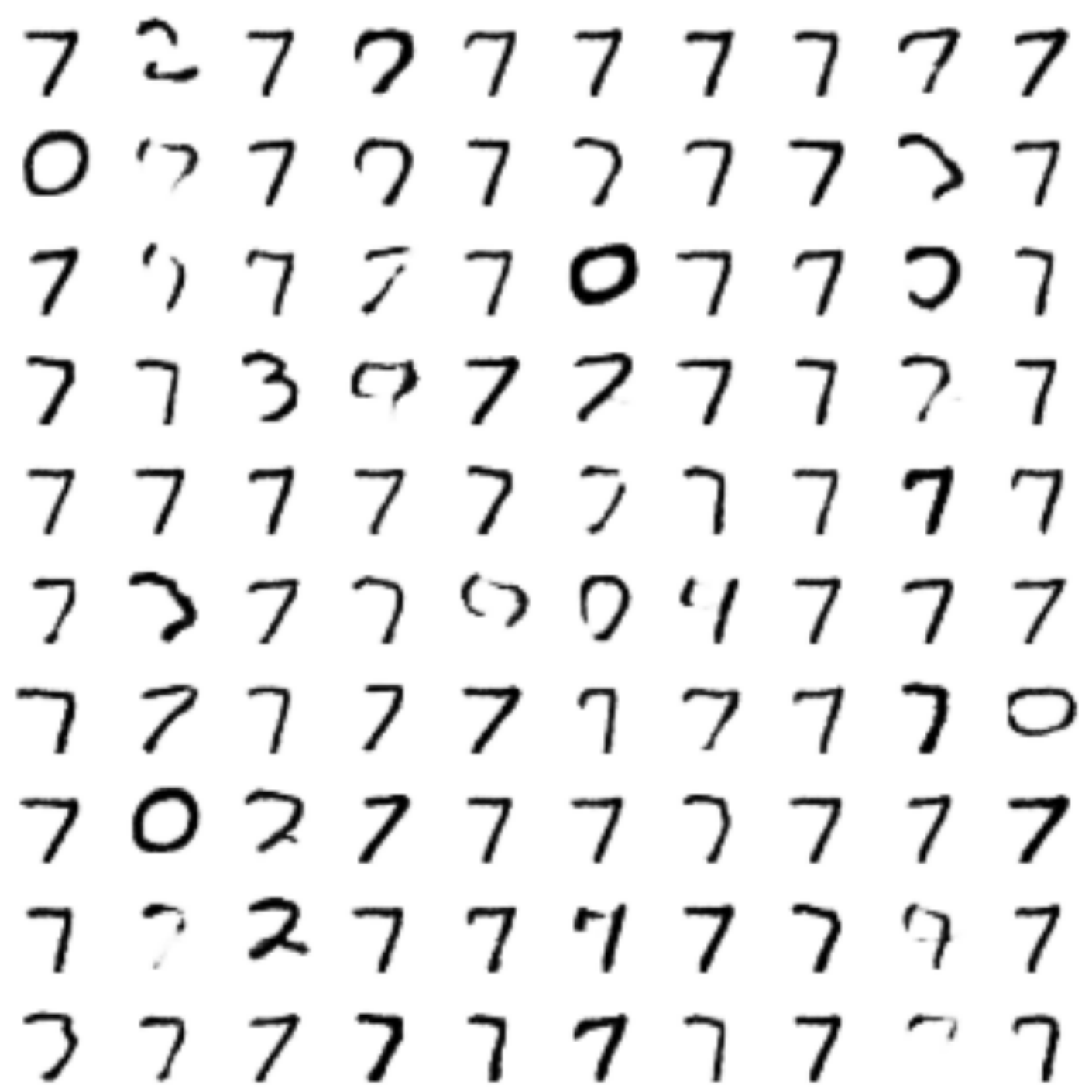}
\includegraphics[scale=0.3]{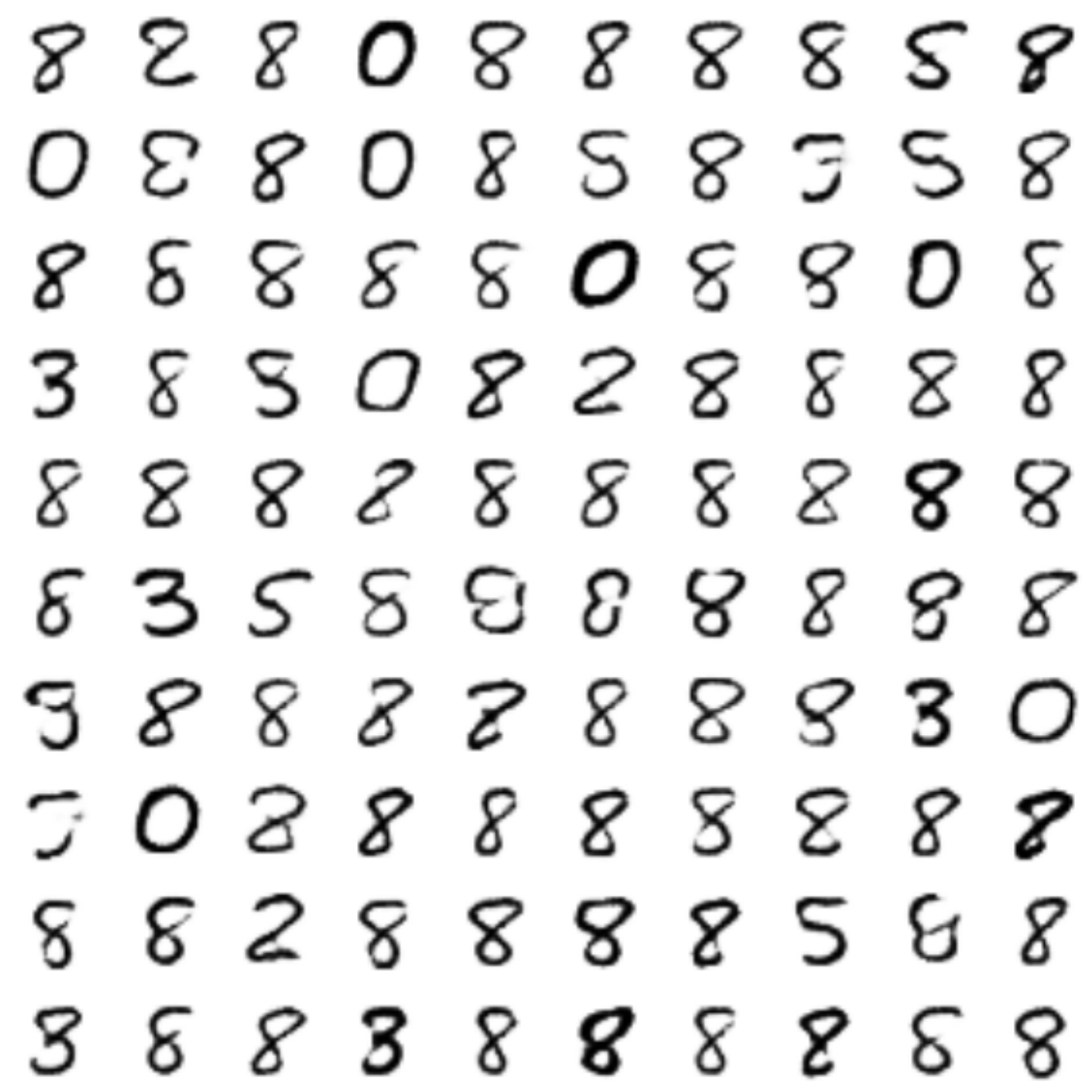}
\includegraphics[scale=0.3]{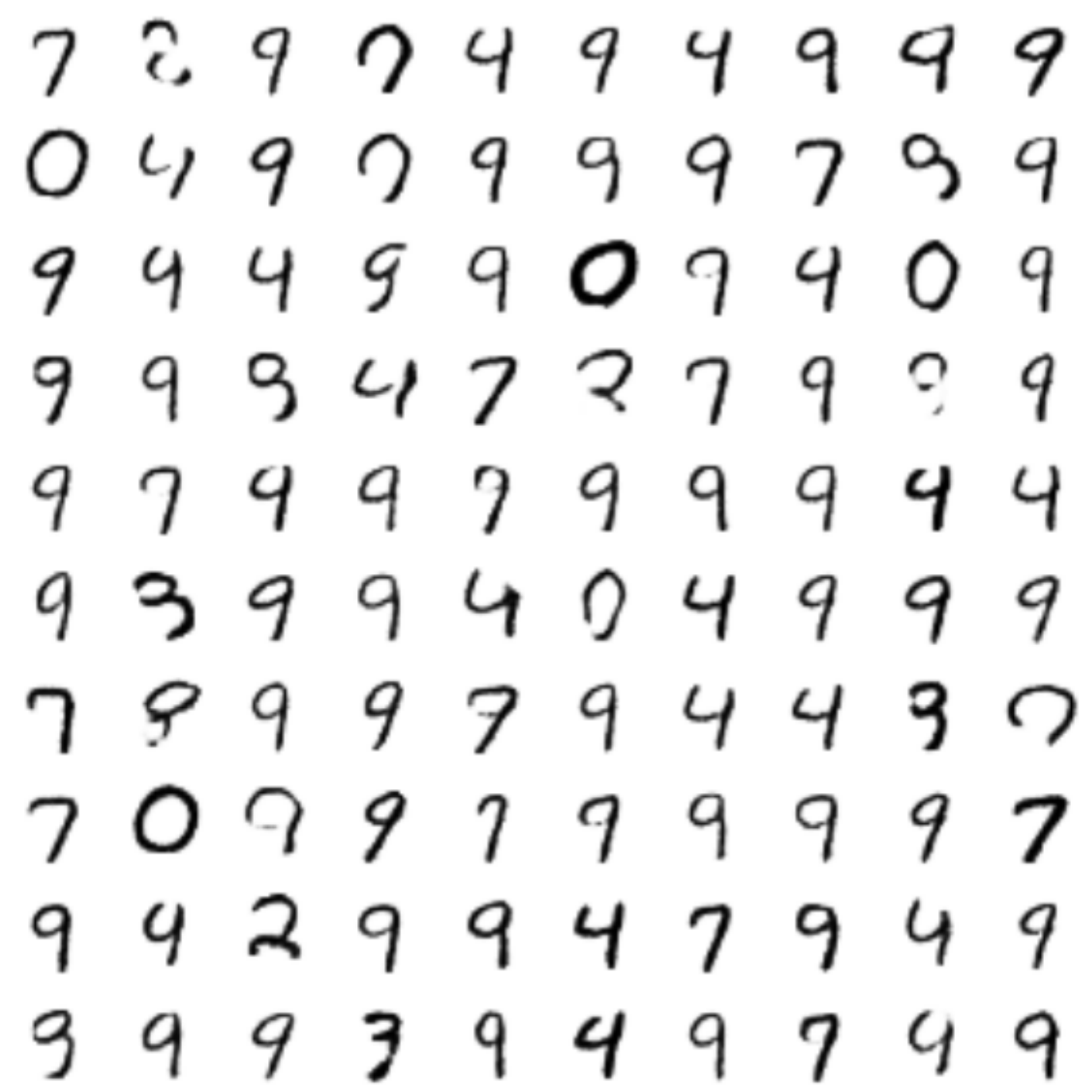}
\end{center}
\caption{\textbf{$L_2$ Optimization Latent Attack (mean latent vector targets):} VAE-GAN reconstructions of adversarial examples with target classes from $1$ through $9$.
Original examples which already belong to the target class are excluded.}
\label{fig:direct-targeted-reconstructions-1-9}
\end{figure}

\begin{figure}[h]
\begin{center}
\includegraphics[scale=0.47]{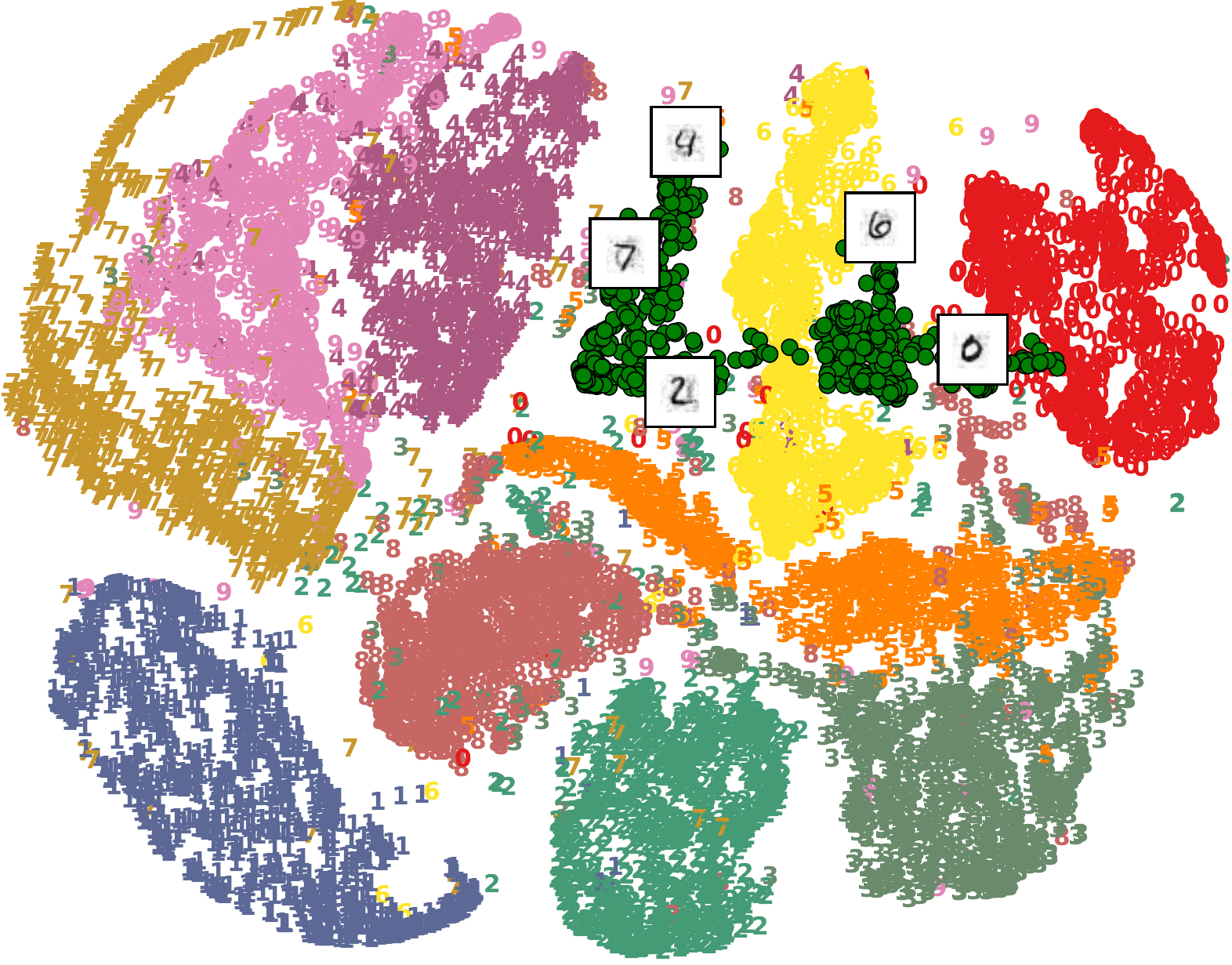}
\end{center}
\caption{\textbf{$L_2$ Optimization Latent Attack (single latent vector target):} t-SNE plot of the latent space, with the addition of green circles representing the adversarial examples for target class $0$.
In this plot, it appears that the adversarial examples cluster around $6$ (yellow) and $0$ (red).}
\label{fig:direct-targeted-random-reconstructions-t-sne}
\end{figure}

\begin{table}[h]
{
\tiny
\begin{center}
\begin{tabular}{|c|c|c|c|c|c|c|c|c|c|c|}
\hline
\textbf{Source} & \textbf{Target 0} & \textbf{Target 1} & \textbf{Target 2} & \textbf{Target 3} & \textbf{Target 4} & \textbf{Target 5} & \textbf{Target 6} & \textbf{Target 7} & \textbf{Target 8} & \textbf{Target 9} \\
\hline
\textbf{0} & - & \begin{tabular}{@{}c@{}}95.18\% \\ (9.64\%)\end{tabular} & \begin{tabular}{@{}c@{}}100.00\% \\ (100.00\%)\end{tabular} & \begin{tabular}{@{}c@{}}98.80\% \\ (93.98\%)\end{tabular} & \begin{tabular}{@{}c@{}}100.00\% \\ (48.19\%)\end{tabular} & \begin{tabular}{@{}c@{}}91.57\% \\ (89.16\%)\end{tabular} & \begin{tabular}{@{}c@{}}100.00\% \\ (89.16\%)\end{tabular} & \begin{tabular}{@{}c@{}}73.49\% \\ (43.37\%)\end{tabular} & \begin{tabular}{@{}c@{}}100.00\% \\ (87.95\%)\end{tabular} & \begin{tabular}{@{}c@{}}100.00\% \\ (25.30\%)\end{tabular} \\
\hline
\textbf{1} & \begin{tabular}{@{}c@{}}100.00\% \\ (100.00\%)\end{tabular} & - & \begin{tabular}{@{}c@{}}100.00\% \\ (100.00\%)\end{tabular} & \begin{tabular}{@{}c@{}}100.00\% \\ (100.00\%)\end{tabular} & \begin{tabular}{@{}c@{}}100.00\% \\ (92.80\%)\end{tabular} & \begin{tabular}{@{}c@{}}100.00\% \\ (97.60\%)\end{tabular} & \begin{tabular}{@{}c@{}}100.00\% \\ (98.40\%)\end{tabular} & \begin{tabular}{@{}c@{}}100.00\% \\ (76.00\%)\end{tabular} & \begin{tabular}{@{}c@{}}100.00\% \\ (100.00\%)\end{tabular} & \begin{tabular}{@{}c@{}}100.00\% \\ (90.40\%)\end{tabular} \\
\hline
\textbf{2} & \begin{tabular}{@{}c@{}}98.25\% \\ (98.25\%)\end{tabular} & \begin{tabular}{@{}c@{}}83.33\% \\ (48.25\%)\end{tabular} & - & \begin{tabular}{@{}c@{}}100.00\% \\ (100.00\%)\end{tabular} & \begin{tabular}{@{}c@{}}88.60\% \\ (43.86\%)\end{tabular} & \begin{tabular}{@{}c@{}}99.12\% \\ (63.16\%)\end{tabular} & \begin{tabular}{@{}c@{}}74.56\% \\ (71.93\%)\end{tabular} & \begin{tabular}{@{}c@{}}99.12\% \\ (63.16\%)\end{tabular} & \begin{tabular}{@{}c@{}}93.86\% \\ (92.98\%)\end{tabular} & \begin{tabular}{@{}c@{}}99.12\% \\ (21.05\%)\end{tabular} \\
\hline
\textbf{3} & \begin{tabular}{@{}c@{}}99.07\% \\ (98.13\%)\end{tabular} & \begin{tabular}{@{}c@{}}57.01\% \\ (42.99\%)\end{tabular} & \begin{tabular}{@{}c@{}}99.07\% \\ (99.07\%)\end{tabular} & - & \begin{tabular}{@{}c@{}}82.24\% \\ (36.45\%)\end{tabular} & \begin{tabular}{@{}c@{}}89.72\% \\ (88.79\%)\end{tabular} & \begin{tabular}{@{}c@{}}99.07\% \\ (61.68\%)\end{tabular} & \begin{tabular}{@{}c@{}}57.01\% \\ (37.38\%)\end{tabular} & \begin{tabular}{@{}c@{}}98.13\% \\ (92.52\%)\end{tabular} & \begin{tabular}{@{}c@{}}67.29\% \\ (18.69\%)\end{tabular} \\
\hline
\textbf{4} & \begin{tabular}{@{}c@{}}100.00\% \\ (100.00\%)\end{tabular} & \begin{tabular}{@{}c@{}}100.00\% \\ (37.27\%)\end{tabular} & \begin{tabular}{@{}c@{}}100.00\% \\ (100.00\%)\end{tabular} & \begin{tabular}{@{}c@{}}100.00\% \\ (99.09\%)\end{tabular} & - & \begin{tabular}{@{}c@{}}100.00\% \\ (80.00\%)\end{tabular} & \begin{tabular}{@{}c@{}}98.18\% \\ (93.64\%)\end{tabular} & \begin{tabular}{@{}c@{}}100.00\% \\ (94.55\%)\end{tabular} & \begin{tabular}{@{}c@{}}100.00\% \\ (99.09\%)\end{tabular} & \begin{tabular}{@{}c@{}}86.36\% \\ (80.00\%)\end{tabular} \\
\hline
\textbf{5} & \begin{tabular}{@{}c@{}}100.00\% \\ (100.00\%)\end{tabular} & \begin{tabular}{@{}c@{}}97.70\% \\ (19.54\%)\end{tabular} & \begin{tabular}{@{}c@{}}100.00\% \\ (98.85\%)\end{tabular} & \begin{tabular}{@{}c@{}}98.85\% \\ (98.85\%)\end{tabular} & \begin{tabular}{@{}c@{}}85.06\% \\ (44.83\%)\end{tabular} & - & \begin{tabular}{@{}c@{}}95.40\% \\ (88.51\%)\end{tabular} & \begin{tabular}{@{}c@{}}93.10\% \\ (45.98\%)\end{tabular} & \begin{tabular}{@{}c@{}}96.55\% \\ (96.55\%)\end{tabular} & \begin{tabular}{@{}c@{}}87.36\% \\ (34.48\%)\end{tabular} \\
\hline
\textbf{6} & \begin{tabular}{@{}c@{}}100.00\% \\ (100.00\%)\end{tabular} & \begin{tabular}{@{}c@{}}96.55\% \\ (58.62\%)\end{tabular} & \begin{tabular}{@{}c@{}}100.00\% \\ (98.85\%)\end{tabular} & \begin{tabular}{@{}c@{}}100.00\% \\ (98.85\%)\end{tabular} & \begin{tabular}{@{}c@{}}100.00\% \\ (86.21\%)\end{tabular} & \begin{tabular}{@{}c@{}}100.00\% \\ (97.70\%)\end{tabular} & - & \begin{tabular}{@{}c@{}}100.00\% \\ (56.32\%)\end{tabular} & \begin{tabular}{@{}c@{}}100.00\% \\ (96.55\%)\end{tabular} & \begin{tabular}{@{}c@{}}95.40\% \\ (43.68\%)\end{tabular} \\
\hline
\textbf{7} & \begin{tabular}{@{}c@{}}100.00\% \\ (100.00\%)\end{tabular} & \begin{tabular}{@{}c@{}}80.81\% \\ (40.40\%)\end{tabular} & \begin{tabular}{@{}c@{}}100.00\% \\ (100.00\%)\end{tabular} & \begin{tabular}{@{}c@{}}100.00\% \\ (98.99\%)\end{tabular} & \begin{tabular}{@{}c@{}}100.00\% \\ (92.93\%)\end{tabular} & \begin{tabular}{@{}c@{}}100.00\% \\ (87.88\%)\end{tabular} & \begin{tabular}{@{}c@{}}100.00\% \\ (62.63\%)\end{tabular} & - & \begin{tabular}{@{}c@{}}100.00\% \\ (97.98\%)\end{tabular} & \begin{tabular}{@{}c@{}}100.00\% \\ (88.89\%)\end{tabular} \\
\hline
\textbf{8} & \begin{tabular}{@{}c@{}}100.00\% \\ (100.00\%)\end{tabular} & \begin{tabular}{@{}c@{}}44.32\% \\ (18.18\%)\end{tabular} & \begin{tabular}{@{}c@{}}100.00\% \\ (100.00\%)\end{tabular} & \begin{tabular}{@{}c@{}}100.00\% \\ (100.00\%)\end{tabular} & \begin{tabular}{@{}c@{}}30.68\% \\ (28.41\%)\end{tabular} & \begin{tabular}{@{}c@{}}78.41\% \\ (76.14\%)\end{tabular} & \begin{tabular}{@{}c@{}}89.77\% \\ (81.82\%)\end{tabular} & \begin{tabular}{@{}c@{}}75.00\% \\ (38.64\%)\end{tabular} & - & \begin{tabular}{@{}c@{}}22.73\% \\ (15.91\%)\end{tabular} \\
\hline
\textbf{9} & \begin{tabular}{@{}c@{}}100.00\% \\ (100.00\%)\end{tabular} & \begin{tabular}{@{}c@{}}98.91\% \\ (17.39\%)\end{tabular} & \begin{tabular}{@{}c@{}}100.00\% \\ (100.00\%)\end{tabular} & \begin{tabular}{@{}c@{}}100.00\% \\ (100.00\%)\end{tabular} & \begin{tabular}{@{}c@{}}97.83\% \\ (92.39\%)\end{tabular} & \begin{tabular}{@{}c@{}}100.00\% \\ (89.13\%)\end{tabular} & \begin{tabular}{@{}c@{}}100.00\% \\ (92.39\%)\end{tabular} & \begin{tabular}{@{}c@{}}98.91\% \\ (94.57\%)\end{tabular} & \begin{tabular}{@{}c@{}}100.00\% \\ (100.00\%)\end{tabular} & - \\
\hline
\end{tabular}
\end{center}
}
\caption{\textbf{$L_2$ Optimization $\calL_{\VAE}$ Attack (mean reconstruction target):} $\ASuntargeted$\ ($\AStargeted$\ in parentheses) for all source and target class pairs using adversarial examples generated on the VAE-GAN model.
The mean image for each target class (over all of the images of that class in the training set) is used as the target.
Higher values indicate more successful attacks against the generative model.}
\end{table}

\begin{table}[h]
{
\tiny
\begin{center}
\begin{tabular}{|c|c|c|c|c|c|c|c|c|c|c|}
\hline
\textbf{Source} & \textbf{Target 0} & \textbf{Target 1} & \textbf{Target 2} & \textbf{Target 3} & \textbf{Target 4} & \textbf{Target 5} & \textbf{Target 6} & \textbf{Target 7} & \textbf{Target 8} & \textbf{Target 9} \\
\hline
\textbf{0} & - & \begin{tabular}{@{}c@{}}92.77\% \\ (38.55\%)\end{tabular} & \begin{tabular}{@{}c@{}}100.00\% \\ (100.00\%)\end{tabular} & \begin{tabular}{@{}c@{}}100.00\% \\ (66.27\%)\end{tabular} & \begin{tabular}{@{}c@{}}100.00\% \\ (34.94\%)\end{tabular} & \begin{tabular}{@{}c@{}}100.00\% \\ (22.89\%)\end{tabular} & \begin{tabular}{@{}c@{}}100.00\% \\ (100.00\%)\end{tabular} & \begin{tabular}{@{}c@{}}79.52\% \\ (63.86\%)\end{tabular} & \begin{tabular}{@{}c@{}}97.59\% \\ (90.36\%)\end{tabular} & \begin{tabular}{@{}c@{}}100.00\% \\ (62.65\%)\end{tabular} \\
\hline
\textbf{1} & \begin{tabular}{@{}c@{}}100.00\% \\ (100.00\%)\end{tabular} & - & \begin{tabular}{@{}c@{}}100.00\% \\ (100.00\%)\end{tabular} & \begin{tabular}{@{}c@{}}100.00\% \\ (99.20\%)\end{tabular} & \begin{tabular}{@{}c@{}}100.00\% \\ (90.40\%)\end{tabular} & \begin{tabular}{@{}c@{}}100.00\% \\ (0.80\%)\end{tabular} & \begin{tabular}{@{}c@{}}100.00\% \\ (100.00\%)\end{tabular} & \begin{tabular}{@{}c@{}}100.00\% \\ (100.00\%)\end{tabular} & \begin{tabular}{@{}c@{}}100.00\% \\ (100.00\%)\end{tabular} & \begin{tabular}{@{}c@{}}100.00\% \\ (100.00\%)\end{tabular} \\
\hline
\textbf{2} & \begin{tabular}{@{}c@{}}97.37\% \\ (97.37\%)\end{tabular} & \begin{tabular}{@{}c@{}}97.37\% \\ (57.02\%)\end{tabular} & - & \begin{tabular}{@{}c@{}}100.00\% \\ (87.72\%)\end{tabular} & \begin{tabular}{@{}c@{}}98.25\% \\ (42.11\%)\end{tabular} & \begin{tabular}{@{}c@{}}100.00\% \\ (50.88\%)\end{tabular} & \begin{tabular}{@{}c@{}}100.00\% \\ (99.12\%)\end{tabular} & \begin{tabular}{@{}c@{}}97.37\% \\ (89.47\%)\end{tabular} & \begin{tabular}{@{}c@{}}89.47\% \\ (89.47\%)\end{tabular} & \begin{tabular}{@{}c@{}}100.00\% \\ (81.58\%)\end{tabular} \\
\hline
\textbf{3} & \begin{tabular}{@{}c@{}}100.00\% \\ (100.00\%)\end{tabular} & \begin{tabular}{@{}c@{}}89.72\% \\ (85.05\%)\end{tabular} & \begin{tabular}{@{}c@{}}100.00\% \\ (100.00\%)\end{tabular} & - & \begin{tabular}{@{}c@{}}62.62\% \\ (48.60\%)\end{tabular} & \begin{tabular}{@{}c@{}}91.59\% \\ (45.79\%)\end{tabular} & \begin{tabular}{@{}c@{}}100.00\% \\ (99.07\%)\end{tabular} & \begin{tabular}{@{}c@{}}95.33\% \\ (90.65\%)\end{tabular} & \begin{tabular}{@{}c@{}}97.20\% \\ (94.39\%)\end{tabular} & \begin{tabular}{@{}c@{}}90.65\% \\ (79.44\%)\end{tabular} \\
\hline
\textbf{4} & \begin{tabular}{@{}c@{}}100.00\% \\ (100.00\%)\end{tabular} & \begin{tabular}{@{}c@{}}95.45\% \\ (67.27\%)\end{tabular} & \begin{tabular}{@{}c@{}}100.00\% \\ (100.00\%)\end{tabular} & \begin{tabular}{@{}c@{}}100.00\% \\ (73.64\%)\end{tabular} & - & \begin{tabular}{@{}c@{}}100.00\% \\ (30.00\%)\end{tabular} & \begin{tabular}{@{}c@{}}100.00\% \\ (100.00\%)\end{tabular} & \begin{tabular}{@{}c@{}}100.00\% \\ (99.09\%)\end{tabular} & \begin{tabular}{@{}c@{}}100.00\% \\ (99.09\%)\end{tabular} & \begin{tabular}{@{}c@{}}99.09\% \\ (99.09\%)\end{tabular} \\
\hline
\textbf{5} & \begin{tabular}{@{}c@{}}100.00\% \\ (100.00\%)\end{tabular} & \begin{tabular}{@{}c@{}}98.85\% \\ (79.31\%)\end{tabular} & \begin{tabular}{@{}c@{}}100.00\% \\ (100.00\%)\end{tabular} & \begin{tabular}{@{}c@{}}73.56\% \\ (73.56\%)\end{tabular} & \begin{tabular}{@{}c@{}}83.91\% \\ (34.48\%)\end{tabular} & - & \begin{tabular}{@{}c@{}}100.00\% \\ (100.00\%)\end{tabular} & \begin{tabular}{@{}c@{}}90.80\% \\ (87.36\%)\end{tabular} & \begin{tabular}{@{}c@{}}100.00\% \\ (100.00\%)\end{tabular} & \begin{tabular}{@{}c@{}}87.36\% \\ (82.76\%)\end{tabular} \\
\hline
\textbf{6} & \begin{tabular}{@{}c@{}}100.00\% \\ (100.00\%)\end{tabular} & \begin{tabular}{@{}c@{}}86.21\% \\ (79.31\%)\end{tabular} & \begin{tabular}{@{}c@{}}100.00\% \\ (100.00\%)\end{tabular} & \begin{tabular}{@{}c@{}}100.00\% \\ (88.51\%)\end{tabular} & \begin{tabular}{@{}c@{}}95.40\% \\ (71.26\%)\end{tabular} & \begin{tabular}{@{}c@{}}10.34\% \\ (10.34\%)\end{tabular} & - & \begin{tabular}{@{}c@{}}100.00\% \\ (83.91\%)\end{tabular} & \begin{tabular}{@{}c@{}}100.00\% \\ (97.70\%)\end{tabular} & \begin{tabular}{@{}c@{}}100.00\% \\ (70.11\%)\end{tabular} \\
\hline
\textbf{7} & \begin{tabular}{@{}c@{}}100.00\% \\ (100.00\%)\end{tabular} & \begin{tabular}{@{}c@{}}91.92\% \\ (79.80\%)\end{tabular} & \begin{tabular}{@{}c@{}}100.00\% \\ (100.00\%)\end{tabular} & \begin{tabular}{@{}c@{}}100.00\% \\ (87.88\%)\end{tabular} & \begin{tabular}{@{}c@{}}100.00\% \\ (63.64\%)\end{tabular} & \begin{tabular}{@{}c@{}}100.00\% \\ (58.59\%)\end{tabular} & \begin{tabular}{@{}c@{}}100.00\% \\ (100.00\%)\end{tabular} & - & \begin{tabular}{@{}c@{}}100.00\% \\ (100.00\%)\end{tabular} & \begin{tabular}{@{}c@{}}100.00\% \\ (100.00\%)\end{tabular} \\
\hline
\textbf{8} & \begin{tabular}{@{}c@{}}100.00\% \\ (100.00\%)\end{tabular} & \begin{tabular}{@{}c@{}}88.64\% \\ (73.86\%)\end{tabular} & \begin{tabular}{@{}c@{}}100.00\% \\ (100.00\%)\end{tabular} & \begin{tabular}{@{}c@{}}100.00\% \\ (46.59\%)\end{tabular} & \begin{tabular}{@{}c@{}}95.45\% \\ (44.32\%)\end{tabular} & \begin{tabular}{@{}c@{}}96.59\% \\ (31.82\%)\end{tabular} & \begin{tabular}{@{}c@{}}100.00\% \\ (100.00\%)\end{tabular} & \begin{tabular}{@{}c@{}}96.59\% \\ (94.32\%)\end{tabular} & - & \begin{tabular}{@{}c@{}}95.45\% \\ (79.55\%)\end{tabular} \\
\hline
\textbf{9} & \begin{tabular}{@{}c@{}}100.00\% \\ (100.00\%)\end{tabular} & \begin{tabular}{@{}c@{}}96.74\% \\ (72.83\%)\end{tabular} & \begin{tabular}{@{}c@{}}100.00\% \\ (100.00\%)\end{tabular} & \begin{tabular}{@{}c@{}}100.00\% \\ (59.78\%)\end{tabular} & \begin{tabular}{@{}c@{}}66.30\% \\ (63.04\%)\end{tabular} & \begin{tabular}{@{}c@{}}100.00\% \\ (28.26\%)\end{tabular} & \begin{tabular}{@{}c@{}}100.00\% \\ (100.00\%)\end{tabular} & \begin{tabular}{@{}c@{}}98.91\% \\ (98.91\%)\end{tabular} & \begin{tabular}{@{}c@{}}100.00\% \\ (100.00\%)\end{tabular} & - \\
\hline
\end{tabular}
\end{center}
}
\caption{\textbf{$L_2$ Optimization $\calL_{\VAE}$ Attack on MNIST (single image target):} $\ASuntargeted$\ ($\AStargeted$\ in parentheses) for different source and target class pairs using adversarial examples generated on the VAE-GAN model.
Higher values indicate more successful attacks against the generative model.}
\end{table}


\begin{figure}[h]
\begin{center}
\includegraphics[scale=0.4]{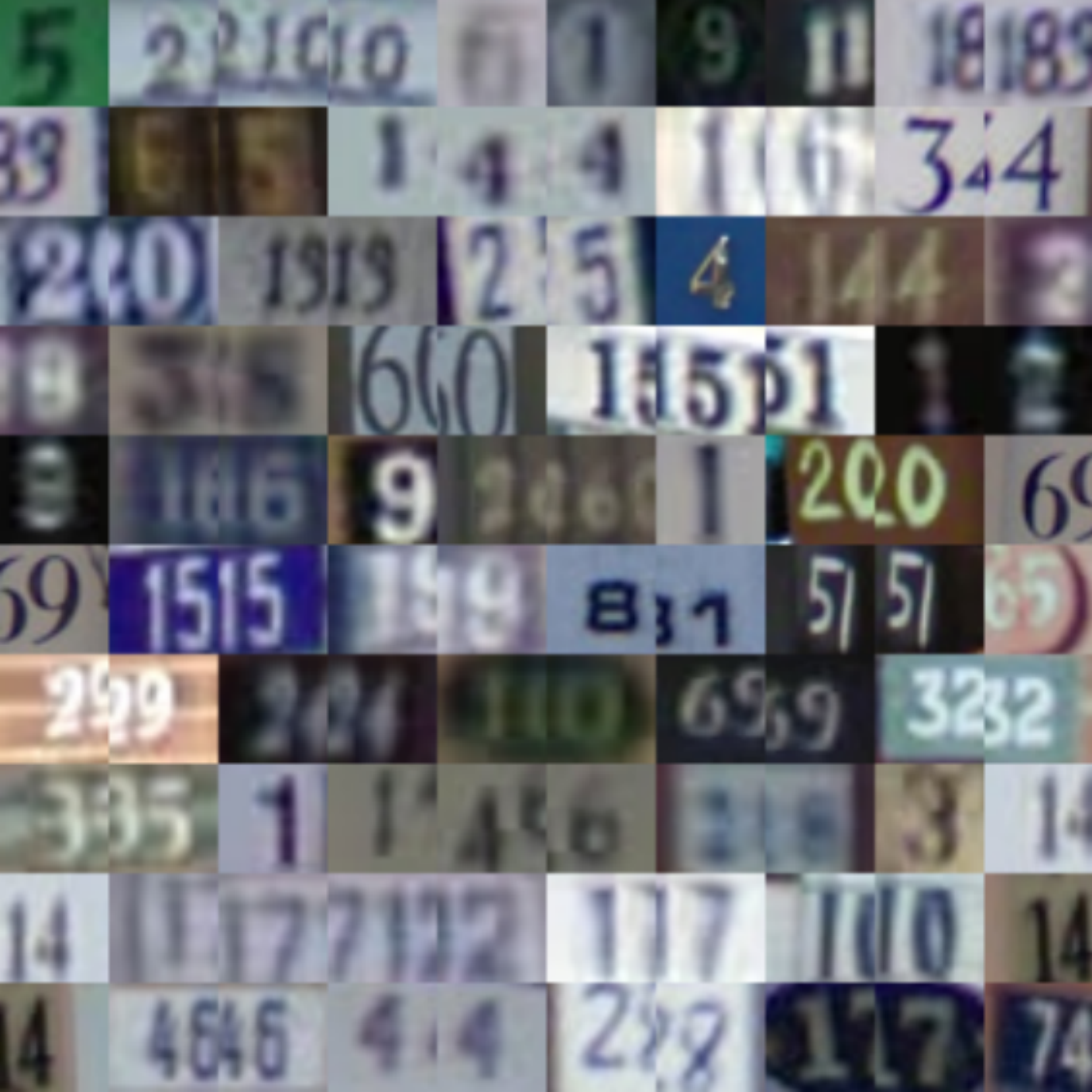}
\rule{0.4pt}{5.7cm}
\includegraphics[scale=0.4]{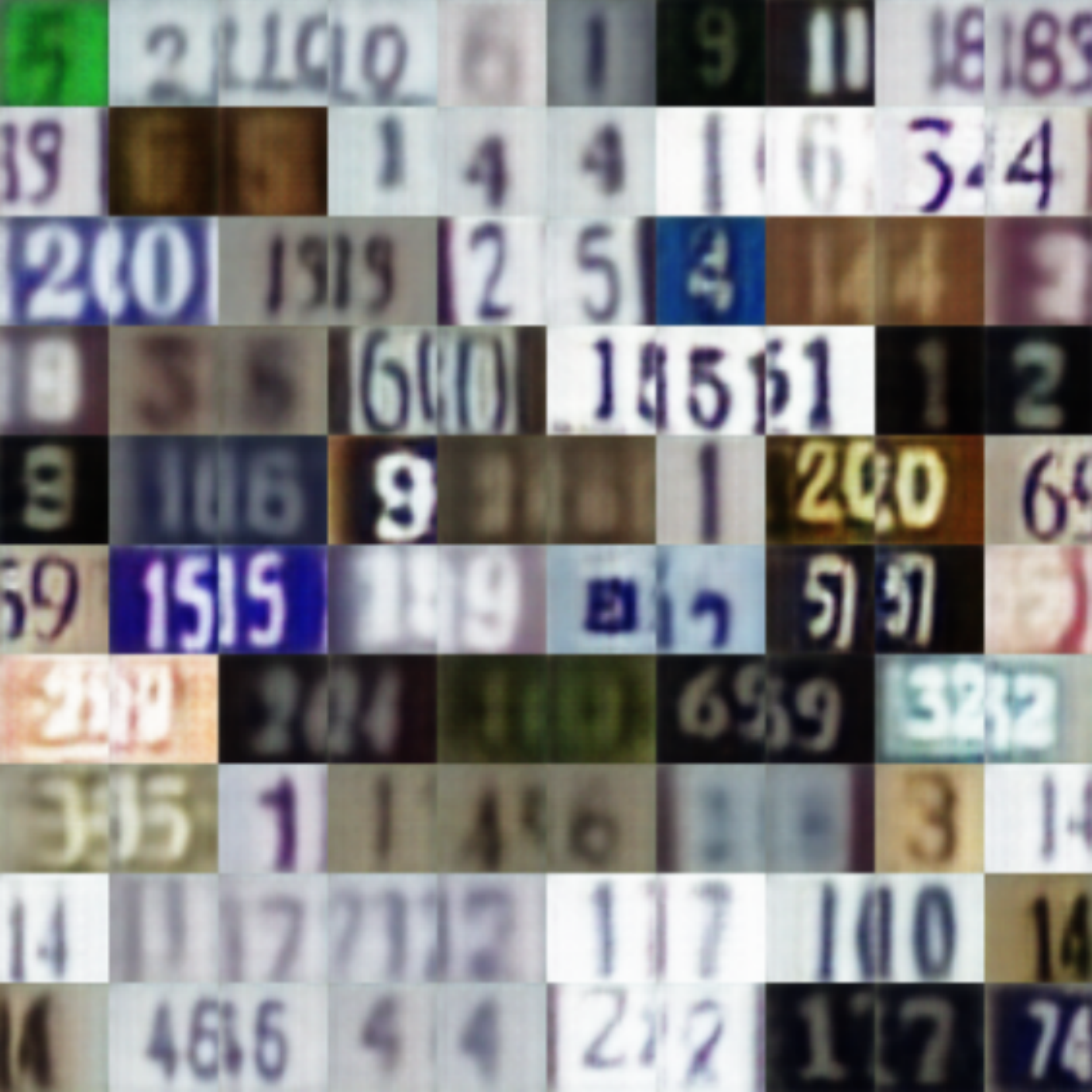}
\end{center}
\caption{\textbf{Original Inputs and Reconstructions:} The first $100$ images from the SVHN validation set (left) reconstructed by VAE-GAN (right).}
\label{fig:svhn-original-reconstructions}
\end{figure}

\begin{table}
{
\tiny
\begin{center}
\begin{tabular}{|c|c|c|c|c|c|c|c|c|c|c|}
\hline
\textbf{Source} & \textbf{Target 0} & \textbf{Target 1} & \textbf{Target 2} & \textbf{Target 3} & \textbf{Target 4} & \textbf{Target 5} & \textbf{Target 6} & \textbf{Target 7} & \textbf{Target 8} & \textbf{Target 9} \\
\hline
\textbf{0} & - & \begin{tabular}{@{}c@{}}64.29\% \\ (40.00\%)\end{tabular} & \begin{tabular}{@{}c@{}}78.57\% \\ (61.43\%)\end{tabular} & \begin{tabular}{@{}c@{}}92.86\% \\ (80.00\%)\end{tabular} & \begin{tabular}{@{}c@{}}84.29\% \\ (57.14\%)\end{tabular} & \begin{tabular}{@{}c@{}}98.57\% \\ (98.57\%)\end{tabular} & \begin{tabular}{@{}c@{}}94.29\% \\ (38.57\%)\end{tabular} & \begin{tabular}{@{}c@{}}88.57\% \\ (54.29\%)\end{tabular} & \begin{tabular}{@{}c@{}}95.71\% \\ (11.43\%)\end{tabular} & \begin{tabular}{@{}c@{}}95.71\% \\ (25.71\%)\end{tabular} \\
\hline
\textbf{1} & \begin{tabular}{@{}c@{}}76.80\% \\ (70.72\%)\end{tabular} & - & \begin{tabular}{@{}c@{}}74.59\% \\ (67.40\%)\end{tabular} & \begin{tabular}{@{}c@{}}93.37\% \\ (88.95\%)\end{tabular} & \begin{tabular}{@{}c@{}}75.69\% \\ (65.19\%)\end{tabular} & \begin{tabular}{@{}c@{}}98.34\% \\ (97.79\%)\end{tabular} & \begin{tabular}{@{}c@{}}86.74\% \\ (24.86\%)\end{tabular} & \begin{tabular}{@{}c@{}}46.96\% \\ (36.46\%)\end{tabular} & \begin{tabular}{@{}c@{}}96.13\% \\ (4.97\%)\end{tabular} & \begin{tabular}{@{}c@{}}96.13\% \\ (28.73\%)\end{tabular} \\
\hline
\textbf{2} & \begin{tabular}{@{}c@{}}82.93\% \\ (65.85\%)\end{tabular} & \begin{tabular}{@{}c@{}}57.93\% \\ (42.68\%)\end{tabular} & - & \begin{tabular}{@{}c@{}}90.24\% \\ (86.59\%)\end{tabular} & \begin{tabular}{@{}c@{}}53.66\% \\ (46.34\%)\end{tabular} & \begin{tabular}{@{}c@{}}99.39\% \\ (98.17\%)\end{tabular} & \begin{tabular}{@{}c@{}}82.93\% \\ (14.02\%)\end{tabular} & \begin{tabular}{@{}c@{}}71.34\% \\ (57.32\%)\end{tabular} & \begin{tabular}{@{}c@{}}71.34\% \\ (6.71\%)\end{tabular} & \begin{tabular}{@{}c@{}}24.39\% \\ (23.17\%)\end{tabular} \\
\hline
\textbf{3} & \begin{tabular}{@{}c@{}}92.17\% \\ (64.35\%)\end{tabular} & \begin{tabular}{@{}c@{}}58.26\% \\ (41.74\%)\end{tabular} & \begin{tabular}{@{}c@{}}83.48\% \\ (68.70\%)\end{tabular} & - & \begin{tabular}{@{}c@{}}84.35\% \\ (49.57\%)\end{tabular} & \begin{tabular}{@{}c@{}}96.52\% \\ (95.65\%)\end{tabular} & \begin{tabular}{@{}c@{}}53.91\% \\ (23.48\%)\end{tabular} & \begin{tabular}{@{}c@{}}90.43\% \\ (56.52\%)\end{tabular} & \begin{tabular}{@{}c@{}}93.04\% \\ (5.22\%)\end{tabular} & \begin{tabular}{@{}c@{}}93.91\% \\ (33.91\%)\end{tabular} \\
\hline
\textbf{4} & \begin{tabular}{@{}c@{}}74.44\% \\ (55.56\%)\end{tabular} & \begin{tabular}{@{}c@{}}47.78\% \\ (43.33\%)\end{tabular} & \begin{tabular}{@{}c@{}}70.00\% \\ (61.11\%)\end{tabular} & \begin{tabular}{@{}c@{}}86.67\% \\ (77.78\%)\end{tabular} & - & \begin{tabular}{@{}c@{}}100.00\% \\ (98.89\%)\end{tabular} & \begin{tabular}{@{}c@{}}93.33\% \\ (35.56\%)\end{tabular} & \begin{tabular}{@{}c@{}}90.00\% \\ (36.67\%)\end{tabular} & \begin{tabular}{@{}c@{}}85.56\% \\ (14.44\%)\end{tabular} & \begin{tabular}{@{}c@{}}94.44\% \\ (27.78\%)\end{tabular} \\
\hline
\textbf{5} & \begin{tabular}{@{}c@{}}75.31\% \\ (50.62\%)\end{tabular} & \begin{tabular}{@{}c@{}}59.26\% \\ (43.21\%)\end{tabular} & \begin{tabular}{@{}c@{}}88.89\% \\ (58.02\%)\end{tabular} & \begin{tabular}{@{}c@{}}97.53\% \\ (88.89\%)\end{tabular} & \begin{tabular}{@{}c@{}}72.84\% \\ (53.09\%)\end{tabular} & - & \begin{tabular}{@{}c@{}}37.04\% \\ (18.52\%)\end{tabular} & \begin{tabular}{@{}c@{}}80.25\% \\ (41.98\%)\end{tabular} & \begin{tabular}{@{}c@{}}32.10\% \\ (6.17\%)\end{tabular} & \begin{tabular}{@{}c@{}}92.59\% \\ (30.86\%)\end{tabular} \\
\hline
\textbf{6} & \begin{tabular}{@{}c@{}}67.44\% \\ (47.67\%)\end{tabular} & \begin{tabular}{@{}c@{}}56.98\% \\ (27.91\%)\end{tabular} & \begin{tabular}{@{}c@{}}84.88\% \\ (55.81\%)\end{tabular} & \begin{tabular}{@{}c@{}}86.05\% \\ (79.07\%)\end{tabular} & \begin{tabular}{@{}c@{}}65.12\% \\ (39.53\%)\end{tabular} & \begin{tabular}{@{}c@{}}94.19\% \\ (94.19\%)\end{tabular} & - & \begin{tabular}{@{}c@{}}90.70\% \\ (33.72\%)\end{tabular} & \begin{tabular}{@{}c@{}}58.14\% \\ (10.47\%)\end{tabular} & \begin{tabular}{@{}c@{}}87.21\% \\ (22.09\%)\end{tabular} \\
\hline
\textbf{7} & \begin{tabular}{@{}c@{}}87.34\% \\ (63.29\%)\end{tabular} & \begin{tabular}{@{}c@{}}55.70\% \\ (48.10\%)\end{tabular} & \begin{tabular}{@{}c@{}}79.75\% \\ (74.68\%)\end{tabular} & \begin{tabular}{@{}c@{}}92.41\% \\ (79.75\%)\end{tabular} & \begin{tabular}{@{}c@{}}69.62\% \\ (41.77\%)\end{tabular} & \begin{tabular}{@{}c@{}}97.47\% \\ (89.87\%)\end{tabular} & \begin{tabular}{@{}c@{}}93.67\% \\ (18.99\%)\end{tabular} & - & \begin{tabular}{@{}c@{}}91.14\% \\ (7.59\%)\end{tabular} & \begin{tabular}{@{}c@{}}97.47\% \\ (17.72\%)\end{tabular} \\
\hline
\textbf{8} & \begin{tabular}{@{}c@{}}98.33\% \\ (63.33\%)\end{tabular} & \begin{tabular}{@{}c@{}}78.33\% \\ (38.33\%)\end{tabular} & \begin{tabular}{@{}c@{}}80.00\% \\ (63.33\%)\end{tabular} & \begin{tabular}{@{}c@{}}100.00\% \\ (88.33\%)\end{tabular} & \begin{tabular}{@{}c@{}}93.33\% \\ (48.33\%)\end{tabular} & \begin{tabular}{@{}c@{}}98.33\% \\ (96.67\%)\end{tabular} & \begin{tabular}{@{}c@{}}96.67\% \\ (35.00\%)\end{tabular} & \begin{tabular}{@{}c@{}}96.67\% \\ (50.00\%)\end{tabular} & - & \begin{tabular}{@{}c@{}}95.00\% \\ (31.67\%)\end{tabular} \\
\hline
\textbf{9} & \begin{tabular}{@{}c@{}}87.88\% \\ (66.67\%)\end{tabular} & \begin{tabular}{@{}c@{}}72.73\% \\ (43.94\%)\end{tabular} & \begin{tabular}{@{}c@{}}92.42\% \\ (80.30\%)\end{tabular} & \begin{tabular}{@{}c@{}}93.94\% \\ (86.36\%)\end{tabular} & \begin{tabular}{@{}c@{}}80.30\% \\ (51.52\%)\end{tabular} & \begin{tabular}{@{}c@{}}95.45\% \\ (93.94\%)\end{tabular} & \begin{tabular}{@{}c@{}}98.48\% \\ (27.27\%)\end{tabular} & \begin{tabular}{@{}c@{}}92.42\% \\ (62.12\%)\end{tabular} & \begin{tabular}{@{}c@{}}93.94\% \\ (9.09\%)\end{tabular} & - \\
\hline
\end{tabular}
\end{center}
}
\caption{\textbf{$L_2$ Optimization Latent Attack on SVHN (single latent vector target):} $\ASuntargeted$\ ($\AStargeted$\ in parentheses) after one reconstruction loop for different source and target class pairs on the VAE-GAN model.
The latent representation of a random image from the target class is used to generate the target latent vector.
Higher values indicate more successful attacks against the generative model.}
\label{tab:svhn-direct-targeted-random-numerical}
\end{table}

\begin{table}[h]
{
\tiny
\begin{center}
\begin{tabular}{|c|c|c|c|c|c|c|c|c|c|c|}
\hline
\textbf{Source} & \textbf{Target 0} & \textbf{Target 1} & \textbf{Target 2} & \textbf{Target 3} & \textbf{Target 4} & \textbf{Target 5} & \textbf{Target 6} & \textbf{Target 7} & \textbf{Target 8} & \textbf{Target 9} \\
\hline
\textbf{0} & - & \begin{tabular}{@{}c@{}}30.00\% \\ (12.86\%)\end{tabular} & \begin{tabular}{@{}c@{}}32.86\% \\ (5.71\%)\end{tabular} & \begin{tabular}{@{}c@{}}34.29\% \\ (5.71\%)\end{tabular} & \begin{tabular}{@{}c@{}}28.57\% \\ (0.00\%)\end{tabular} & \begin{tabular}{@{}c@{}}30.00\% \\ (1.43\%)\end{tabular} & \begin{tabular}{@{}c@{}}30.00\% \\ (5.71\%)\end{tabular} & \begin{tabular}{@{}c@{}}30.00\% \\ (0.00\%)\end{tabular} & \begin{tabular}{@{}c@{}}30.00\% \\ (1.43\%)\end{tabular} & \begin{tabular}{@{}c@{}}31.43\% \\ (0.00\%)\end{tabular} \\
\hline
\textbf{1} & \begin{tabular}{@{}c@{}}13.26\% \\ (1.10\%)\end{tabular} & - & \begin{tabular}{@{}c@{}}7.73\% \\ (1.66\%)\end{tabular} & \begin{tabular}{@{}c@{}}18.78\% \\ (4.97\%)\end{tabular} & \begin{tabular}{@{}c@{}}13.26\% \\ (3.31\%)\end{tabular} & \begin{tabular}{@{}c@{}}12.15\% \\ (0.00\%)\end{tabular} & \begin{tabular}{@{}c@{}}11.60\% \\ (0.55\%)\end{tabular} & \begin{tabular}{@{}c@{}}9.94\% \\ (1.10\%)\end{tabular} & \begin{tabular}{@{}c@{}}10.50\% \\ (1.10\%)\end{tabular} & \begin{tabular}{@{}c@{}}16.02\% \\ (0.55\%)\end{tabular} \\
\hline
\textbf{2} & \begin{tabular}{@{}c@{}}23.17\% \\ (0.61\%)\end{tabular} & \begin{tabular}{@{}c@{}}13.41\% \\ (3.66\%)\end{tabular} & - & \begin{tabular}{@{}c@{}}17.07\% \\ (3.05\%)\end{tabular} & \begin{tabular}{@{}c@{}}14.63\% \\ (1.83\%)\end{tabular} & \begin{tabular}{@{}c@{}}14.63\% \\ (2.44\%)\end{tabular} & \begin{tabular}{@{}c@{}}15.24\% \\ (0.00\%)\end{tabular} & \begin{tabular}{@{}c@{}}15.24\% \\ (1.22\%)\end{tabular} & \begin{tabular}{@{}c@{}}14.02\% \\ (0.61\%)\end{tabular} & \begin{tabular}{@{}c@{}}15.24\% \\ (1.22\%)\end{tabular} \\
\hline
\textbf{3} & \begin{tabular}{@{}c@{}}30.43\% \\ (0.87\%)\end{tabular} & \begin{tabular}{@{}c@{}}26.09\% \\ (7.83\%)\end{tabular} & \begin{tabular}{@{}c@{}}30.43\% \\ (2.61\%)\end{tabular} & - & \begin{tabular}{@{}c@{}}30.43\% \\ (0.00\%)\end{tabular} & \begin{tabular}{@{}c@{}}29.57\% \\ (6.96\%)\end{tabular} & \begin{tabular}{@{}c@{}}27.83\% \\ (0.00\%)\end{tabular} & \begin{tabular}{@{}c@{}}27.83\% \\ (1.74\%)\end{tabular} & \begin{tabular}{@{}c@{}}28.70\% \\ (2.61\%)\end{tabular} & \begin{tabular}{@{}c@{}}33.91\% \\ (6.09\%)\end{tabular} \\
\hline
\textbf{4} & \begin{tabular}{@{}c@{}}21.11\% \\ (0.00\%)\end{tabular} & \begin{tabular}{@{}c@{}}15.56\% \\ (5.56\%)\end{tabular} & \begin{tabular}{@{}c@{}}16.67\% \\ (2.22\%)\end{tabular} & \begin{tabular}{@{}c@{}}25.56\% \\ (4.44\%)\end{tabular} & - & \begin{tabular}{@{}c@{}}16.67\% \\ (1.11\%)\end{tabular} & \begin{tabular}{@{}c@{}}18.89\% \\ (0.00\%)\end{tabular} & \begin{tabular}{@{}c@{}}16.67\% \\ (1.11\%)\end{tabular} & \begin{tabular}{@{}c@{}}18.89\% \\ (2.22\%)\end{tabular} & \begin{tabular}{@{}c@{}}22.22\% \\ (0.00\%)\end{tabular} \\
\hline
\textbf{5} & \begin{tabular}{@{}c@{}}32.10\% \\ (0.00\%)\end{tabular} & \begin{tabular}{@{}c@{}}28.40\% \\ (3.70\%)\end{tabular} & \begin{tabular}{@{}c@{}}27.16\% \\ (3.70\%)\end{tabular} & \begin{tabular}{@{}c@{}}32.10\% \\ (8.64\%)\end{tabular} & \begin{tabular}{@{}c@{}}24.69\% \\ (2.47\%)\end{tabular} & - & \begin{tabular}{@{}c@{}}28.40\% \\ (6.17\%)\end{tabular} & \begin{tabular}{@{}c@{}}23.46\% \\ (0.00\%)\end{tabular} & \begin{tabular}{@{}c@{}}27.16\% \\ (3.70\%)\end{tabular} & \begin{tabular}{@{}c@{}}27.16\% \\ (0.00\%)\end{tabular} \\
\hline
\textbf{6} & \begin{tabular}{@{}c@{}}27.91\% \\ (4.65\%)\end{tabular} & \begin{tabular}{@{}c@{}}25.58\% \\ (4.65\%)\end{tabular} & \begin{tabular}{@{}c@{}}26.74\% \\ (0.00\%)\end{tabular} & \begin{tabular}{@{}c@{}}33.72\% \\ (3.49\%)\end{tabular} & \begin{tabular}{@{}c@{}}30.23\% \\ (2.33\%)\end{tabular} & \begin{tabular}{@{}c@{}}20.93\% \\ (4.65\%)\end{tabular} & - & \begin{tabular}{@{}c@{}}31.40\% \\ (0.00\%)\end{tabular} & \begin{tabular}{@{}c@{}}24.42\% \\ (3.49\%)\end{tabular} & \begin{tabular}{@{}c@{}}32.56\% \\ (0.00\%)\end{tabular} \\
\hline
\textbf{7} & \begin{tabular}{@{}c@{}}30.38\% \\ (0.00\%)\end{tabular} & \begin{tabular}{@{}c@{}}27.85\% \\ (12.66\%)\end{tabular} & \begin{tabular}{@{}c@{}}26.58\% \\ (10.13\%)\end{tabular} & \begin{tabular}{@{}c@{}}31.65\% \\ (5.06\%)\end{tabular} & \begin{tabular}{@{}c@{}}31.65\% \\ (0.00\%)\end{tabular} & \begin{tabular}{@{}c@{}}30.38\% \\ (0.00\%)\end{tabular} & \begin{tabular}{@{}c@{}}32.91\% \\ (0.00\%)\end{tabular} & - & \begin{tabular}{@{}c@{}}30.38\% \\ (0.00\%)\end{tabular} & \begin{tabular}{@{}c@{}}34.18\% \\ (1.27\%)\end{tabular} \\
\hline
\textbf{8} & \begin{tabular}{@{}c@{}}40.00\% \\ (5.00\%)\end{tabular} & \begin{tabular}{@{}c@{}}35.00\% \\ (0.00\%)\end{tabular} & \begin{tabular}{@{}c@{}}33.33\% \\ (3.33\%)\end{tabular} & \begin{tabular}{@{}c@{}}43.33\% \\ (6.67\%)\end{tabular} & \begin{tabular}{@{}c@{}}40.00\% \\ (3.33\%)\end{tabular} & \begin{tabular}{@{}c@{}}35.00\% \\ (1.67\%)\end{tabular} & \begin{tabular}{@{}c@{}}41.67\% \\ (11.67\%)\end{tabular} & \begin{tabular}{@{}c@{}}38.33\% \\ (0.00\%)\end{tabular} & - & \begin{tabular}{@{}c@{}}36.67\% \\ (0.00\%)\end{tabular} \\
\hline
\textbf{9} & \begin{tabular}{@{}c@{}}34.85\% \\ (6.06\%)\end{tabular} & \begin{tabular}{@{}c@{}}33.33\% \\ (12.12\%)\end{tabular} & \begin{tabular}{@{}c@{}}33.33\% \\ (9.09\%)\end{tabular} & \begin{tabular}{@{}c@{}}40.91\% \\ (4.55\%)\end{tabular} & \begin{tabular}{@{}c@{}}31.82\% \\ (3.03\%)\end{tabular} & \begin{tabular}{@{}c@{}}31.82\% \\ (0.00\%)\end{tabular} & \begin{tabular}{@{}c@{}}33.33\% \\ (0.00\%)\end{tabular} & \begin{tabular}{@{}c@{}}34.85\% \\ (0.00\%)\end{tabular} & \begin{tabular}{@{}c@{}}31.82\% \\ (1.52\%)\end{tabular} & - \\
\hline
\end{tabular}
\end{center}
}
\caption{\textbf{$L_2$ Optimization $\calL_{\VAE}$ Attack on SVHN (single image target):} $\ASuntargeted$\ ($\AStargeted$\ in parentheses) after one reconstruction loop for different source and target class pairs on the VAE-GAN model.
The latent representation of a random image from the target class is used to generate the target latent vector.
Higher values indicate more successful attacks against the generative model.}
\label{tab:svhn-lvae-targeted-random-numerical}
\end{table}

\begin{figure}[h]
\begin{center}
\includegraphics[scale=0.4]{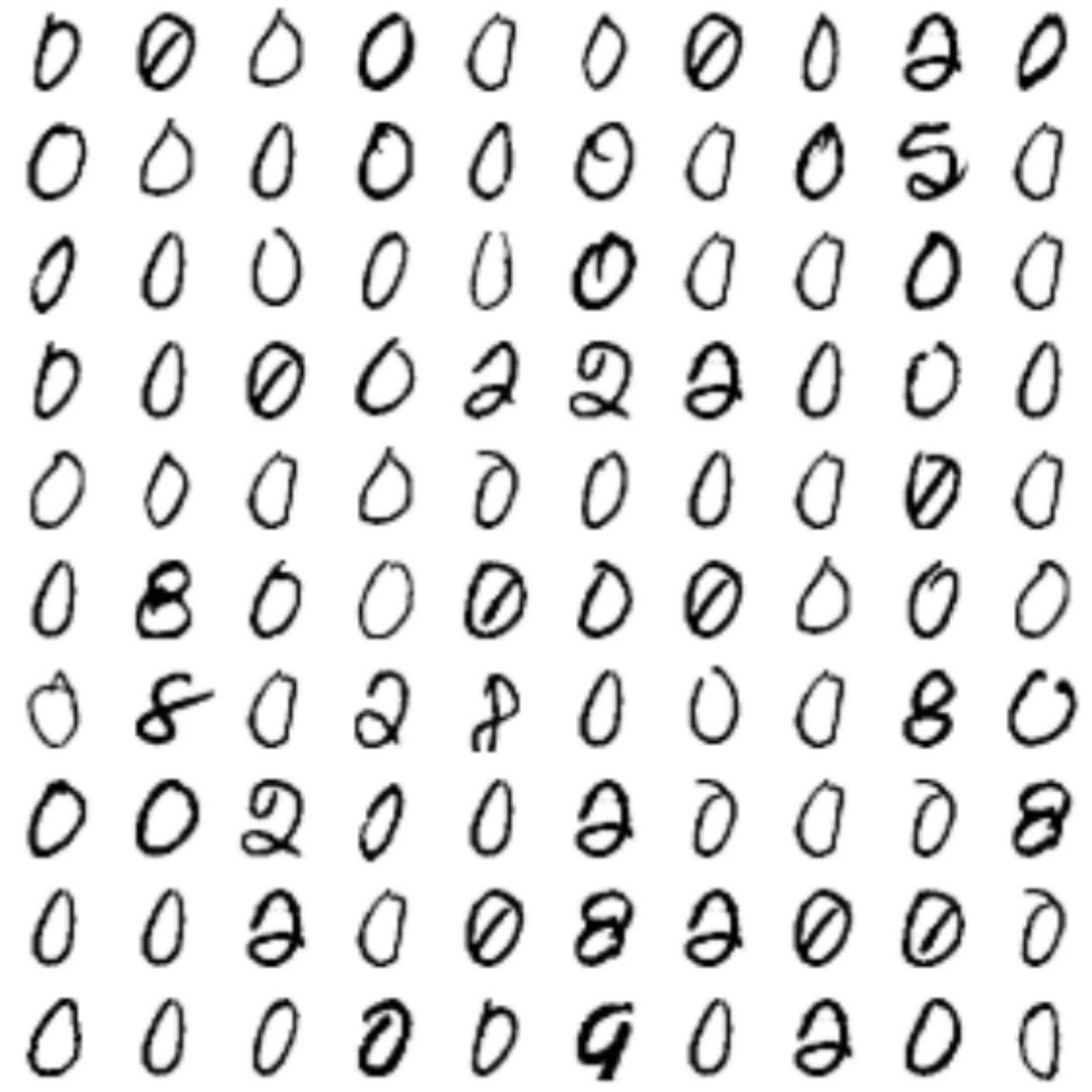}
\rule{0.4pt}{5.7cm}
\includegraphics[scale=0.4]{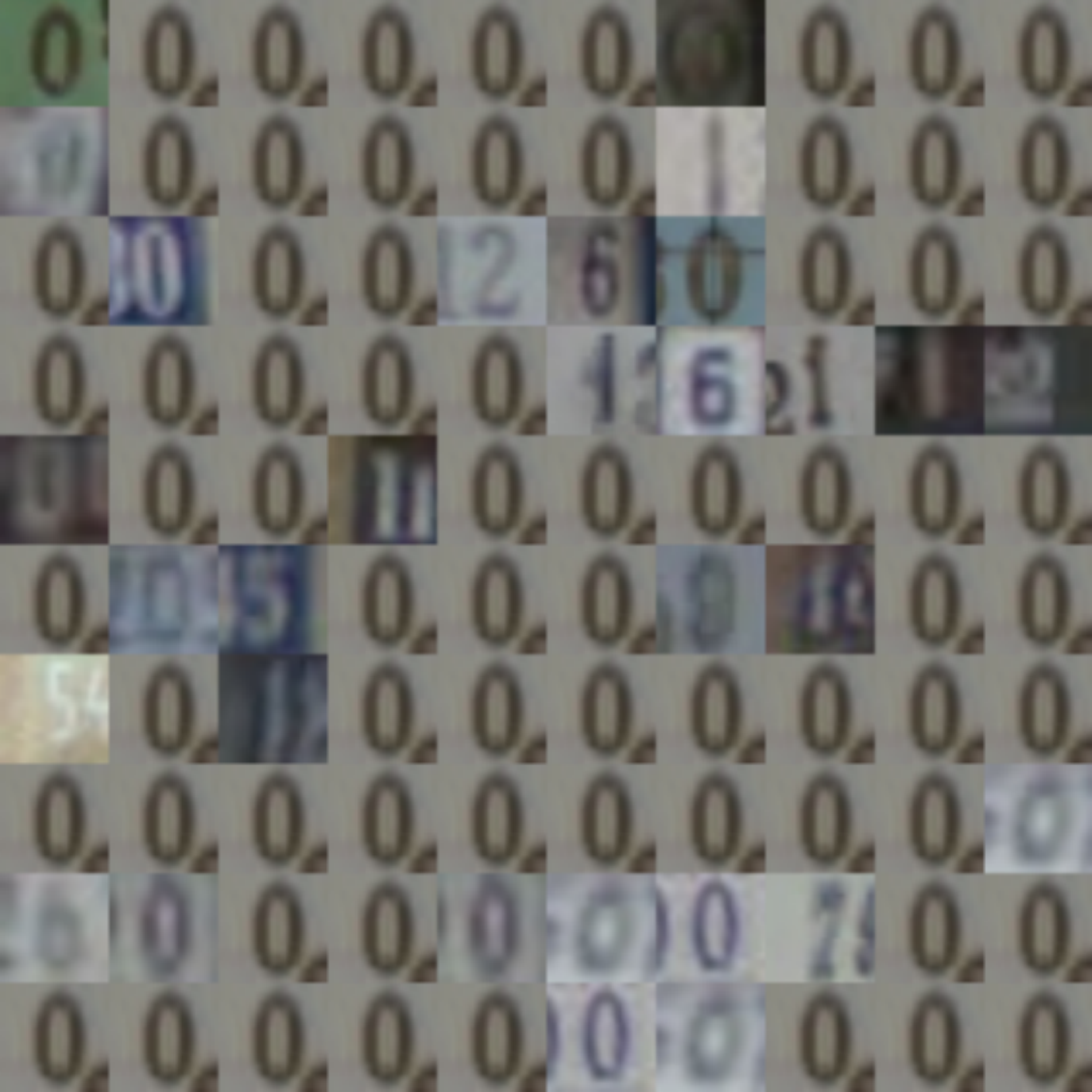}
\end{center}
\caption{\textbf{$L_2$ Optimization Latent Attack (single latent vector target):} Nearest neighbors in latent space for generated adversarial examples (target class $0$) on the first $100$ images from the MNIST (left) and SVHN (right) validation sets.
}
\label{fig:l2-latent-nearest-neighbors}
\end{figure}


\begin{figure}[h]
\begin{center}
\includegraphics[scale=0.4]{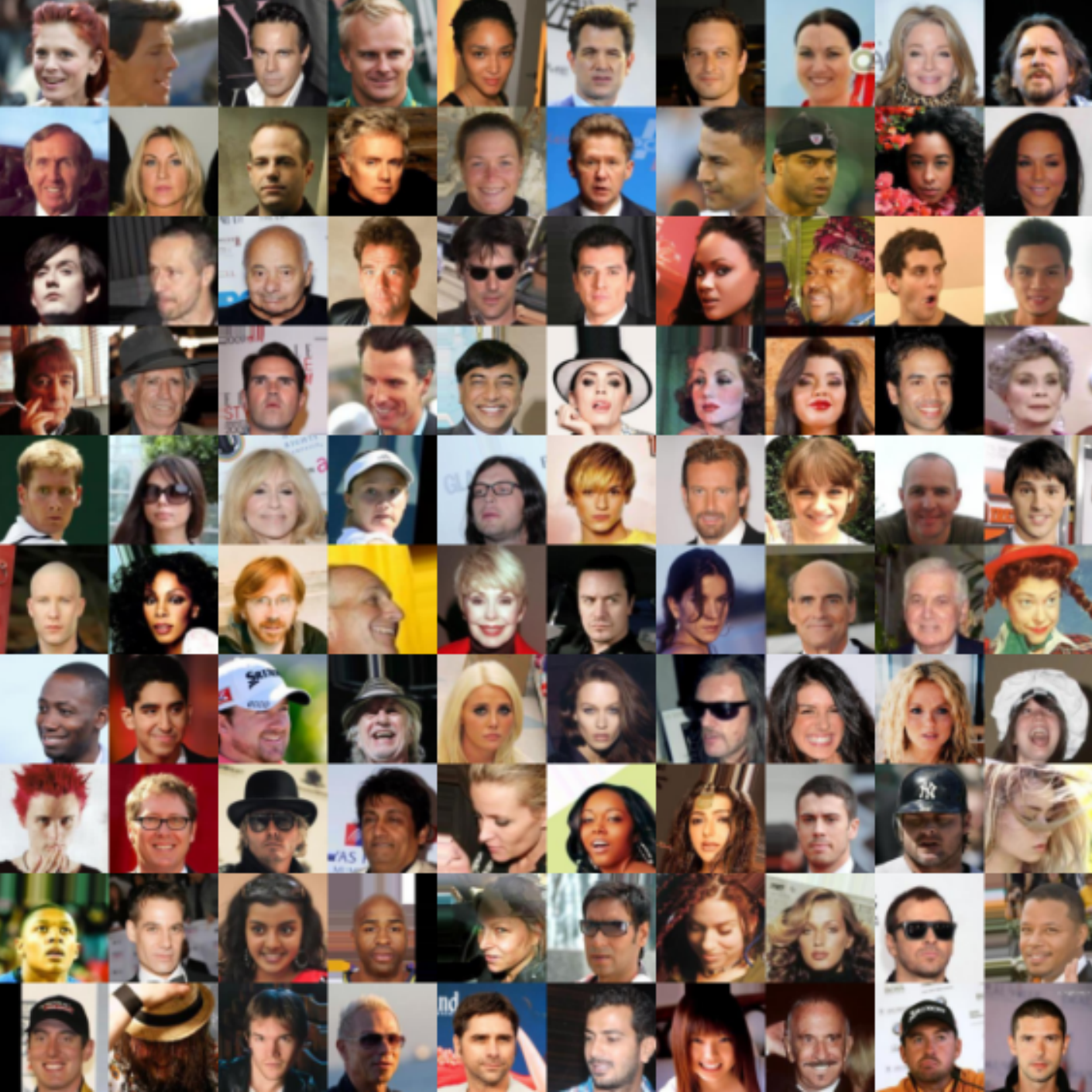}
\rule{0.4pt}{5.7cm}
\includegraphics[scale=0.4]{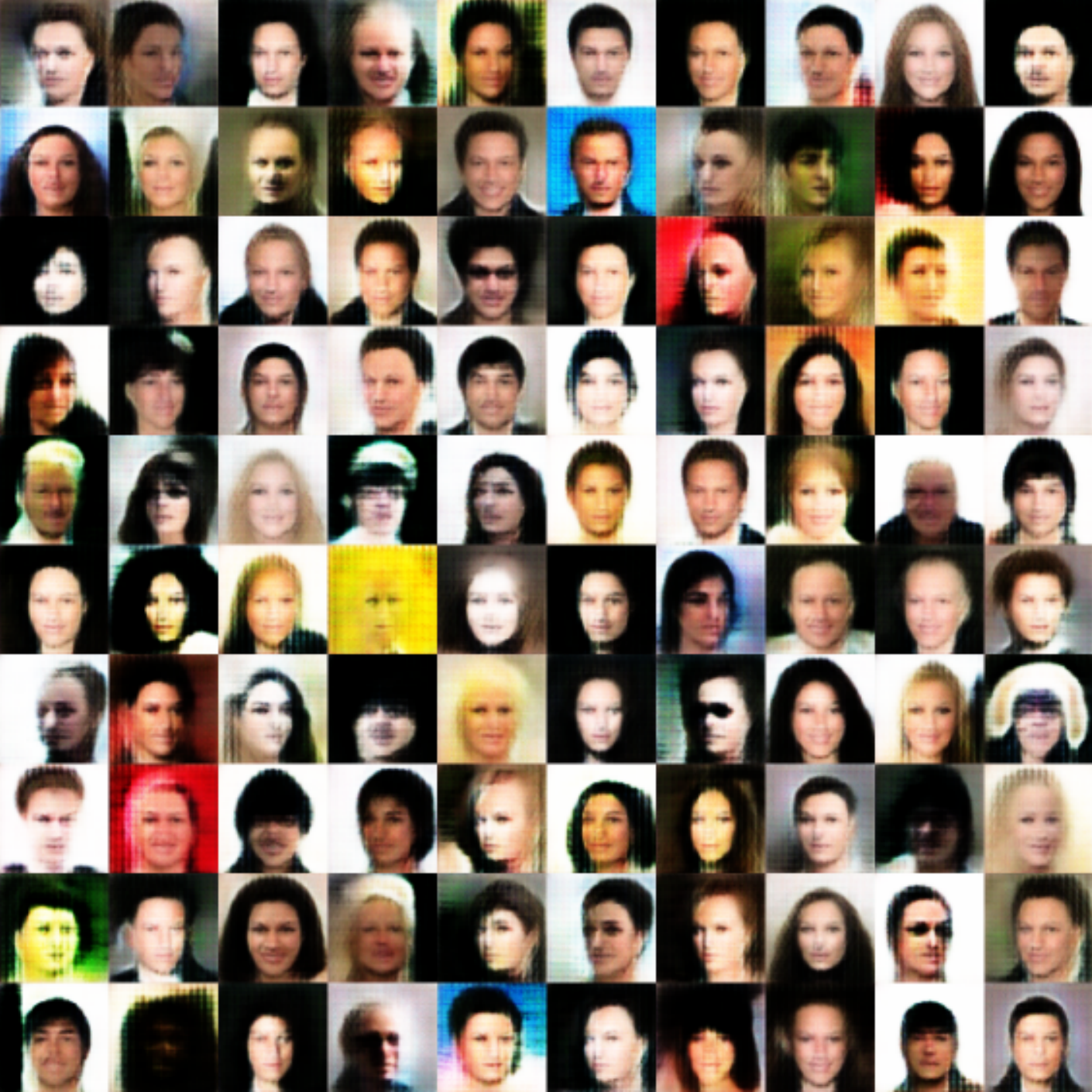}
\end{center}
\caption{Original images in the CelebA dataset (left) and their VAE-GAN reconstructions (right).}
\label{fig:faces-original-reconstructions}
\end{figure}

\begin{figure}[h]
\begin{center}
\includegraphics[scale=0.4]{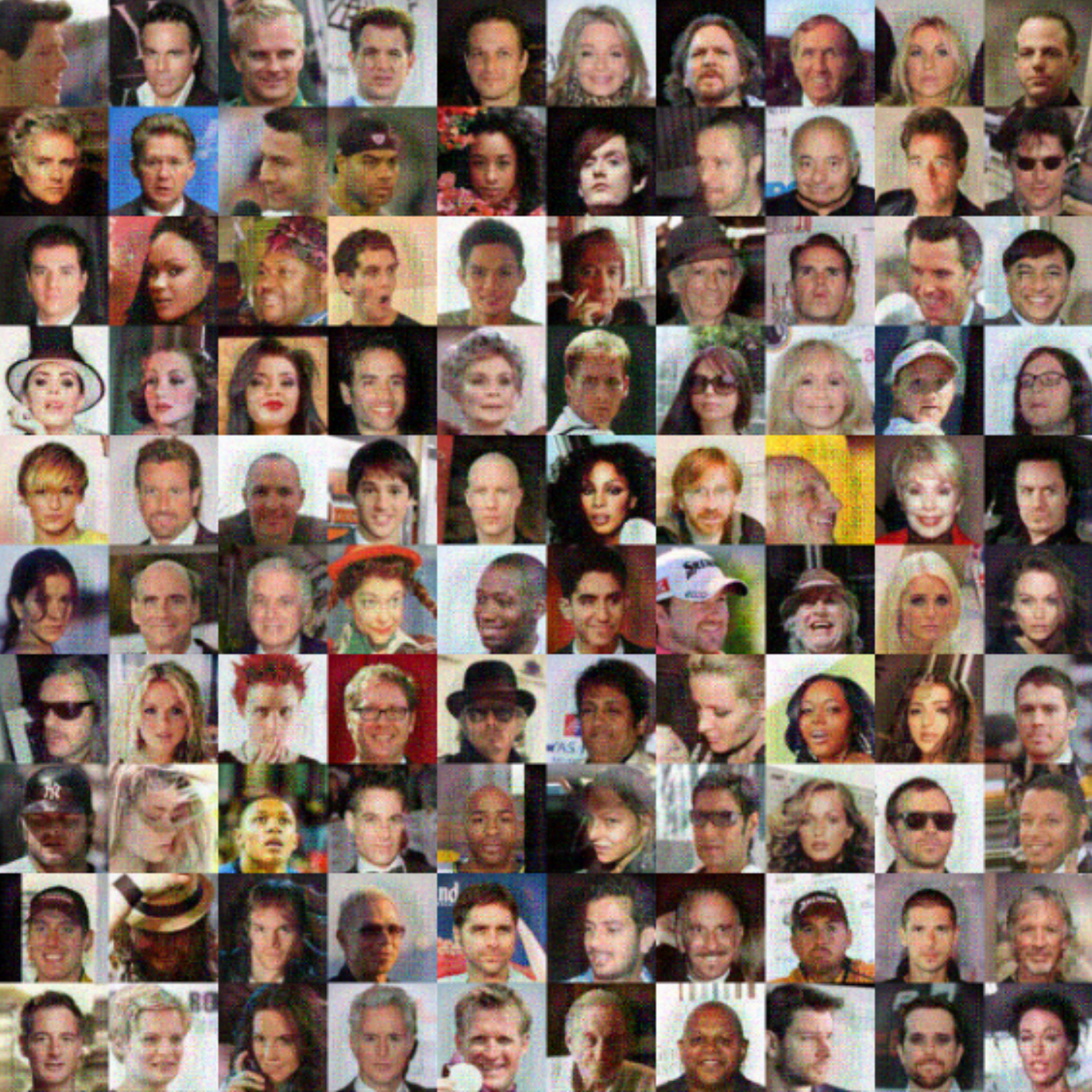}
\rule{0.4pt}{5.7cm}
\includegraphics[scale=0.4]{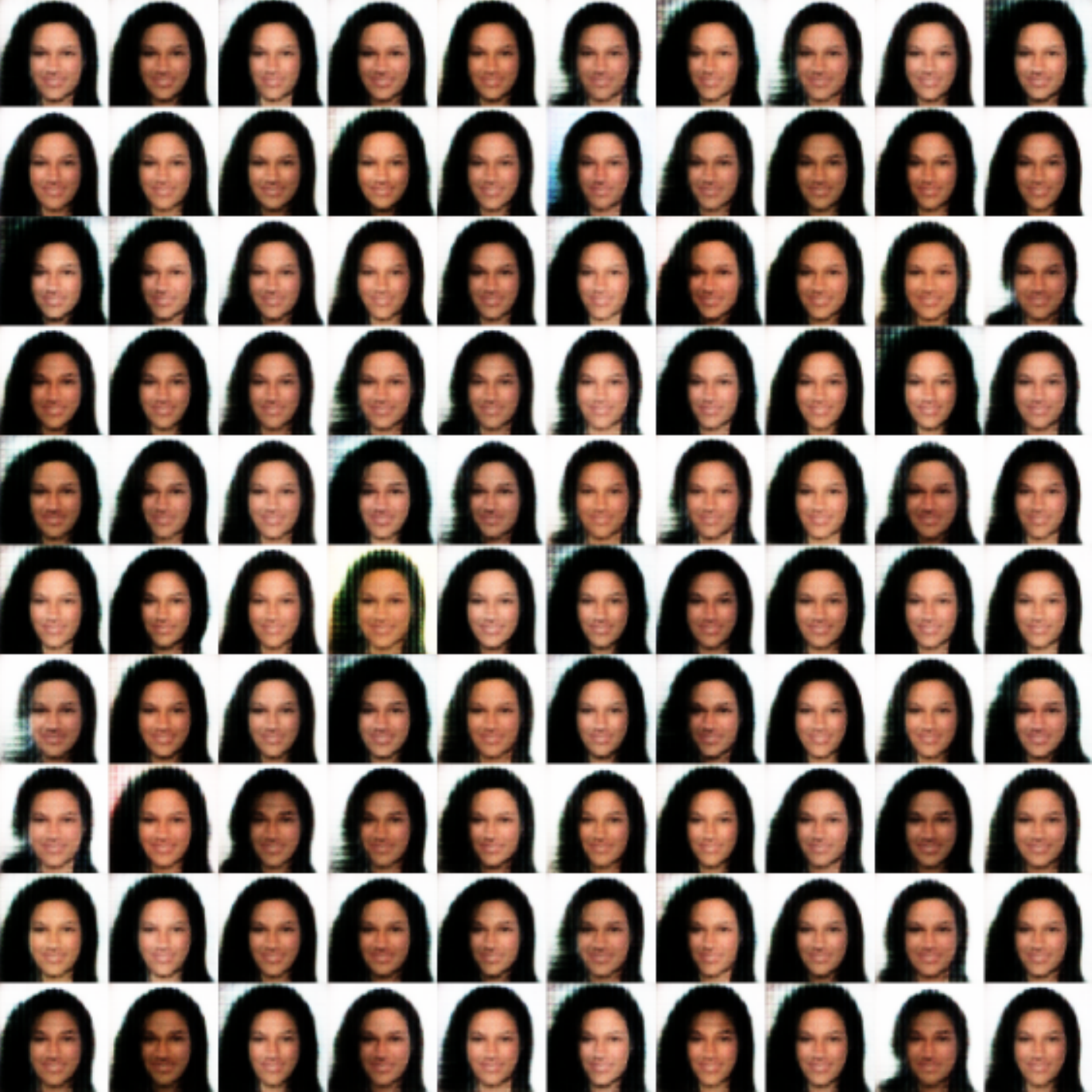}
\end{center}
\caption{\textbf{$L_2$ Optimization Latent Attack on CelebA Dataset (single latent vector target):} Adversarial examples generated for $100$ images from the CelebA dataset (left) and their VAE-GAN reconstructions (right).}
\label{fig:faces-latent-single-vector-target}
\end{figure}

\begin{figure}[h]
\begin{center}
\includegraphics[scale=0.4]{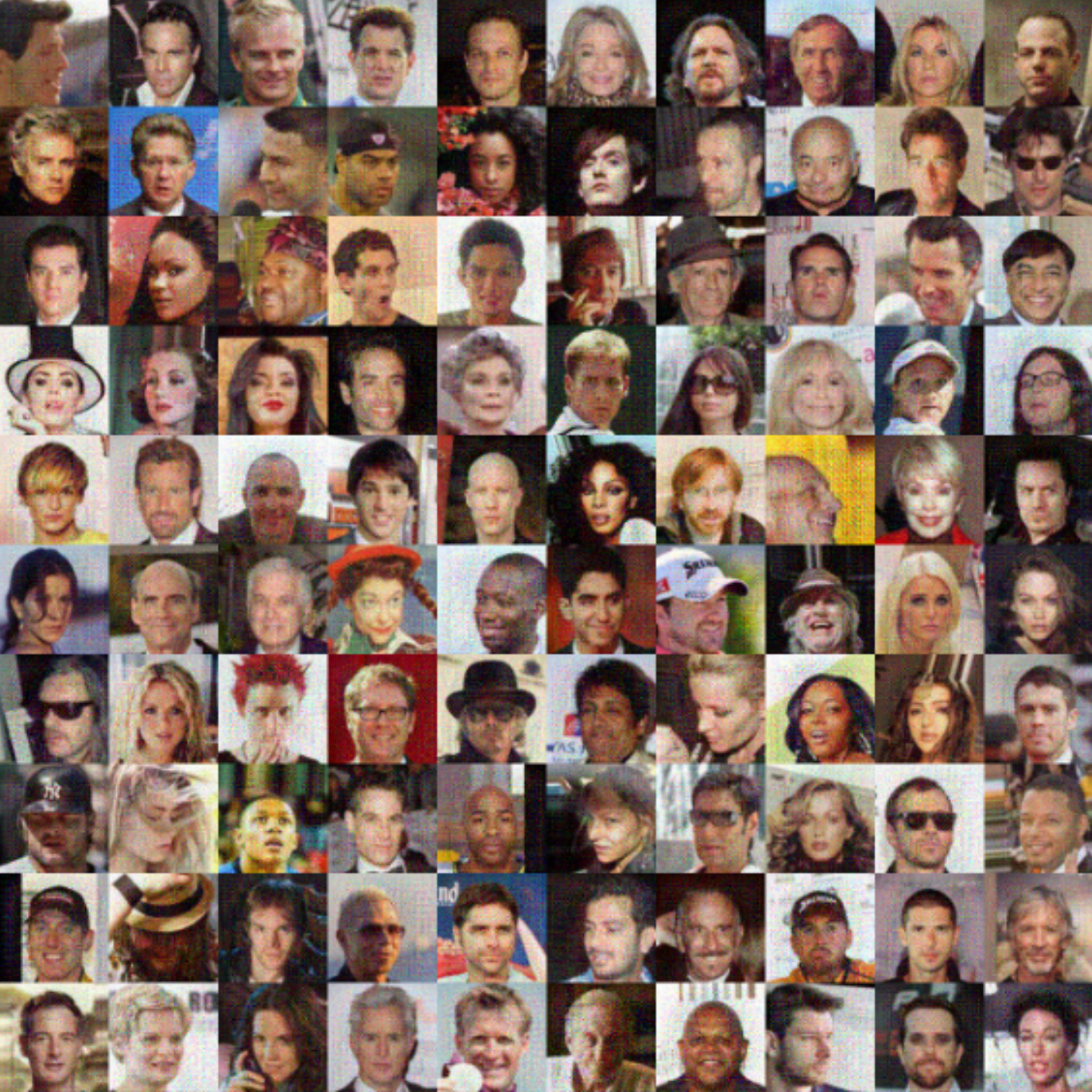}
\rule{0.4pt}{5.7cm}
\includegraphics[scale=0.4]{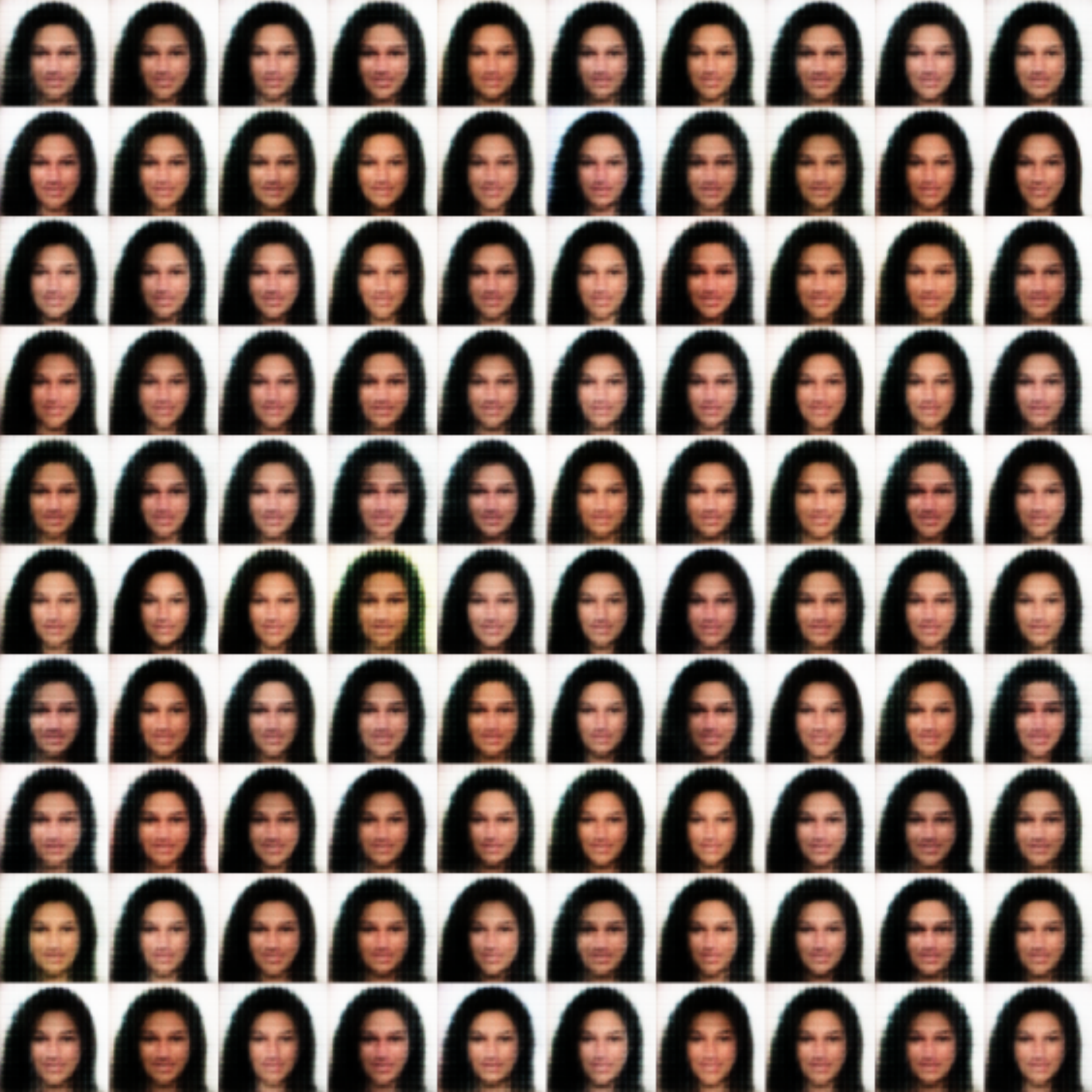}
\end{center}
\caption{\textbf{$L_2$ Optimization $\calL_{\VAE}$ Attack on CelebA Dataset (single image target):} Adversarial examples generated for $100$ images from the CelebA dataset (left) and their VAE-GAN reconstructions (right).}
\label{fig:faces-lvae-single-image-target}
\end{figure}

\begin{figure}[h]
\begin{center}
\begin{tabular}{c c c c c c c c c c}
~~Orig~~ & ~Mean~ & 1 Smp & 12 Smp & 50 Smp & $L_2$ Adv & Mean & 1 Smp & 12 Smp & 50 Smp \\
\end{tabular}
\includegraphics[scale=0.73]{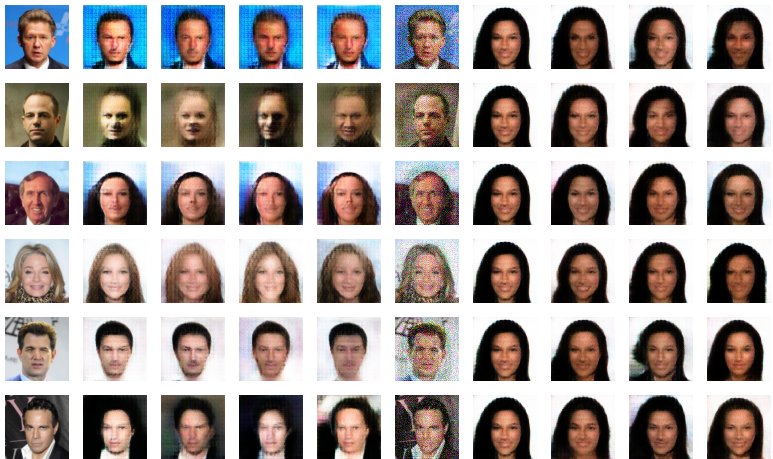}
\end{center}
\caption{Effect of sampling on adversarial reconstructions. Columns in order: original image, reconstruction of the original image (no sampling, just the mean), reconstruction of the original image (1 sample), reconstruction of the original image (12 samples), reconstruction of the original image (50 samples), adversarial example (latent attack), reconstruction of the adversarial example (no sampling, just the mean), reconstruction of the adversarial example (1 sample), reconstruction of the adversarial example (12 samples), reconstruction of the adversarial example (50 samples).}
\label{fig:faces-sampling}
\end{figure}